\documentclass[11pt]{article}
\usepackage[formal]{zihan}   

\usepackage{multirow}
\usepackage{tabularx}
\usepackage{url}
\usepackage{colortbl}
\usepackage{array}
\usepackage{float}
\usepackage{placeins}
\usepackage{adjustbox}

\tcbset{before skip=4pt, after skip=4pt}

\newcommand{\NoSkill}{\textsc{NoSkill}}
\newcommand{\Direct}{\textsc{Direct}}
\newcommand{\SkillOpt}{\textsc{SkillOpt}}
\newcommand{\TraceToSkill}{\textsc{Trace2Skill}}
\definecolor{GainBlue}{HTML}{1F5AA6}
\definecolor{HarmRed}{HTML}{B42318}
\newcommand{\gain}[1]{\begingroup\color{GainBlue}\bfseries\boldmath #1\endgroup}
\newcommand{\harm}[1]{\begingroup\color{HarmRed}\bfseries\boldmath #1\endgroup}
\newcommand{\rev}[1]{#1}
\newenvironment{revblock}{\par}{\par}
\newtcolorbox{finding}[1]{enhanced,colback=zboxbody,colframe=zboxhead,boxrule=0.6pt,arc=6pt,
  left=10pt,right=10pt,top=6pt,bottom=6pt,
  title={#1},fonttitle=\zheadfont\bfseries\small,coltitle=zviolettext,colbacktitle=zboxhead,
  titlerule=0pt,toptitle=3pt,bottomtitle=3pt}

\title{Which Failure Does an Intervention Repair? When Agent Interventions Help, and Where Self-Evolving Skills Stall}
\shorttitle{Which Failure Does an Intervention Repair?}
\renewcommand{\zauthorsep}{\hskip 1em plus 0.5em minus 0.2em\relax}
\authors{Zihan Jin\textsuperscript{1}, Gaoyuan Du\textsuperscript{2*\textdagger}, Jiate Li\textsuperscript{1*}, Dongze Ye\textsuperscript{1*}, Shawn Li\textsuperscript{1}, Yue Zhao\textsuperscript{1}, Defu Cao\textsuperscript{1}}
\affiliations{\textsuperscript{1}University of Southern California \quad \textsuperscript{2}Amazon}
\date{December 2, 2026}
\codeurl{\url{https://github.com/Zihan-Zoe/same-skill-different-harness}}
\dataurl{\url{https://huggingface.co/datasets/Zihan-Zoe/same-skill-different-harness}}
\email{\href{mailto:zihanjin@usc.edu}{zihanjin@usc.edu}}
\venue{\textsuperscript{*}Listed in alphabetical order by last name. \quad \textsuperscript{\textdagger}This work does not relate to the author's position at Amazon.}

\begin{document}
\maketitle

\begin{abstract}
An intervention that improves one agent can break another, and adding a stronger harness on top need not repair the damage. Agent interventions such as skills, harnesses, and extra inference budget are nonetheless usually credited with one average gain, measured under one setup. We ask instead when each intervention helps and which failure it repairs. A matched factorial study over nine benchmarks, four models, four skill conditions, four harness levels, and three budgets, with about 0.72 million trajectories, changes one axis at a time. No configuration is best everywhere, interventions interact rather than add, and a harness can absorb a skill's gain, rescue a failure the skill causes, or leave the gain intact. The failure left by the current system points to the fix: missing knowledge to a stronger model, execution errors to a richer harness. Yet feedback is not recovery: a matched recovery repairs mainly failures that leave a visible error signal. Skills that evolve from the agent's own experience hit a bottleneck: they rarely beat a short human-written skill, can learn dataset shortcuts, break when moved to another model, and cannot supply missing facts, which outside information does. \rev{A benchmark-level switch to the model favored by a 20-task labeled pilot achieves both higher performance and lower cost than always using the stronger model.} Agent interventions should be judged by which failure they repair and under which configuration.
\end{abstract}

\section{Introduction}
\label{sec:intro}

\rev{Can a skill built for one model break another?} A skill optimized for score on Qwen3.5-9B was transferred, unchanged, to a stronger model. On Hotpot Context, that model answers 62.0\% of questions correctly without the skill and 13.6\% with it. A stronger harness that feeds back errors and allows a revision does not repair it: the score stays at 12.2. The model still knows the answer. In 74.1\% of its responses with the skill a correct answer is present, against 76.8\% without, but the skill makes the model fill its answer field with a fact-extraction worksheet instead of the answer. The intervention did not add a capability. It interacted with a model it was not built for and with an answer interface that the harness enforces.

\begin{figure*}[t]
\centering
\includegraphics[width=\linewidth,height=0.85\textheight,keepaspectratio]{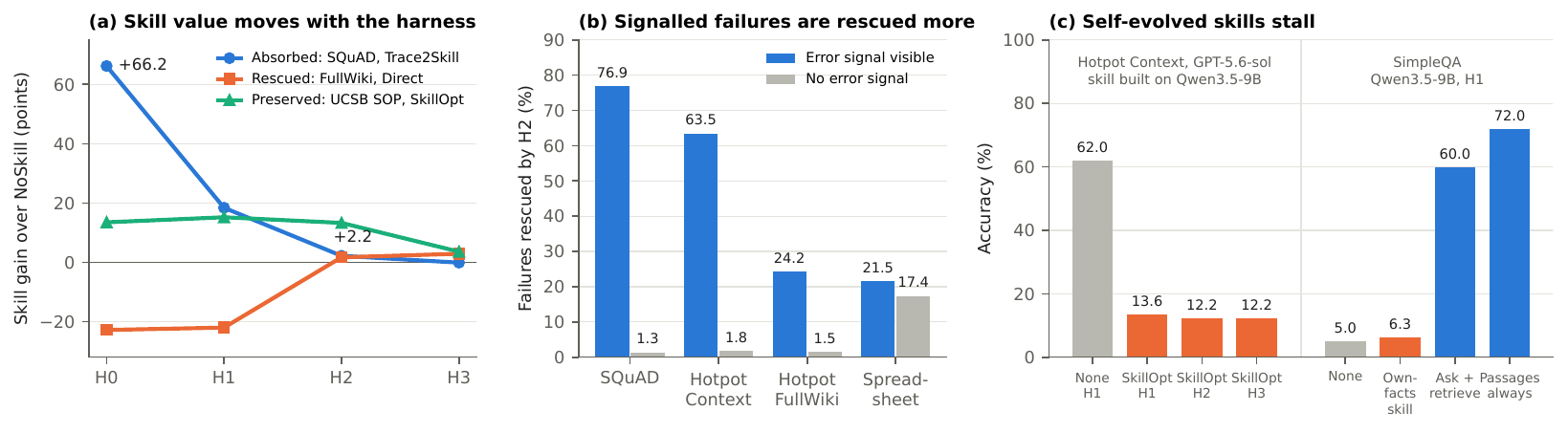}
\caption{\rev{The \rev{first} three findings at a glance. (a) With Qwen3.5-9B, the same frozen skill can lose its gain as the harness rises (absorbed), turn a loss into a gain (rescued), or keep its gain (preserved) (Table~\ref{tab:regimes}). (b) Among Qwen3.5-9B H1 failures, those whose own trajectory shows an error signal are rescued by H2 far more often on the three QA benchmarks, but only modestly more often on Spreadsheet (21.5\% versus 17.4\%) (Section~\ref{sec:which}). (c) A skill optimized on Qwen3.5-9B collapses on GPT-5.6-sol, and richer harnesses do not restore it; on SimpleQA, a skill built from the agent's own facts barely helps, whereas retrieved passages do (Section~\ref{sec:selfevo}).}}
\label{fig:overview}
\end{figure*}

This case is not an anomaly. Modern agents wrap a language model in a stack of interventions: reusable skills \citep{wang2023voyager}, tool interfaces and execution feedback \citep{yao2022react,yang2024swe}, reflection and revision \citep{shinn2023reflexion,madaan2023self}, extra inference budget \citep{snell2025scaling}, and, increasingly, skills that the agent evolves from its own trajectories \citep{yang2026skillopt,ni2026trace2skill}. Each new intervention is reported as a gain under its authors' model, harness, and budget. Agent benchmarks score the stack as a whole \citep{liu2024agentbench,ma2024agentboard,yao2024tau}. Neither tells us whether a gain belongs to the intervention, which failure it repaired, or whether it survives a change in the rest of the system.

We therefore ask: \emph{when does an intervention help, and which failure does it repair?} Answering this requires changing one intervention at a time on the same task instances. We build a matched factorial matrix over task instance, model, skill, harness, and budget (Section~\ref{sec:design}): 1,872 test instances from nine benchmarks, four models, four skill conditions (no skill, a one-shot self-generated skill, and two self-evolving methods, all built by the same model and frozen before testing), four harness levels, and three budgets, most cells run three times. The matrix contains 718,848 trajectories. The study supports \rev{four} groups of findings\rev{, summarized in Figure~\ref{fig:overview}}.
\begin{itemize}[leftmargin=*,topsep=2pt,itemsep=2pt]
\item \emph{When interventions help (Section~\ref{sec:when}).} No configuration is best everywhere, and interventions interact rather than add. Even with the model fixed, a harness can absorb a skill's gain, rescue a failure the skill causes, or leave the gain intact: an 18-word SQuAD skill adds 66.2 points under a minimal harness and 2.2 under execution feedback.
\item \emph{Which failure they repair (Section~\ref{sec:which}).} The failure left by the current system points to the fix, but feedback is not recovery: a matched recovery repairs mainly failures that leave a visible error signal.
\item \emph{Where self-evolution stalls (Section~\ref{sec:selfevo}).} Skills evolved from the agent's own experience rarely beat a short human-written skill, can learn dataset shortcuts, break when moved to another model (the opening case), and cannot supply missing facts.
\item \rev{\emph{How to choose an intervention (Section~\ref{sec:choose}).} Choosing per task does not pay: neither a router nor escalation on failure signals beats a simple fixed configuration. Choosing per benchmark does: a small labeled pilot beats always switching to a stronger model at lower cost.}
\end{itemize}

\section{A Matched Factorial Study}
\label{sec:design}

\begin{figure}[t]
\centering
\includegraphics[width=\linewidth,height=0.85\textheight,keepaspectratio]{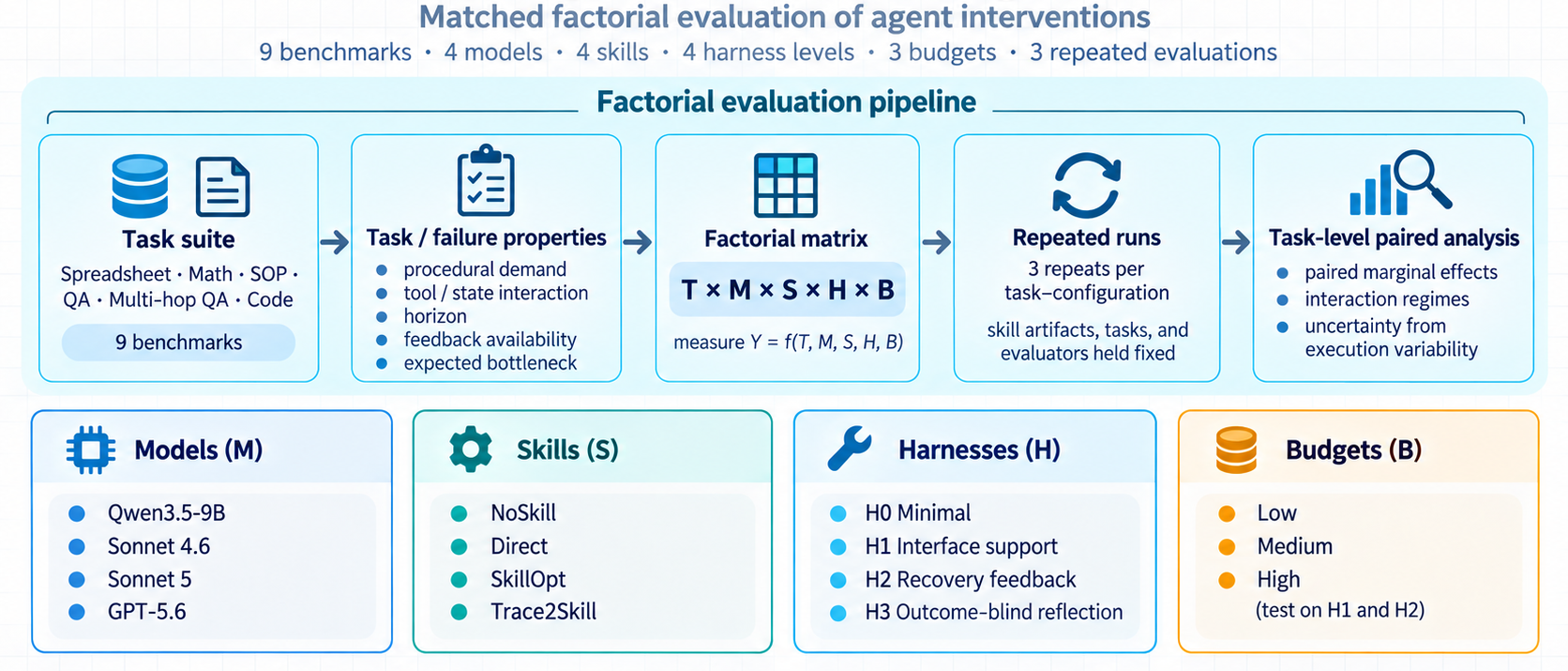}
\caption{The matched factorial design. Each comparison changes one axis (model, skill, harness, or budget) on the same task instances and holds the others fixed.}
\label{fig:design}
\end{figure}

The study crosses five factors (Figure~\ref{fig:design}). \textbf{Tasks} are 1,872 test instances from nine benchmarks whose dominant bottlenecks differ: missing knowledge (SimpleQA), an answer contract (SQuAD), multi-hop integration with supplied evidence or with search (Hotpot Context and FullWiki), code (BigCodeBench-Hard), procedure following (Amazon SOP and UCSB SOP), spreadsheet manipulation (SpreadsheetBench), and multiple-choice mathematics (LiveMathematicianBench). \textbf{Models} are Qwen3.5-9B, Sonnet~4.6, Sonnet~5, and GPT-5.6-sol. \textbf{Skills} are \NoSkill{}; \Direct{}, written in one call from training items; \SkillOpt{}, optimized from scored rollouts; and \TraceToSkill{}, distilled from successful and failed trajectories. \rev{In the main matrix,} Qwen3.5-9B builds every skill from training items only, and each skill is frozen and reused unchanged for every model\rev{; Section~\ref{sec:selfevo} adds a stronger constructor, GPT-5.6-sol, on two benchmarks}. The \textbf{harness} rises one mechanism at a time (Table~\ref{tab:ladder}), and the \textbf{budget} sweeps Low, Medium, and High caps on tokens, turns, and time for H1 and H2. Every cell is run three times, for nearly 0.9 million trajectories.

\begin{table}[t]
\caption{The harness ladder. Each level adds one mechanism to the level above it; no level sees evaluator outcomes or gold answers.}
\label{tab:ladder}
\centering
\footnotesize
\begin{tabularx}{\linewidth}{@{}lX@{}}
\toprule
H0 Minimal & The benchmark's own prompt, with the skill appended when present \\
H1 Interface & $+$ a standardized answer or tool interface with legal-action guidance \\
H2 Execution & $+$ execution errors or observations fed back, with up to three bounded retries \\
H3 Reflection & $+$ one revision of the response or trajectory that does not see the outcome \\
\bottomrule
\end{tabularx}
\end{table}

A comparison changes exactly one axis on the same task instances and repeats, so every difference is paired. For each failed execution we also record why it failed, using logged errors, evaluator records, and answer rules, and we freeze that label before any matched outcome is inspected. An independent judge and an author check of 200 sampled labels validate them; LiveMath, whose labels the independent judge does not reproduce, is excluded from every failure-conditioned analysis. Appendix~\ref{app:measurement-details} gives each benchmark's harness implementation, budget ceilings, splits, metrics, and statistical methods; Appendix~\ref{app:skill-provenance} gives skill construction and provenance; and Appendix~\ref{sec:root-cause-appendix} gives the failure taxonomy and label validation.

\section{When Interventions Help}
\label{sec:when}

\providecommand{\rowbest}[1]{\textbf{\textcolor{GainBlue}{#1}}}
\providecommand{\blockbest}[1]{\cellcolor{HarmRed!15}#1}
\providecommand{\benchbest}[1]{\cellcolor{HarmRed!15}\begingroup\setlength{\fboxsep}{0.8pt}\fbox{\strut #1}\endgroup}
\begin{table*}[!t]
\centering
\caption{Combined Default-budget Skill $\times$ Harness performance matrix. Rows follow the Benchmark $\times$ Skill layout; columns group each Model into H0--H3 Harness conditions. Each entry is the mean benchmark-native performance on a 0--100 scale over three complete repeats. \textbf{\textcolor{GainBlue}{Bold blue}},  a \colorbox{HarmRed!15}{red-shaded cell}, and a \fbox{box} mark the row, block and benchmark maxima respectively.}
\label{tab:default-matrix-transposed}
\scriptsize
\setlength{\tabcolsep}{1.7pt}
\renewcommand{\arraystretch}{0.8}
\resizebox{\textwidth}{!}{%
\begin{tabular}{@{}ll*{16}{r}@{}}
\toprule
Benchmark & Skill / contrast & \multicolumn{4}{c}{Qwen3.5-9B} & \multicolumn{4}{c}{Sonnet 4.6} & \multicolumn{4}{c}{Sonnet 5} & \multicolumn{4}{c}{GPT-5.6-sol} \\
\cmidrule(lr){3-6}\cmidrule(lr){7-10}\cmidrule(lr){11-14}\cmidrule(lr){15-18}
 & & H0 & H1 & H2 & H3 & H0 & H1 & H2 & H3 & H0 & H1 & H2 & H3 & H0 & H1 & H2 & H3 \\
\midrule
\multirow{4}{*}{Spreadsheet} & \NoSkill{} & 12.5 & 9.3 & 24.9 & \blockbest{30.4} & 48.6 & 49.2 & 51.1 & \benchbest{\blockbest{\rowbest{53.1}}} & 33.5 & 31.9 & \blockbest{34.5} & 34.2 & 35.1 & 35.0 & 36.0 & \blockbest{37.5} \\
 & \Direct{} & 14.3 & 13.3 & 28.6 & 28.8 & 45.8 & 43.3 & 45.8 & \rowbest{48.9} & 30.1 & 31.4 & 34.0 & 28.2 & 33.7 & {33.0} & 35.0 & 36.0 \\
 & \SkillOpt{} & {8.0} & 9.3 & 22.5 & 26.2 & {34.2} & 39.6 & \rowbest{40.4} & 36.1 & 30.2 & 30.5 & 33.2 & 29.0 & 33.3 & 34.0 & 34.8 & 37.3 \\
 & \TraceToSkill{} & 9.3 & 10.1 & 22.7 & 25.1 & 34.5 & 36.7 & 38.0 & \rowbest{39.3} & 26.1 & 25.7 & 26.9 & {24.5} & 33.2 & 33.7 & 34.5 & 36.5 \\
\midrule
\multirow{4}{*}{LiveMath} & \NoSkill{} & 34.1 & 34.9 & 34.4 & 36.3 & 31.7 & 32.0 & 30.9 & 31.7 & 40.1 & 38.7 & 37.9 & 35.2 & 52.2 & \rowbest{53.5} & 51.9 & {49.2} \\
 & \Direct{} & 33.3 & 32.3 & 33.3 & {30.9} & {28.5} & 29.8 & 31.5 & 29.8 & 36.3 & 36.6 & 38.4 & 37.4 & 50.0 & \rowbest{51.9} & 49.7 & 51.1 \\
 & \SkillOpt{} & 40.3 & 40.6 & 39.8 & 40.6 & 31.5 & 33.6 & 33.1 & 35.5 & 34.7 & 37.4 & 37.1 & {32.5} & 55.4 & 55.1 & \rowbest{58.6} & 54.3 \\
 & \TraceToSkill{} & 53.2 & \blockbest{58.1} & 56.5 & 54.6 & 33.9 & 34.4 & 33.6 & \blockbest{37.1} & \blockbest{40.6} & 39.0 & 36.8 & 35.5 & 61.6 & \benchbest{\blockbest{\rowbest{62.1}}} & 57.8 & 60.5 \\
\midrule
\multirow{4}{*}{UCSB SOP} & \NoSkill{} & 52.8 & {51.0} & 52.6 & 59.2 & 61.7 & 60.2 & 60.2 & 60.2 & 66.7 & 65.9 & 66.7 & 64.2 & 69.9 & 69.3 & \rowbest{70.3} & 68.4 \\
 & \Direct{} & 54.1 & 57.5 & 55.2 & 60.4 & 59.0 & 58.7 & 59.4 & 63.0 & 66.5 & 65.5 & 66.1 & {61.3} & 68.2 & \rowbest{68.6} & 68.2 & {66.9} \\
 & \SkillOpt{} & \blockbest{66.3} & \blockbest{66.3} & 65.9 & 62.9 & \blockbest{67.4} & 67.2 & 65.9 & 65.1 & 68.6 & 67.8 & \blockbest{69.3} & 67.6 & 72.6 & \benchbest{\blockbest{\rowbest{73.1}}} & 71.6 & 70.3 \\
 & \TraceToSkill{} & 62.1 & 61.9 & 61.9 & 62.1 & 58.9 & {57.5} & 60.6 & 62.9 & 65.1 & 66.1 & 65.0 & 64.0 & \rowbest{71.8} & 70.3 & 71.2 & 68.2 \\
\midrule
\multirow{4}{*}{Amazon SOP} & \NoSkill{} & 28.3 & 33.9 & 88.6 & 95.1 & 82.7 & 82.7 & \benchbest{\blockbest{\rowbest{100.0}}} & 99.6 & 2.6 & 4.2 & 58.9 & 87.8 & 58.5 & 57.1 & \benchbest{\blockbest{\rowbest{100.0}}} & 99.0 \\
 & \Direct{} & {1.7} & {1.7} & {1.7} & {1.7} & 90.8 & 80.2 & \rowbest{99.4} & 99.2 & 2.6 & {1.7} & 85.7 & \blockbest{89.5} & 57.8 & 57.8 & 95.7 & 99.0 \\
 & \SkillOpt{} & 8.8 & 6.7 & 94.8 & \blockbest{98.2} & \benchbest{\blockbest{\rowbest{100.0}}} & \benchbest{\blockbest{\rowbest{100.0}}} & \benchbest{\blockbest{\rowbest{100.0}}} & \benchbest{\blockbest{\rowbest{100.0}}} & {1.7} & 9.0 & 61.5 & 80.0 & 59.1 & 58.5 & \benchbest{\blockbest{\rowbest{100.0}}} & 99.3 \\
 & \TraceToSkill{} & 19.6 & 8.6 & 80.6 & 93.3 & {77.1} & 82.9 & \benchbest{\blockbest{\rowbest{100.0}}} & 99.9 & 9.4 & 6.8 & 58.5 & 85.1 & 57.1 & {55.9} & 98.0 & 99.3 \\
\midrule
\multirow{4}{*}{SimpleQA} & \NoSkill{} & 10.7 & 5.1 & {4.7} & 5.0 & \blockbest{33.8} & 30.1 & 29.9 & 29.7 & \blockbest{31.6} & 29.2 & 30.4 & 29.6 & \benchbest{\blockbest{\rowbest{69.4}}} & 67.8 & 68.0 & 68.8 \\
 & \Direct{} & 8.0 & 6.3 & 6.1 & 6.4 & {28.2} & 29.7 & 30.1 & 29.3 & 27.6 & 25.0 & {24.6} & 25.7 & {65.0} & 66.1 & 67.0 & \rowbest{69.2} \\
 & \SkillOpt{} & \blockbest{11.3} & 6.3 & 6.3 & 5.9 & 33.4 & 30.1 & 30.1 & 32.9 & 27.7 & 30.9 & 29.3 & 30.8 & 67.6 & 68.0 & 66.6 & \rowbest{68.4} \\
 & \TraceToSkill{} & 9.9 & 6.0 & 5.4 & 5.6 & 33.3 & 29.8 & 29.0 & 31.1 & 30.3 & 31.2 & 31.2 & 28.0 & 67.8 & 67.4 & 66.9 & \rowbest{68.8} \\
\midrule
\multirow{4}{*}{SQuAD} & \NoSkill{} & {10.7} & 36.6 & 78.6 & 79.3 & {0.0} & 85.4 & 85.6 & 86.3 & {1.8} & \rowbest{90.0} & 89.0 & 89.4 & {23.9} & 63.4 & 63.1 & 68.3 \\
 & \Direct{} & 46.7 & 76.4 & 75.9 & 76.3 & 42.4 & 80.4 & 80.3 & 82.0 & 37.6 & \rowbest{89.2} & 88.3 & 87.4 & 59.3 & 73.6 & 74.4 & 77.2 \\
 & \SkillOpt{} & 29.0 & 77.0 & 77.6 & 77.4 & {0.0} & 82.4 & 82.1 & 82.6 & 50.3 & \rowbest{87.1} & 86.9 & 84.7 & 49.7 & 63.8 & 64.1 & 66.6 \\
 & \TraceToSkill{} & 76.9 & 55.0 & \blockbest{80.8} & 79.2 & 16.1 & 86.7 & 86.9 & \blockbest{87.2} & 69.3 & \benchbest{\blockbest{\rowbest{91.0}}} & 90.6 & 90.7 & 72.8 & 80.6 & 80.8 & \blockbest{81.1} \\
\midrule
\multirow{4}{*}{Hotpot Context} & \NoSkill{} & 18.6 & 59.3 & \blockbest{59.7} & 58.6 & {0.0} & 66.2 & 66.2 & \rowbest{67.8} & 3.4 & 64.3 & 64.6 & \blockbest{65.8} & 22.6 & 62.0 & 61.9 & 61.3 \\
 & \Direct{} & 21.3 & 55.2 & 55.3 & 57.3 & 16.6 & 66.8 & 66.9 & \benchbest{\blockbest{\rowbest{69.2}}} & 9.9 & 62.9 & 62.7 & 64.8 & 29.7 & \blockbest{64.6} & 64.3 & 60.9 \\
 & \SkillOpt{} & {0.0} & 44.0 & 47.9 & \blockbest{59.7} & {0.0} & 68.2 & 68.4 & \rowbest{68.6} & {0.0} & 63.9 & 63.0 & 64.9 & {0.0} & 13.6 & 12.2 & 12.2 \\
 & \TraceToSkill{} & 26.2 & 58.4 & 58.6 & 59.2 & 1.0 & 68.6 & 68.1 & \rowbest{68.7} & 17.6 & 62.8 & 62.1 & 65.4 & 28.0 & 61.6 & 62.0 & 61.3 \\
\midrule
\multirow{4}{*}{Hotpot FullWiki} & \NoSkill{} & 24.2 & 25.9 & 26.8 & 26.9 & 29.1 & 30.4 & 29.3 & 31.1 & 31.1 & 30.9 & 31.3 & \rowbest{33.2} & 21.0 & 22.6 & 23.1 & 19.9 \\
 & \Direct{} & {1.4} & 3.9 & 28.4 & \rowbest{29.8} & 26.7 & 26.1 & 26.7 & 28.0 & 27.2 & 26.7 & 27.8 & 27.4 & 27.2 & 27.2 & 27.2 & 26.8 \\
 & \SkillOpt{} & 21.1 & 24.0 & 34.9 & \rowbest{35.6} & {19.8} & 20.7 & 21.0 & 23.9 & {20.1} & 24.0 & 23.6 & 24.4 & {15.2} & 20.0 & 17.1 & 16.8 \\
 & \TraceToSkill{} & 32.7 & 34.1 & 36.0 & \blockbest{39.1} & \blockbest{44.0} & 41.6 & 41.7 & 42.7 & 41.1 & \benchbest{\blockbest{\rowbest{45.0}}} & 43.2 & 41.9 & 31.6 & 30.1 & \blockbest{32.6} & 28.9 \\
\bottomrule
\end{tabular}}
\end{table*}

\begin{figure}[t]
\centering
\includegraphics[width=0.88\linewidth,height=0.85\textheight,keepaspectratio]{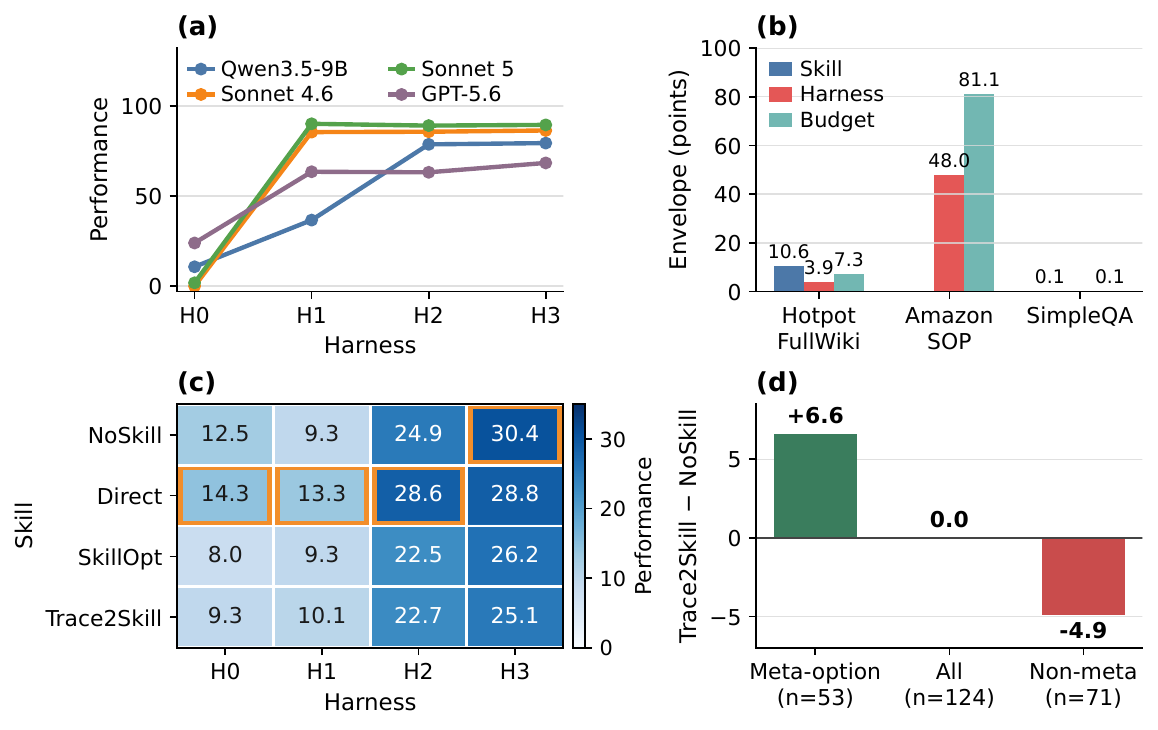}
\caption{Intervention response is heterogeneous. (a) SQuAD without a skill: the best model changes with the harness. (b) Within-axis envelopes for three benchmarks: each responds to a different axis. (c) Spreadsheet with Qwen3.5-9B: the best skill (outlined) changes with the harness. (d) LiveMath with Sonnet~5: \TraceToSkill{} helps on instances with a recurring answer option and hurts on the rest.}
\label{fig:heterogeneity}
\end{figure}

\paragraph{No configuration is best everywhere.} Across the 512 Default-budget cells (Table~\ref{tab:default-matrix-transposed}), the best model changes with the benchmark, the skill, and the harness; skills, harnesses, and budgets help on some tasks and do little or harm on others; and within one benchmark, the same intervention can help some instances and hurt others (Figure~\ref{fig:heterogeneity}). On LiveMath with Sonnet~5, \TraceToSkill{} gains 6.60 points on the 53 instances that carry a recurring correct answer option and loses 4.93 on the other 71, so its average effect is near zero.

\paragraph{Interventions interact rather than add.} If effects simply added up, each cell's score would follow from a model effect, a skill effect, and a harness effect alone. Letting each pair of axes interact instead predicts held-out cells better on seven of eight benchmarks (mean leave-one-cell-out $R^2$ from 0.729 to 0.812; paired sign-flip $p=0.0078$; Appendix Figure~\ref{fig:interaction}). Appendix~\ref{app:interaction-details} gives the fitting procedure and the test.

\paragraph{Skill value moves with the harness.} The skill--harness interaction is the easiest to trace, because the model and the frozen skill stay fixed while the harness rises one mechanism at a time. For Qwen3.5-9B, three regimes recur (Table~\ref{tab:regimes}).
\begin{itemize}[leftmargin=*,topsep=2pt,itemsep=2pt]
\item \emph{Absorbed.} Asked which company won a contest to have its ad shown free during Super Bowl 50, the model without a skill answers in a full sentence and fails exact match; with an 18-word skill, \emph{``Use only the supplied context and return the shortest span that answers the question,''} it answers ``Death Wish Coffee''. At H1 the model without a skill gives the same short answer, because the answer interface imposes the same contract. The skill's gain falls from 66.2 at H0 to 2.2 at H2, and adding H1 to the skill even lowers the score from 76.9 to 55.0. Re-scored by whether a response contains a gold answer, \NoSkill{} already reaches 96.1 at H0: the absorbed gain is answer format.
\item \emph{Rescued.} On Hotpot FullWiki, \Direct{} leads the agent into repeated malformed actions until the turn limit, so its gain is $-22.0$ at H1. At H2 each malformed action is returned as an error, and the gain becomes $+1.7$; 215 of the 221 failure-to-success transitions in this cell follow such feedback.
\item \emph{Preserved.} On UCSB SOP, the agent without a skill executes a payment whose precondition fails but raises no error, so H2 has nothing to feed back. \SkillOpt{} adds the missing check and keeps a 13.3--15.2-point gain through H0--H2.
\end{itemize}
Across the four models, SQuAD absorption and Hotpot FullWiki preservation recur; the rescued cells are specific to Qwen3.5-9B, because the stronger models show no large skill-induced penalty for H2 to remove. \rev{Extended to the full matrix, the same classification reproduces the Qwen3.5-9B regimes of Table~\ref{tab:regimes} and finds 18/96 skill--model--benchmark combinations absorbed, 12/96 preserved, and 4/96 rescued (Table~\ref{tab:regime-prevalence}).} Appendix~\ref{app:skill-harness-details} reports the regimes for every model and the format-versus-content re-scoring.

\begin{table}[t]
\caption{Skill gain over \NoSkill{} at each harness level for Qwen3.5-9B (points, Default budget). The interaction column is the change in gain from H0 to the level where the regime appears, with a 95\% instance-cluster bootstrap interval.}
\label{tab:regimes}
\centering
\scriptsize
\resizebox{\linewidth}{!}{%
\begin{tabular}{lllrrrrl}
\toprule
Regime & Benchmark & Skill & H0 & H1 & H2 & H3 & Interaction [95\% CI] \\
\midrule
Absorbed & SQuAD & \TraceToSkill{} & $+66.2$ & $+18.4$ & $+2.2$ & $-0.1$ & H2: $-64.0$ [$-69.9$, $-58.2$] \\
 & SQuAD & \Direct{} & $+36.0$ & $+39.9$ & $-2.7$ & $-3.0$ & H2: $-38.7$ [$-44.6$, $-32.4$] \\
\midrule
Rescued & Hotpot FullWiki & \Direct{} & $-22.8$ & $-22.0$ & $+1.7$ & $+2.9$ & H2: $+24.4$ [$+19.1$, $+29.8$] \\
 & Amazon SOP & \SkillOpt{} & $-19.5$ & $-27.2$ & $+6.2$ & $+3.0$ & H2: $+25.7$ [$+12.1$, $+39.8$] \\
\midrule
Preserved & Hotpot FullWiki & \TraceToSkill{} & $+8.4$ & $+8.2$ & $+9.2$ & $+12.2$ & H2: $+0.8$ [$-2.6$, $+4.0$] \\
 & UCSB SOP & \SkillOpt{} & $+13.5$ & $+15.2$ & $+13.3$ & $+3.6$ & H2: $-0.2$ [$-4.8$, $+4.4$] \\
\bottomrule
\end{tabular}}
\end{table}

\begin{figure*}[t]
\centering
\includegraphics[width=\linewidth,height=0.85\textheight,keepaspectratio]{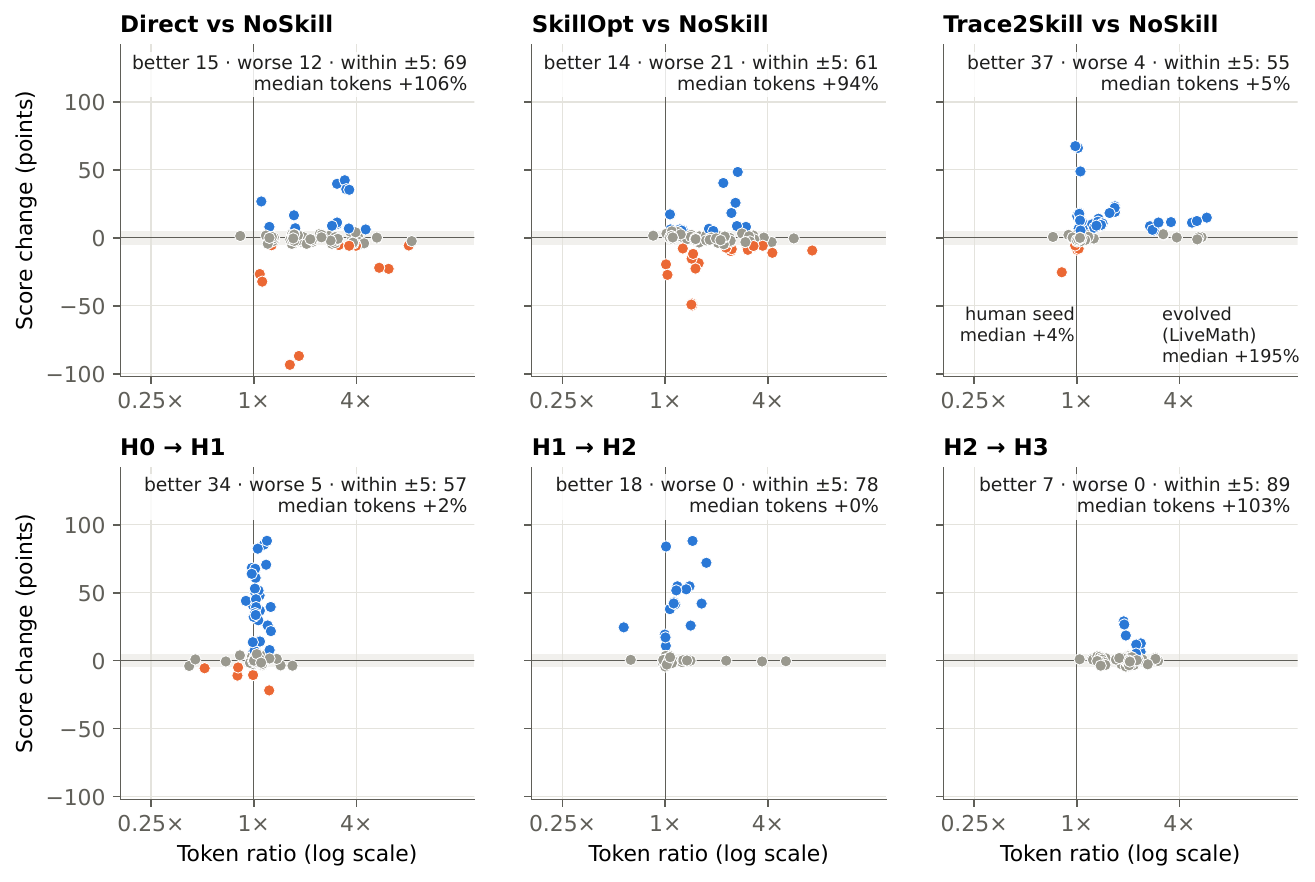}
\caption{\rev{Quality and cost of each intervention. Each point is one matched comparison at the Default budget: a skill against \NoSkill{} at the same model, benchmark, and harness, or one harness step at the same model, benchmark, and skill (96 per panel; four models, six benchmarks). The x-axis is the ratio of mean total tokens per instance; the y-axis is the change in native score. Blue and orange points change the score by at least five points up or down; the grey band marks $\pm5$. Spreadsheet and UCSB SOP log no tokens and are excluded; for GPT-5.6-sol, whose input-token field is not recorded, total tokens are used for every model. The \TraceToSkill{} panel shows the human-written seed on every benchmark except LiveMath, where the evolved candidate passed its validation gate.}}
\label{fig:quality-cost}
\end{figure*}

\paragraph{\rev{Interventions also differ in cost.}} \rev{On the six benchmarks with token telemetry, Figure~\ref{fig:quality-cost} sets each intervention's score change against its change in total tokens on the same cells, for all four models. \Direct{} and \SkillOpt{} roughly double token use (median $+106\%$ and $+94\%$), leave the score within five points in 69 and 61 of 96 comparisons, and when they do move it, hurt about as often as they help (15 better and 12 worse; 14 and 21). \TraceToSkill{} looks cheap (median $+5\%$) only because on five of these six benchmarks its validation gate kept the human-written seed of 17--85 words (Table~\ref{tab:trace2skill-gate}); that seed helps in 28 of 80 comparisons and hurts in 4, whereas the evolved LiveMath skill, at 2,603 words, costs a median $+195\%$. The H0$\rightarrow$H1 and H1$\rightarrow$H2 steps are nearly free in median tokens ($+2\%$ and $+0\%$), and H1$\rightarrow$H2 never lowers the score by five points or more, whereas the H3 revision doubles token use ($+103\%$) and helps in only 7. A single accuracy number hides these prices.}

\begin{finding}{Finding 1: Intervention value is conditional}
No configuration is best everywhere, and interventions interact rather than add. A harness can absorb a skill's gain, rescue a failure the skill causes, or leave the gain intact, so a skill gain reported without its harness cannot be interpreted.
\end{finding}

\section{Which Failure an Intervention Repairs}
\label{sec:which}

\begin{table*}[!t]
\centering
\caption{Room to help on each axis, per benchmark (points). $\Delta_S^+$, $\Delta_H^+$, and $\Delta_B^+$ are the largest gains of any alternative skill, any harness above H0, and any budget above Low; $\Delta_M$ is the best-to-worst span across models. Bold blue marks the largest envelope per row; the columns are not comparable across axes. $^\ddagger$Contains native-score floor cells. Relaxed $\Delta_H^+$ re-scores the same trajectories with a relaxed answer check (SQuAD and Hotpot Context: the response contains a gold answer; Amazon SOP: field-level relaxed parser); -- not re-scored.}
\label{tab:benchmark-marginal-summary}
\small
\begin{adjustbox}{max width=\linewidth,max totalheight=0.85\textheight}
\begin{tabular}{lrrrrr}
\toprule
Benchmark & $\Delta_S^+$ & $\Delta_H^+$ & Relaxed $\Delta_H^+$ & $\Delta_B^+$ & $\Delta_M$ \\
\midrule
Spreadsheet & 0.00 & 5.54 & -- & 4.18 & \textbf{\textcolor{GainBlue}{24.33}} \\
LiveMath & 8.15 & 0.79 & -- & 2.67 & \textbf{\textcolor{GainBlue}{22.26}} \\
UCSB SOP & 5.55 & 0.00 & -- & NA & \textbf{\textcolor{GainBlue}{10.42}} \\
Amazon SOP$^\ddagger$ & 0.00 & 48.03 & 46.33 & \textbf{\textcolor{GainBlue}{81.13}} & 53.10 \\
SimpleQA & 0.13 & 0.00 & -- & 0.10 & \textbf{\textcolor{GainBlue}{60.85}} \\
SQuAD$^\ddagger$ & 17.13 & \textbf{\textcolor{GainBlue}{44.34}} & 2.12 & 0.24 & 11.88 \\
Hotpot Context$^\ddagger$ & 1.71 & \textbf{\textcolor{GainBlue}{48.18}} & 0.00 & 0.00 & 9.57 \\
Hotpot FullWiki & \textbf{\textcolor{GainBlue}{10.58}} & 3.92 & -- & 7.26 & 6.99 \\
\bottomrule
\end{tabular}
\end{adjustbox}

\end{table*}

\paragraph{The bottleneck points to the axis.} Averaged over a benchmark, the axis with room to help follows its bottleneck (Table~\ref{tab:benchmark-marginal-summary}). SimpleQA moves only with the model (a 60.85-point span across models, against at most 0.13 points from any skill), UCSB SOP is led by the skill, and on Spreadsheet and Amazon SOP the harness gain sits at the step from H1 to H2, where execution feedback lets a failed step be retried. On SQuAD and Hotpot Context, the large harness gain is answer format: re-scored on answer content, their harness envelopes fall from 44.34 to 2.12 and from 48.18 to 0.00, whereas Amazon SOP's barely moves (48.03 to 46.33). The case studies in Appendix~\ref{app:benchmark-case-studies} read each benchmark in turn: its expected bottleneck and observed failure labels, its best configuration, and the mechanism behind it. Spreadsheet, for example, is limited by execution, so a capable model with execution feedback beats any added skill (Appendix~\ref{app:case-spreadsheet}), whereas SimpleQA moves only with the model (Appendix~\ref{app:case-simpleqa}).

\begin{figure}[t]
\centering
\includegraphics[width=0.8\linewidth,height=0.85\textheight,keepaspectratio]{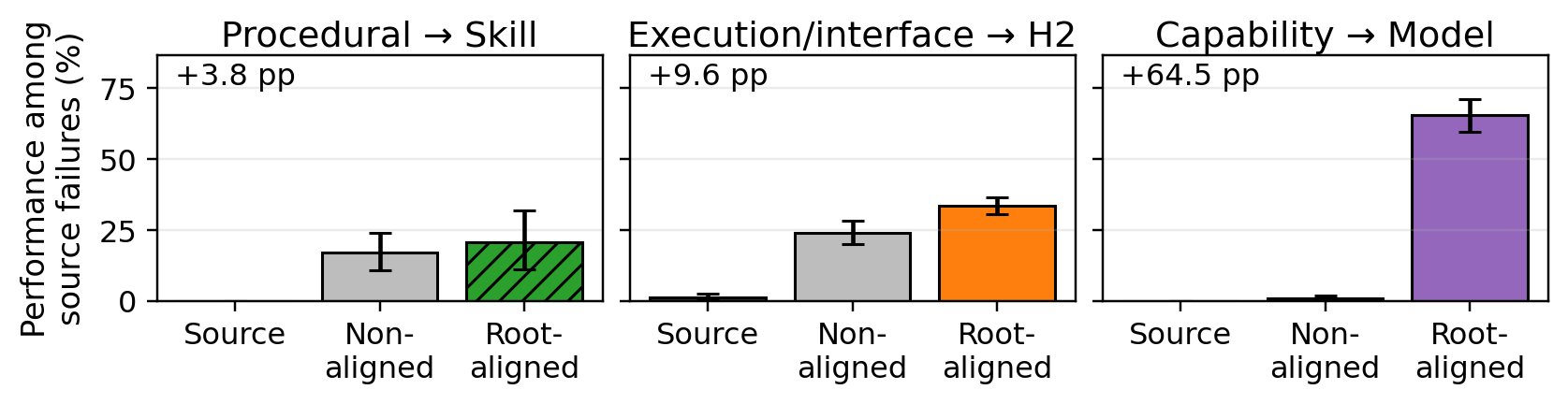}
\caption{Success under the action mapped from each failure type versus the other axes. Hatched: gain over the non-aligned mean not significant.}
\label{fig:aligned}
\end{figure}

\paragraph{The residual failure points to the fix.} We then map each frozen failure label to an axis before loading any outcome (procedural to skill; execution, state, and answer contract to harness; reasoning, factual knowledge, and abstention to model) and replay every failure under the mapped action and under the other axes (Figure~\ref{fig:aligned}). Execution failures gain $+9.58$ points from the harness over the mean of the other axes [$+4.77$, $+13.93$], and capability failures gain $+64.54$ from a stronger model [$+58.68$, $+70.16$]; both survive the estimated label errors. Procedural failures do not map to a skill ($+3.77$ [$-5.66$, $+14.47$]). A stronger model rescues the most in every row, but it replaces the agent rather than repairing it; among fixes that keep the model, the richer harness is the strongest on execution failures (33.50\% against 18.41\% for the best alternative). Most failures stay out of reach: in 59.7\% of comparisons, no single tested action reaches success. Appendix~\ref{app:rq3-details} gives the root composition and the aligned contrasts, and Appendix~\ref{app:rq5-details} the capability boundaries.

\paragraph{Feedback is not recovery.} A matched recovery helps only if the failure travels the whole way from exposure to a correct final answer: the error must be exposed to the model, the model must act on it, the fix must work locally, and the repaired run must end correct. These stages break in different places. For Qwen3.5-9B at H2, BigCode exposes an error in 70--76\% of executions, yet the model repairs at most 2\% of them, because coarse test verdicts do not point to a fix. Spreadsheet repairs about two thirds of the errors it responds to, but only about one fifth of those runs end correct, because a local fix rarely makes the whole workbook right. Across the four models, H2 rescues 27.1\% of the Hotpot FullWiki failures that were visible, actionable, and locally correctable, but only 1.3\% of such failures on Spreadsheet.

\paragraph{Recovery needs a visible signal.} To test this within each benchmark and without using H2 outcomes, we code every Qwen3.5-9B H1 failure by its own trajectory: an interface or contract violation, a runtime or tool error, or no error signal. Failures with a signal are rescued far more often on SQuAD (76.9\% versus 1.3\%), Hotpot Context (63.5\% versus 1.8\%), and Hotpot FullWiki (24.2\% versus 1.5\%), and modestly more often on Spreadsheet (21.5\% versus 17.4\%). BigCode's signals are visible but rarely acted on, and UCSB SOP shows no difference. On SQuAD, 98.7\% of H2 rescues come from interface violations, so recovery there is almost entirely format repair. Appendix~\ref{app:skill-harness-details} gives the recovery-stage and signal tables.

\begin{finding}{Finding 2: The failure points to the fix, but feedback is not recovery}
Missing knowledge calls for a stronger model, and execution errors call for a richer harness. Even a matched recovery repairs a failure only when the error is exposed, the model can act on it, and the repair survives to the final answer; it succeeds mainly when the failure leaves a visible signal.
\end{finding}

\section{Where Self-Evolving Skills Stall}
\label{sec:selfevo}

Self-evolving skills promise to improve an agent from its own experience without changing the model or the harness. The matched design shows four ways in which they fall short, and three experiments test why.

\paragraph{Four shortfalls.}
\begin{itemize}[leftmargin=*,topsep=2pt,itemsep=2pt]
\item \emph{They rarely beat a few human-written lines.} \TraceToSkill{}'s evolved candidate does not beat its human-written initial skill on five of seven benchmarks, and on the test split that initial skill also beats \SkillOpt{} by 10--15 points on three QA benchmarks. \rev{The few human-written lines are also cheap: where the gate kept them, they add a median 4\% tokens, whereas the evolved 2,603-word LiveMath skill adds 195\% (Figure~\ref{fig:quality-cost}).}
\item \emph{They can learn shortcuts.} On LiveMath, where one answer option recurs and is correct whenever present, the evolved skill encodes a rule for picking it: for Qwen3.5-9B it raises accuracy from 11.2\% to 71.5\% on the 53 items with that option and lowers it from 52.7\% to 43.7\% on the other 71.
\item \emph{They overfit to the model they were built on.} This is the opening case. On Hotpot Context, GPT-5.6-sol scores 62.0 at H1 without a skill and 13.6 with the \SkillOpt{} skill optimized on Qwen3.5-9B, which makes it fill its answer field with a fact-extraction worksheet; H2 and H3 both stay at 12.2, because the answer stays well-formed and the outcome-blind revision does not correct it.
\item \emph{They cannot supply missing facts.} On SimpleQA, no skill moves accuracy, whereas a stronger model does.
\end{itemize}

\begin{table}[t]
\centering
\caption{Evidence on the skill bottleneck (locked test). (a) Skills rebuilt with GPT-5.6-sol, evaluated on Qwen3.5-9B; paired differences from the row reference. (b) Same Qwen3.5-9B constructor, different procedure, averaged over Models; \textbf{\textcolor{GainBlue}{bold blue}} marks the row maximum, and a \fbox{box} and a \colorbox{HarmRed!15}{red-shaded cell} mark \NoSkill{} and \Direct{} scores above \SkillOpt{}. (c) Rejected \TraceToSkill{} candidates against their human-written seeds, on Qwen3.5-9B. (d) SimpleQA, Qwen3.5-9B at H1; $\Delta$ against \NoSkill{}.}
\label{tab:skill-bottleneck}
\footnotesize
\noindent\makebox[\columnwidth][l]{\hspace*{-2em}%
\begin{minipage}[t]{0.56\columnwidth}
\centering
(a)\\[2pt]
\setlength{\tabcolsep}{3pt}
\begin{adjustbox}{max width=\linewidth,max totalheight=0.85\textheight}
\begin{tabular}{llrr}
\toprule
Benchmark & Ref. & \SkillOpt{} & \TraceToSkill{} \\
\midrule
\multirow{2}{*}{UCSB SOP} & \NoSkill{} & $+5.38$ & $+13.48$ \\
 & Seed & \textcolor{HarmRed}{$\mathbf{-2.71}$} & \textcolor{GainBlue}{$\mathbf{+5.38}$} \\
\midrule
\multirow{2}{*}{SimpleQA} & \NoSkill{} & $+0.75$ & $+0.61$ \\
 & Seed & $+0.67$ & $+0.09$ \\
\bottomrule
\end{tabular}
\end{adjustbox}
\end{minipage}\hspace{-0.4em}%
\begin{minipage}[t]{0.42\columnwidth}
\centering
(b)\\[2pt]
\setlength{\tabcolsep}{3pt}
\begin{adjustbox}{max width=\linewidth,max totalheight=0.85\textheight}
\begin{tabular}{lrrr}
\toprule
Benchmark & \NoSkill{} & \Direct{} & \SkillOpt{} \\
\midrule
Hotpot Context & \begingroup\setlength{\fboxsep}{0.8pt}\fbox{\strut 50.14}\endgroup & \cellcolor{HarmRed!15}\textbf{\textcolor{GainBlue}{51.77}} & 36.66 \\
SQuAD & 59.47 & \cellcolor{HarmRed!15}\textbf{\textcolor{GainBlue}{71.73}} & 66.33 \\
Spreadsheet & \begingroup\setlength{\fboxsep}{0.8pt}\fbox{\strut \textbf{\textcolor{GainBlue}{34.78}}}\endgroup & \cellcolor{HarmRed!15}33.15 & 29.91 \\
Hotpot FullWiki & \begingroup\setlength{\fboxsep}{0.8pt}\fbox{\strut \textbf{\textcolor{GainBlue}{27.31}}}\endgroup & \cellcolor{HarmRed!15}24.28 & 22.63 \\
\bottomrule
\end{tabular}
\end{adjustbox}
\end{minipage}}

\vspace{8pt}
\begin{minipage}[t]{0.56\columnwidth}
\centering
(c)\\[2pt]
\setlength{\tabcolsep}{3pt}
\begin{adjustbox}{max width=\linewidth,max totalheight=0.85\textheight}
\begin{tabular}{lrrl}
\toprule
Benchmark & Seed & Cand. & Cand. $-$ Seed \\
\midrule
UCSB SOP & 62.00 & 49.19 & $-12.81$ [$-17.76,-8.19$] \\
Amazon SOP & 50.52 & 45.38 & $-5.14$ [$-16.00,+5.24$] \\
Hotpot Context & 63.69 & 61.94 & $-1.75$ [$-3.46,+0.05$] \\
SimpleQA & 6.75 & 6.34 & $-0.41$ [$-2.00,+1.22$] \\
SQuAD (F1) & 83.39 & 85.43 & $+2.04$ [$-0.18,+4.29$] \\
SQuAD (EM) & 72.97 & 67.03 & $-5.94$ [$-9.78,-2.11$] \\
Hotpot FullWiki & 48.95 & 60.57 & $+11.62$ [$+8.11,+15.16$] \\
\bottomrule
\end{tabular}
\end{adjustbox}
\end{minipage}\hfill
\begin{minipage}[t]{0.42\columnwidth}
\centering
(d)\\[2pt]
\setlength{\tabcolsep}{3pt}
\begin{adjustbox}{max width=\linewidth,max totalheight=0.85\textheight}
\begin{tabular}{lrrr}
\toprule
Skill & Acc. & $\Delta$ & Wrong \\
\midrule
\NoSkill{} & 5.0 & -- & 263 \\
Own-data facts & 6.3 & $+1.3$ & 251 \\
Knowledge boundary & 2.0 & $-3.0$ & 45 \\
\quad + search on request & 60.0 & $+55.0$ & 54 \\
\NoSkill{} + search always & 72.0 & $+67.0$ & 34 \\
\bottomrule
\end{tabular}
\end{adjustbox}
\end{minipage}
\end{table}

\paragraph{Why they stall.}
\rev{The four shortfalls share one source. An evolved skill can encode only what the agent's own trajectories and training score reveal. The trajectories contain no fact the model lacks, so the skill cannot supply one. The score rewards whatever raises it on the training split, including a recurring answer option or an answer layout that suits the constructor model, so the skill can learn a shortcut or a model-specific format as readily as a reusable procedure. The three experiments below test the parts of this account that can be separated: the constructor, the procedure, and the missing outside information.}
\begin{itemize}[leftmargin=*,topsep=2pt,itemsep=2pt]
\item \emph{The constructor may be too weak, but a stronger one is not sufficient.} Rebuilding both methods with GPT-5.6-sol (Table~\ref{tab:skill-bottleneck}a) lets \TraceToSkill{} beat the human-written skill on UCSB SOP ($+5.38$ [$+1.71$, $+9.14$]) but makes \SkillOpt{} worse than its Qwen-built version ($-6.05$), and leaves SimpleQA unchanged.
\item \emph{The procedure can be the limit.} With the constructor fixed, the one-shot \Direct{} skill beats the iteratively optimized \SkillOpt{} skill on four of eight benchmarks (Table~\ref{tab:skill-bottleneck}b), and \TraceToSkill{}'s rejected candidates are no better than the human-written seed on five of six (Table~\ref{tab:skill-bottleneck}c).
\item \emph{When knowledge is missing, the agent guesses instead of asking, so asking must be designed in.} On SimpleQA at H1, a skill built from the agent's own training facts moves accuracy from 5.0\% to 6.3\%, and a short human-written skill that asks the model to request facts it does not recall removes most wrong guesses but cannot raise accuracy on its own. When each request is answered with passages from the cited sources, the same model reaches 60.0\%, and 72.0\% when the passages come with every question (Table~\ref{tab:skill-bottleneck}d).
\end{itemize}
Appendix~\ref{app:skill-experiments} gives the design and full results of the four experiments.

\begin{finding}{Finding 3: Self-evolution needs something from outside the agent}
Skills evolved from the agent's own experience rarely beat a short human-written skill, can learn dataset shortcuts, break when moved to another model, and cannot supply missing facts. A stronger constructor does not fix this; guidance or information from outside the agent's experience does.
\end{finding}

\section{\texorpdfstring{\rev{How to Choose an Intervention}}{How to Choose an Intervention}}
\label{sec:choose}

\paragraph{Two policies that did not pay.} If the failure points to the fix, a controller could choose the intervention per task. A router that chooses an action from the task text, or from a predicted failure type, does not beat the fixed \TraceToSkill{}+H2 configuration (39.42\% and 40.10\% against 45.29\%), although an oracle that chooses on one repeat and is scored on the others shows that headroom exists (49.24\%). Behavior before recovery does predict which Qwen3.5-9B H1 failures H2 will rescue (AUROC 0.945, against 0.796 for benchmark identity), but escalating only H1 failures reaches 42.1\% against 41.6\% for always using H2: it cuts regressions from 2.20 to 0.87 per 100 instances yet costs 68\% more tokens, because H1 must run first. The failure state is informative, but a policy that turns it into accuracy or cost gains remains open. Appendices~\ref{app:pre-recovery} and~\ref{app:routing} give the methods and offline replays.

\paragraph{\rev{A policy that does pay.}} \rev{The failure signal is too sparse to choose per task online. Escalating Qwen3.5-9B+H2 to a stronger model only when its trajectory ends in a visible error, an abstention, or an over-long answer adds about one point, because 89\% of its failures at H2 are silent: the signalled ones were already retried inside H2 (Appendix~\ref{app:online-policies}, Table~\ref{tab:online-cascade}). Choosing per benchmark from a small labeled pilot works. Running Qwen3.5-9B+H2 and a stronger model on 20 labeled tasks, and switching the whole benchmark only if the stronger model scores higher on them, reaches 57.4 against 53.5 for always using GPT-5.6-sol, at 36\% lower cost, because the stronger model is worse on some benchmark--skill combinations (on SQuAD, for example; Appendix~\ref{app:online-policies}, Table~\ref{tab:online-pilot}). The same pilot gains nothing over always using H2 rather than H1, and matches always using H3 at 10\% lower cost. Configuration dependence, which makes an average gain misleading, is what a benchmark-level pilot exploits (Appendix~\ref{app:online-policies}).}

\begin{finding}{\rev{Finding 4: Choose the intervention per benchmark, not per task}}
\rev{Per-task signals are too sparse to choose an intervention online: most residual failures at H2 are silent. A small labeled pilot per benchmark works: 20 labeled tasks are enough to keep the stronger model only where it helps, scoring higher than always switching at 36--42\% lower cost.}
\end{finding}

\section{Discussion}
\label{sec:discussion}

\begin{revblock}
\paragraph{Implications for building skills.} Four practices follow for anyone who writes or evolves skills.
\begin{itemize}[leftmargin=*,topsep=2pt,itemsep=2pt]
\item \emph{Put the answer contract in the harness, not in the skill.} A skill that mainly enforces the answer contract is absorbed once the harness enforces it (SQuAD: $+66.2$ at H0, $+2.2$ at H2), and a skill that rewrites the format can break the answer (the worksheet that drops GPT-5.6-sol from 62.0 to 13.6). Formats, schemas, and tool-call rules that code can check belong in the harness; the skill can then carry what the harness cannot see, such as a precondition that fails silently.
\item \emph{Treat a skill as validated for one model and one harness.} A skill's gain did not carry over unchanged to another harness (Table~\ref{tab:regimes}) or to a stronger model (the opening case), so a skill kept across a model or harness upgrade should be re-tested against no skill, and removed if it no longer helps.
\item \emph{Judge an evolved skill by its gain per token, against a short human-written baseline.} The human-written lines the gate kept add a median 4\% tokens; the evolved LiveMath skill adds 195\% and partly encodes a shortcut. An evolved skill is worth its cost only if it beats the short baseline on the target model and harness.
\item \emph{Do not take an offline optimization gain at face value.} A high training score from automatic prompt or skill optimization may come from a shortcut rather than a reusable procedure: on LiveMath, the evolved skill learns to pick a recurring option, raising accuracy from 11.2\% to 71.5\% where that option is correct but lowering it from 52.7\% to 43.7\% elsewhere. Validate an evolved skill on out-of-distribution test sets or across tasks, and track its tokens, so that a spurious gain of a few points does not cost twice the compute (the LiveMath skill adds $+195\%$ tokens).
\end{itemize}
\end{revblock}

\begin{figure}[t]
\centering
\includegraphics[width=0.82\linewidth,height=0.85\textheight,keepaspectratio]{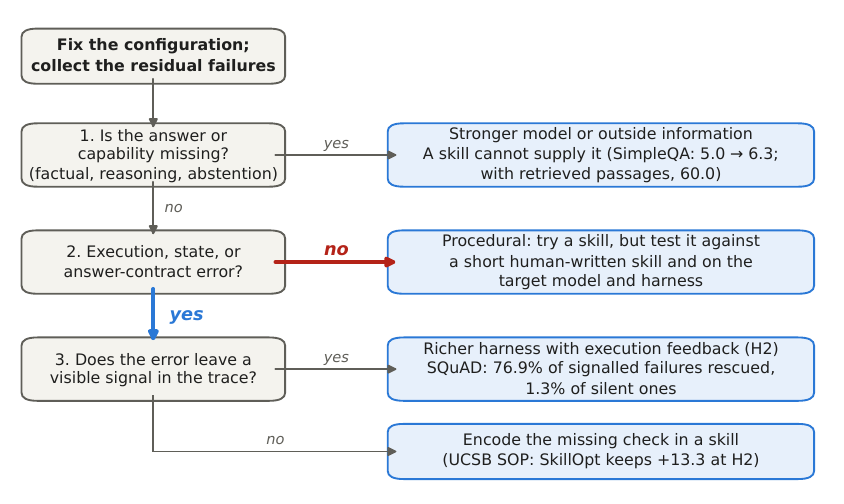}
\caption{\rev{A diagnostic flow for choosing an intervention from a system's residual failures. Each branch cites the result it rests on; procedural failures do not map reliably to a skill ($+3.77$ [$-5.66$, $+14.47$]), so a candidate skill must be tested against a short human-written one.}}
\label{fig:diagnostic}
\end{figure}

\paragraph{A diagnostic flow.} Figure~\ref{fig:diagnostic} turns the findings into an order of questions for a system designer to ask before choosing an intervention; it is not a per-task controller.

\begin{revblock}
\paragraph{Limits.} \emph{Skill construction.} Main-matrix skills are built by Qwen3.5-9B and reused unchanged, so the cross-model results, including the opening collapse, test transfer rather than construction by each model; the stronger constructor is tested on two benchmarks only. \emph{Model scope.} The skill--harness regimes, the signal coding, and the pre-recovery prediction use Qwen3.5-9B, and rescue appears only for it. \emph{Feedback versus retry.} H2 bundles error feedback with up to three retries, so their contributions are not separated. \emph{Failure labels.} Only 56.5\% of source failures are labeled, the labels are validated at the level of the intervention axis, and LiveMath is excluded. \emph{Tested space.} ``Out of reach'' means beyond every tested action, at temperature 0 and with budgets swept only at H1 and H2.
\end{revblock}

\section{Related Work}
\label{sec:related}

\paragraph{Evaluating agents.} Agent benchmarks compare model backbones under fixed scaffolds \citep{liu2024agentbench,ma2024agentboard,yao2024tau}, and HAL evaluates models, scaffolds, and benchmarks jointly \citep{kapoor2026hal}. \rev{Others fix the task and vary the agent's visible context \citep{weng2026temporalbench} or simulated user \citep{li2026personacuabench}, or check tool actions against executable task contracts \citep{banerjee2026crmagentbench}. We vary one intervention at a time on the same task instances and repeats, so each effect is paired and attributed to one axis.}

\paragraph{\rev{Skills and harnesses.}} SkillsBench compares no-skill, curated, and self-generated skills across model--harness configurations \citep{li2026skillsbench}; SkillLearnBench scores skill generation on the skill, the trajectory, and the outcome \citep{zhong2026skilllearnbench}; \citet{shaposhnikov2026framework} evaluate about 500 skills across 19 agent--model configurations; and \citet{ding2026agentskill} survey skill evolution and evaluation. \rev{Other work builds skill libraries \citep{sha2026skillcenter}, routes among skills \citep{wei2026dsr}, and learns the harness from experience \citep{huang2026memoharness}; added components can also over-privilege a skill \citep{li2026fortisbenchmarkingoverprivilegeagent} or break agent behavior \citep{li2026defensespromptattackslearn,li2026autonomytaxdefensetraining}.} We evaluate self-evolving methods rather than only curated or one-shot skills, and link each effect\rev{, gain or loss,} to the failure it acts on.

\paragraph{\rev{Feedback, recovery, and asking.}} Intrinsic self-correction often fails without external feedback \citep{huang2024cannot}, works best with reliable feedback \citep{kamoi2024when}, and depends on feedback quality in code \citep{olausson2024selfrepair}\rev{ and in time-series agents \citep{ye2026tsreasoner}}. \rev{Agents can also learn when to ask an expert \citep{yang2026adaptive} or a patient \citep{gao2026medexagent}.} We add stage-wise recovery evidence across benchmarks\rev{, and show that when knowledge is missing, evolving skills from the agent's own experience cannot repair the failure: the agent guesses, so the missing information must come from outside that experience}.

\paragraph{\rev{Allocation and failure diagnosis.}} LLM routers choose which model answers a query \citep{chen2023frugalgpt,ong2025routellm,hu2024routerbench}, \rev{or reweight a model ensemble over several turns \citep{cao2025tsorchestra},} and compute-optimal scaling allocates test-time computation by difficulty \citep{snell2025scaling}; we instead fix the model and choose which intervention to apply. \rev{Prior work diagnoses} why agent systems fail \citep{cemri2026multi}\rev{, when a failure can be caught \citep{zhao2026catchbench}, and whether a change to an agent is safe to deploy \citep{zhao2026agentrisk}}; we test whether such diagnoses select the fix\rev{, and compare per-task with per-benchmark choice: switching models per benchmark from a small labeled pilot beats always using the stronger model, at lower cost}.

\section{\texorpdfstring{\rev{Conclusion}}{Conclusion}}
\label{sec:conclusion}

\rev{We asked when an agent intervention helps and which failure it repairs, using a matched factorial study that changes one axis at a time over nine benchmarks and four models. The answer is conditional but not arbitrary. No configuration is best everywhere, and interventions interact rather than add. The failure left by the current system points to the axis with room to help, and feedback repairs mainly failures that leave a visible signal. Skills evolved from the agent's own experience stall unless guidance or information comes from outside that experience.} \rev{Finally, the choice of intervention pays off per benchmark rather than per task: a small labeled pilot that keeps the stronger model only where it helps beats always using that model, at lower cost.}

\paragraph{How to judge an intervention.} An intervention should therefore be judged not by its average gain, but by which failure it repairs and under which configuration. Self-evolving methods should be compared against a short human-written skill and tested beyond the model they were built on, and an agent that learns from its own experience should also learn when that experience is insufficient and where to look \citep{ren2023robots,wang2024mint,schick2023toolformer,asai2024selfrag}.

\clearpage
\bibliographystyle{plainnat}
\bibliography{references}

\appendix

\clearpage
\section{Experimental Details and Measurement Definitions}
\label{app:measurement-details}

\paragraph{Design Matrix.}
We organize the matched study as $T\times M\times S\times H\times B$, use three repeats per available cell, and hold task instances, frozen Skill artifacts, and evaluators fixed. Figure~\ref{fig:design} summarizes the matrix; the remainder of this section reports the intervention definitions, task-property annotations, execution ledger, protocol details, and uncertainty calibration.

\subsection{Intervention Axes}
\label{app:intervention-axes}

\subsubsection{Model (\texorpdfstring{$M$}{M})}
\label{app:model-axis}

\begin{table}[!htbp]
\centering
\caption{Model axis. Qwen3.5-9B also constructs every \Direct{}, \SkillOpt{}, and \TraceToSkill{} artifact, so the API models test whether a frozen skill transfers to stronger targets.}
\label{tab:model-axis}
\small
\begin{adjustbox}{max width=\linewidth,max totalheight=0.85\textheight}
\begin{tabular}{lll}
\toprule
Model & Access & Role \\
\midrule
Qwen3.5-9B & Local server & Target; skill constructor; root-label judge \\
Claude Sonnet 4.6 & API & Target \\
Claude Sonnet 5 & API & Target \\
GPT-5.6-sol & API & Target \\
\bottomrule
\end{tabular}
\end{adjustbox}
\end{table}

\subsubsection{Skill (\texorpdfstring{$S$}{S})}
\label{app:skill-axis}

\begin{figure*}[t]
  \centering
  \includegraphics[width=\textwidth,height=0.85\textheight,keepaspectratio]{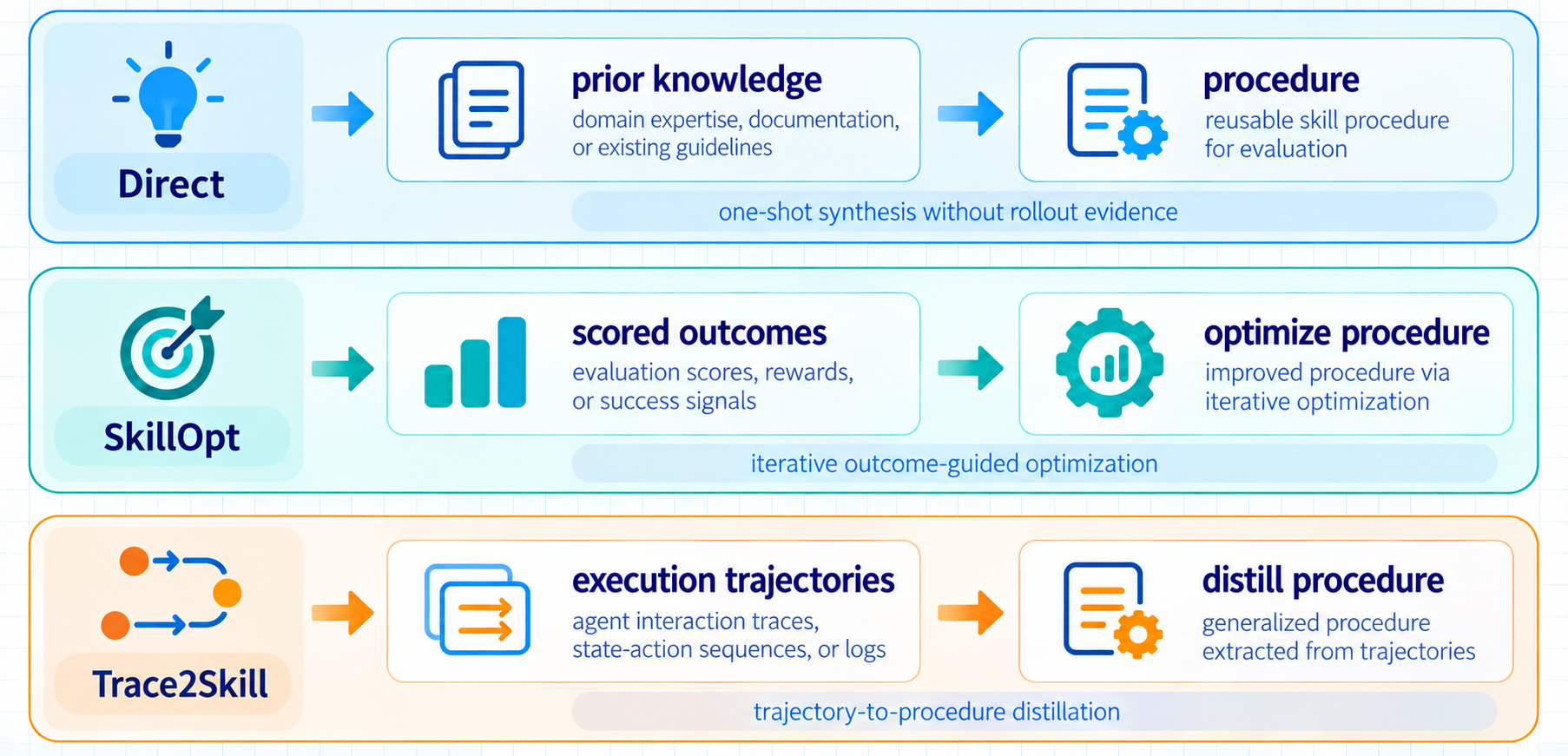}
  \caption{Skill construction pathways. \Direct{} synthesizes a procedure from prior knowledge without rollout evidence; \SkillOpt{} optimizes a procedure using scored outcomes; and \TraceToSkill{} distills a reusable procedure from execution trajectories.}
  \label{fig:skill-construction-pathways}
\end{figure*}

Table~\ref{tab:skill-conditions} states what each condition isolates, and Figure~\ref{fig:skill-construction-pathways} shows the construction pathways. Tables~\ref{tab:skill-construction}--\ref{tab:trace2skill-gate} give the construction settings, the \Direct{} inputs and prompt, and the \TraceToSkill{} validation-gate outcomes.

\begin{table*}[!t]
\centering
\scriptsize
\caption{Skill construction conditions. The Outcome feedback column indicates whether scored outcomes directly drive optimization; Trace2Skill distills execution experience rather than primarily optimizing a procedure against validation scores. All constructed Skill artifacts are frozen before downstream evaluation. All constructed artifacts are built by Qwen3.5-9B on the training split (Table~\ref{tab:skill-construction}). All frozen artifacts are transferred unchanged across target Models.}
\label{tab:skill-conditions}
\setlength{\tabcolsep}{3pt}
\begin{tabularx}{\textwidth}{p{0.12\textwidth}>{\centering\arraybackslash}p{0.10\textwidth}>{\centering\arraybackslash}p{0.18\textwidth}p{0.18\textwidth}X}
\toprule
Condition & Rollout evidence & Outcome feedback & Construction & What it isolates \\
\midrule
\textbf{\NoSkill{}} & No & No & None & Base model + harness capability \\
\textbf{\Direct{}} & No & No & One-shot synthesis & Value of prior-only procedural guidance \\
\textbf{\SkillOpt{}} & Yes & Yes & Iterative optimization & Value of outcome-guided skill learning \\
\textbf{\TraceToSkill{}} & Yes & Not primarily optimization-driven & Trajectory distillation & Value of experience-derived reusable procedures \\
\bottomrule
\end{tabularx}

\end{table*}

\begin{table*}[!t]
\centering
\scriptsize
\caption{Skill construction settings. All constructed skills are frozen before evaluation and transferred unchanged to every Model.}
\label{tab:skill-construction}
\setlength{\tabcolsep}{3pt}
\begin{tabularx}{\textwidth}{p{0.11\textwidth}p{0.27\textwidth}p{0.24\textwidth}X}
\toprule
Condition & Constructor & Construction input & Selection \\
\midrule
\Direct{} & Qwen3.5-9B, one call (temperature 0, thinking disabled, seed 42) & Frozen catalog of training items only (Table~\ref{tab:direct-catalog}) & None: no rollouts, feedback, or validation gate \\
\SkillOpt{} & Qwen3.5-9B as rollout model and optimizer & Scored training rollouts & Step patches gated on method validation \\
\TraceToSkill{} & Qwen3.5-9B as rollout model, error/success analyst, and merge model & Successful and failed training trajectories & Kept only if it strictly beats its human-written initial skill on method validation (Table~\ref{tab:trace2skill-gate}) \\
\bottomrule
\end{tabularx}
\end{table*}

\begin{table}[!htbp]
\centering
\scriptsize
\caption{\Direct{} construction input. No validation, Harness-validation, or test item enters any catalog, and no gold workbook enters the Spreadsheet catalog.}
\label{tab:direct-catalog}
\begin{tabularx}{\columnwidth}{p{0.30\columnwidth}X}
\toprule
Benchmark & Catalog of training items \\
\midrule
SimpleQA, SQuAD, Hotpot & Question, context or evidence where applicable, and reference answer \\
Amazon SOP & Training items, with the official SOP text appended to the prompt \\
Spreadsheet & Instruction and task type of the 80 training items \\
LiveMath, UCSB SOP & Training items, with a benchmark-specific variant of the prompt \\
\bottomrule
\end{tabularx}
\end{table}

The shared \Direct{} prompt reads: ``Create one concise, reusable domain skill in Markdown from the frozen training catalog below. Do not solve or enumerate individual examples. Synthesize general procedures, checks, output discipline, and common failure avoidance. This is a single Direct-Synthesis call: no rollout, execution feedback, validation gate, retrieval, or later update. Return only the Markdown skill.'' Spreadsheet, LiveMath, and UCSB SOP use benchmark-specific variants of the same instruction; all prompts are included in the released code.

\begin{table}[!htbp]
\centering
\small
\caption{\TraceToSkill{} validation gate. Scores are on the small method-validation split (0--1). The evolved candidate is kept only if it strictly beats its initial skill, a short human-written skill of 17--85 words; otherwise the frozen artifact is that initial skill. The Hotpot gate uses the FullWiki view.}
\label{tab:trace2skill-gate}
\begin{adjustbox}{max width=\linewidth,max totalheight=0.85\textheight}
\begin{tabular}{lrrl}
\toprule
Benchmark & Initial skill & Evolved candidate & Frozen artifact \\
\midrule
UCSB SOP & 0.40 & 0.35 & Initial skill \\
Hotpot & 0.55 & 0.50 & Initial skill \\
SQuAD & 0.80 & 0.75 & Initial skill \\
Amazon SOP & 0.00 & 0.00 & Initial skill \\
SimpleQA & 0.10 & 0.10 & Initial skill \\
Spreadsheet & -- & -- & Evolved \\
LiveMath & 0.30 & 0.80 & Evolved \\
\bottomrule
\end{tabular}
\end{adjustbox}
\end{table}

Spreadsheet has no method-validation scores because it follows the original \TraceToSkill{} design, which uses the evolved skill directly.

\FloatBarrier

\subsubsection{Harness (\texorpdfstring{$H$}{H})}
\label{app:harness-axis}

A Harness changes the information and actions available during execution rather than supplying a reusable task policy. Table~\ref{tab:ladder} defines the four levels; Table~\ref{tab:harness-per-benchmark} gives their implementation on every benchmark. H1 targets interface and answer-contract compatibility, whereas H2 jointly changes failure observability and the opportunity to act on it, so an H2 gain cannot be attributed separately to error exposure or retry. H3 tests whether one additional outcome-blind revision improves a completed response or trajectory.

\begin{table*}[!t]
\centering
\scriptsize
\caption{Harness implementation per benchmark. No level exposes evaluator outcomes or gold answers; a privileged diagnostic branch that does is excluded from every analysis. Ceilings are those of the Budget axis (Table~\ref{tab:budget-single-turn}).}
\label{tab:harness-per-benchmark}
\setlength{\tabcolsep}{3pt}
\begin{tabularx}{\textwidth}{p{0.12\textwidth}XXXXp{0.13\textwidth}}
\toprule
Benchmark & H0 & H1 & H2 & H3 & Ceilings that bind \\
\midrule
SQuAD, Hotpot Context, SimpleQA & Benchmark instruction and question (plus passages, if any), skill appended; no answer format, so the whole response is scored & Fixed skill package block; \texttt{<answer>} contract: shortest span (SQuAD, Hotpot) or a concise answer with an explicit ``I don't know'' option (SimpleQA) & Fixed re-ask, only when no valid \texttt{<answer>} field is found; at most three re-asks; no correctness signal & One fixed outcome-blind reflection after H2; the revision replaces the answer only if it parses & Tokens per call; turns cap only the re-asks; per-request timeout \\
LiveMath & System prompt already requires one choice label in \texttt{<answer>} at every level; skill appended & Same prompt; skill in the fixed package block & Re-ask when the parsed label is not a valid choice (at most three) & One outcome-blind reflection, if turns remain & Tokens; turns cap re-asks \\
Spreadsheet & One code generation from the raw skill package & Same, with the standardized package & Generated code is executed and the execution error is fed back (up to three turns by default) & One reflection over the code plus an output preview or the execution error & Tokens, turns, timeout \\
UCSB SOP & Official tool environment, run once, raw package & Standardized package text & Native tool errors returned, with up to three recovery attempts & One trajectory reflection, then a replay from reset & Tokens, turns, timeout \\
Amazon SOP & Strict action loop; the first invalid action ends the episode & Standardized package & Rejected or failed actions fed back, with up to three recoveries & Reflection, then a replay from reset & Tokens, turns, timeout \\
Hotpot FullWiki & Retrieval-tool protocol and \texttt{<answer>} required at every level; tool errors appear only as ``Tool action failed.'' & Standardized package & The actual tool exception text (e.g., unknown title), plus up to three re-asks for a missing final answer & One reflection & Tokens, turns, timeout; tool-call and total-call caps bind at Low \\
\bottomrule
\end{tabularx}
\end{table*}

On single-turn QA, H1 adds the answer interface, H2 re-asks only when that interface is not satisfied, and H3 always adds one revision pass that can change the answer. On SQuAD, for example, the H2 message is: ``The answer interface was empty or invalid. Return one concise context-supported span inside \texttt{<answer>...</answer>}.''

\FloatBarrier

\subsubsection{Budget (\texorpdfstring{$B$}{B})}
\label{app:budget-axis}

Low, Medium, and High ceilings are crossed only with H1 and H2; H0--H3 are also evaluated under each benchmark's historical Default configuration, which is not relabeled Medium. The sweep jointly raises completion-token, turn, and timeout ceilings (Table~\ref{tab:budget-limits}), so its effects cannot be attributed to tokens alone; these are resource allowances, not observed consumption or reasoning-effort settings. Historical UCSB Low restricts workflow feasibility; the separately recorded protocol-preserving corrected Low reduces only the token cap and retains its reference turns, tools, and timeout, so the two endpoints are not interchangeable.

\begin{table}[!htbp]
\centering
\caption{Nominal resource ceilings in the Low/Medium/High sweep.}
\label{tab:budget-limits}
\small
\begin{adjustbox}{max width=\linewidth,max totalheight=0.85\textheight}
\begin{tabular}{lrrr}
\toprule
Budget & Completion tokens & Turns & Timeout (s) \\
\midrule
Low & 4,096 & 3 & 120 \\
Medium & 16,384 & 30 & 360 \\
High & 32,768 & 60 & 720 \\
\bottomrule
\end{tabular}
\end{adjustbox}
\end{table}

\begin{table}[!htbp]
\centering
\small
\caption{How the Budget ceilings bind on single-turn QA (SQuAD, Hotpot Context, SimpleQA). This matches their near-zero Budget envelopes.}
\label{tab:budget-single-turn}
\begin{tabularx}{\columnwidth}{lX}
\toprule
Ceiling & Effect on single-turn QA \\
\midrule
Completion tokens & Applies to every call \\
Turns & Caps only the number of H2 re-asks (two at Low, three otherwise) \\
Timeout & Per-request client timeout \\
\bottomrule
\end{tabularx}
\end{table}

\paragraph{Matched intervention and root-cause protocol.}
For each intervention, we compare matched task-instance--model configurations while changing one axis and holding the rest of the execution condition fixed. For the failure analysis, we assign 2,225 failed source executions to eight root-cause categories using the earliest decisive cause supported by the trajectory; Table~\ref{tab:appendix-root-taxonomy} defines those categories. The source label is frozen before matched target outcomes are inspected. Benchmark-specific deterministic checks support the label when possible; cases requiring additional interpretation use an AgentRx-inspired constraint-grounded judge.

Two properties of this labeling limit what it can support. The judge is Qwen3.5-9B, which is also one of the four evaluated models and the model whose training executions produced every \SkillOpt{} and \TraceToSkill{} artifact, so on one quarter of the matrix it attributes causes to its own failures. Only the 305 LiveMath labels come from this judge (Appendix~\ref{app:root-labeling}), and an independent judge does not reproduce them, so LiveMath is excluded from every root-conditioned analysis; Appendix~\ref{app:label-validation} reports the independent-judge and author validations of the remaining labels. We therefore treat model-assisted labels as hypotheses throughout and mark them as such in every table; Section~\ref{sec:root-cause-appendix} gives the definitions and the evidence boundary.

\paragraph{Measurement validity.}
An unchanged-configuration repeat comparison (A/A) compares repeated executions of the same task instance to measure variation without an intervention. Noise-qualified improvement must exceed the corresponding A/A tolerance, whereas success requires reaching the benchmark's native success threshold. Section~\ref{app:validity} distinguishes these criteria and documents label coverage, budget limits, and the calibration status of the reported descriptive results.

\subsection{Per-Benchmark Capabilities}
\label{app:benchmark-capabilities}

\begin{table}[!htbp]
  \centering
  \caption{Human-authored descriptive task properties and task-structure intervention hypotheses used to diversify benchmark selection. These annotations are not trajectory-derived root labels and are not used as evidence in the root-cause analysis. Proc.\ = procedural; Fb.\ = feedback; Med.\ = medium. ``Expected intervention'' records a design hypothesis, not an observed winner.}
  \label{tab:appendix-copy-task-properties}
  \scriptsize
  \setlength{\tabcolsep}{3pt}
  \begin{tabularx}{\textwidth}{lcccccXX}
    \toprule
    Benchmark & Proc. & Tool/ & Horizon & Fb. & Fb. & Expected bottleneck & Expected intervention \\
     & demand & state & & demand & avail. & & \\
    \midrule
    Spreadsheet     & Med. & High & Med. & High & High & Execution/tool & H2; H3 \\
    LiveMath        & Low  & Low  & Med. & Low  & Low  & Reasoning/decision rule & Skill; Model \\
    UCSB SOP        & High & High & Med. & High & High & Procedural & Skill; H2 \\
    Amazon SOP      & High & High & High & High & High & Execution/tool; horizon & H2; H3; Budget \\
    SimpleQA        & Low  & Low  & Low  & Low  & Low  & Factual knowledge/recall & Model \\
    SQuAD           & Low  & Low  & Low  & Low  & High & Evaluation/answer contract & H1 \\
    Hotpot Context  & Low  & Low  & Med. & Low  & Med. & Evaluation/answer contract & H1 \\
    Hotpot FullWiki & Med. & High & High & High & Med. & Horizon/termination & Skill; H2; Budget \\
    \bottomrule
  \end{tabularx}
  
\end{table}

Table~\ref{tab:appendix-copy-task-properties} reproduces the compact benchmark-property summary here for local reference. These columns are human-authored design annotations used to diversify benchmark selection before any execution; they are not trajectory-derived root labels and enter no root-cause analysis, so the \emph{Expected bottleneck} and \emph{Expected intervention} entries can be read against the measured results without circularity. Table~\ref{tab:benchmark-capabilities} lists the primary capability each benchmark exercises.

\begin{table}[!htbp]
\centering
\scriptsize
\caption{Primary capability exercised by each benchmark.}
\label{tab:benchmark-capabilities}
\begin{tabularx}{\columnwidth}{p{0.30\columnwidth}X}
\toprule
Benchmark & Primary capability \\
\midrule
SpreadsheetBench \citep{ma2024spreadsheetbench} & multi-step, state-changing spreadsheet edits issued through a tool; stresses execution and tool-use reliability under an evolving workbook state \\
LiveMathematicianBench \citep{he2026livemathematicianbench} & recent multiple-choice mathematical reasoning; stresses reasoning together with a reusable answer-selection regularity, namely a recurring meta-option that is systematically correct when present \\
UCSB SOPBench \citep{li2025sopbench} & long, constraint-sensitive standard operating procedures executed against state; stresses procedural compliance, step ordering, and constraint verification \\
Amazon SOP \citep{nandi2026sop} & a long, state-changing inspection workflow with many tool calls; stresses execution reliability and path feasibility over a long horizon \\
SimpleQA \citep{wei2024measuring} & short-form factual questions with no supplied evidence; stresses parametric factual recall and calibrated abstention \\
SQuAD \citep{rajpurkar2016squad} & extractive question answering with the evidence passage supplied; stresses answer-contract and interface compliance rather than retrieval \\
HotpotQA, provided-context \citep{yang2018hotpotqa} & multi-hop questions with gold and distractor paragraphs supplied; stresses multi-hop evidence integration under a fixed answer contract \\
HotpotQA, full-wiki \citep{yang2018hotpotqa} & the same questions with only a frozen Wikipedia index; adds retrieval-policy, grounding, and horizon demands on top of evidence integration \\
BigCodeBench-Hard \citep{zhuo2025bigcodebench} & program synthesis executed against held-out tests; stresses code generation with executable feedback.; appendix-only, since its evaluator environment is not fully comparable across Models \\
\bottomrule
\end{tabularx}
\end{table}

\subsection{Benchmark Samples and Frozen Splits}
\label{app:benchmark-splits}

Table~\ref{tab:benchmark-splits} reports every sample used for Skill construction, method selection, Harness qualification, and final evaluation. Method validation selects or accepts Skill artifacts; Harness validation qualifies execution code and interfaces without selecting Skill content; the locked test split is used only for final evaluation. We provide machine-readable \texttt{split\_manifest.json} files and ordered \texttt{ids.json} files for every role, including the split seed, source revision, and SHA-256 hashes. Where redistribution is permitted, the archive also provides the corresponding \texttt{items.json}. These files reproduce the exact task-instance membership and ordering used in the experiments and permit direct overlap checks across roles.

\begin{table*}[!t]
\centering
\caption{Benchmark sample and split accounting. \texttt{Used} is the number of distinct task instances assigned to the four experimental roles, before multiplying by Models, Skills, Harnesses, Budgets, or repeats. Hotpot Context and FullWiki are two execution settings over the same UID split and are therefore listed separately but do not represent disjoint questions. All project-defined sampling uses seed 42. For SQuAD and HotpotQA, the two source-pool numbers denote the official training and development/validation pools. Amazon has 12 additional audit-reserve examples that are excluded from the four experimental roles. BigCodeBench is appendix-only because its evaluator environment is not fully comparable across Models.}
\label{tab:benchmark-splits}
\scriptsize
\setlength{\tabcolsep}{3.5pt}
\begin{adjustbox}{max width=\linewidth,max totalheight=0.85\textheight}
\begin{tabular}{lrrrrrr}
\toprule
Benchmark & Source pool & Train & Method val. & Harness val. & Test & Used \\
\midrule
SpreadsheetBench & 400 & 80 & 24 & 16 & 280 & 400 \\
LiveMathematicianBench & 177 & 35 & 10 & 8 & 124 & 177 \\
UCSB SOP & 830 & 70 & 20 & 15 & 175 & 280 \\
Amazon SOP & 112 & 30 & 10 & 10 & 50 & 100 \\
SimpleQA & 4,326 & 70 & 20 & 15 & 300 & 405 \\
SQuAD & 87,599+10,570 & 70 & 20 & 15 & 300 & 405 \\
Hotpot Context & 90,447+7,405 & 70 & 20 & 15 & 300 & 405 \\
Hotpot FullWiki & 90,447+7,405 & 70 & 20 & 15 & 300 & 405 \\
BigCodeBench-Hard & 148 & 70 & 20 & 15 & 43 & 148 \\
\bottomrule
\end{tabular}
\end{adjustbox}

\end{table*}

\paragraph{SpreadsheetBench.}
We use the verified 400-instance release at the frozen source revision. Its 80/40/280 source train/validation/test partition is preserved, with the 40 validation instances divided into 24 method-validation and 16 Harness-validation instances. The 280-instance test set remains locked. Spreadsheet skill provenance is described in Section~\ref{app:skill-provenance}.

\paragraph{LiveMathematicianBench.}
We deterministically partition the 177 questions from four monthly releases with a 2:1:7 train/validation/test ratio and seed 42, yielding 35/18/124 task instances. The 18 validation questions are divided in source order into ten method-validation and eight Harness-validation instances.

\paragraph{UCSB SOP.}
From the 830 released SOPBench instances, we freeze 280 examples using deterministic balancing over domain, path length, constraint complexity, tool-call demand, outcome type, and SOP branch. The split contains 70 training, 35 validation, and 175 locked test examples, with 25 test examples from each of seven domains. The validation set is partitioned into 20 method-validation and 15 Harness-validation examples.

\paragraph{Amazon SOP.}
The aircraft-inspection configuration contains 112 released examples. Because the official release is test-only, we define project roles with seed 42: 30 training, ten method-validation, ten Harness-validation, and 50 test examples. The remaining 12 form an audit reserve and are not included in the factorial evaluation.

\paragraph{SimpleQA.}
The official release contains 4,326 test-only questions. We use seed 42 and stratify by topic and answer type to select 405 disjoint examples: 70 training, 20 method-validation, 15 Harness-validation, and 300 locked test questions. The remaining 3,921 questions are not used.

\paragraph{SQuAD.}
Training and validation roles are sampled from the 87,599-question official training set, using one question per development article and stratification by question type and answer-length bucket. The 70/20/15 training, method-validation, and Harness-validation examples are article-disjoint. The 300 locked test questions come from the 10,570-question official development set and cover all 48 development articles.

\paragraph{Hotpot Context.}
We sample 405 question UIDs with seed 42 from the official distractor configuration: 70 training, 20 method-validation, 15 Harness-validation, and 300 test questions. This setting supplies the official gold and distractor paragraphs to the agent.

\paragraph{Hotpot FullWiki.}
FullWiki reuses exactly the same 405 UIDs and role assignments as Hotpot Context, but supplies only the question and permits retrieval from a frozen local 2017 Wikipedia index; gold titles are not used to construct that index. Exact UID pairing makes differences between the two settings attributable to evidence access rather than question sampling.

\paragraph{BigCodeBench.}
We use all 148 task instances in BigCodeBench-Hard v0.1.4 and define disjoint roles of 70/20/15/43 with seed 42. Agent-visible prompts and evaluator-private solutions/tests are stored separately. BigCodeBench is retained as appendix evidence because its evaluator environment is not fully comparable across Models.

\paragraph{Evaluation and aggregation.}
Every benchmark retains its native evaluator and success semantics. Native scores are never pooled directly across benchmarks; cross-benchmark summaries use equal-weight macro averaging so that large test sets do not dominate. Exact evaluator versions and instance UIDs are tied to the released split manifests and execution ledger.

\subsection{Execution ledger}
\label{app:compute-ledger}

Table~\ref{tab:compute-ledger} states the nominal size of the full nine-benchmark study. The eight primary comparable benchmarks contribute 1,829 test instances, and appendix-only BigCode contributes 43, for 1,872 test instances in total. A small number of cells are unavailable rather than executed, most notably the protocol-preserving Low endpoints for the three API models on UCSB SOP, so the realized total is slightly below the nominal one.

\begin{table}[!htbp]
\centering
\caption{Nominal execution ledger over all nine benchmarks (1,872 test instances). BigCode is retained as appendix-only evidence; derived main-analysis cohorts use the eight comparable benchmarks. $^{a}$Four Models $\times$ four Skills $\times$ H0--H3 $\times$ three repeats. $^{b}$Four Models $\times$ four Skills $\times$ H1/H2 $\times$ three Budgets $\times$ three repeats. The same formulas include the 43 BigCode test instances in the nominal nine-benchmark count; BigCode remains excluded from primary comparative analyses because its evaluator environment is not fully comparable across models. Qwen3.5-9B was served locally, so its executions have no per-token API price and no comparable latency or dollar ledger. Wall-clock and accelerator hours were not logged uniformly across the campaign and we do not report them.}
\label{tab:compute-ledger}
\small
\begin{adjustbox}{max width=\linewidth,max totalheight=0.85\textheight}
\begin{tabular}{lr}
\toprule
Quantity & Count \\
\midrule
Primary comparable test instances & 1,829 \\
BigCode appendix test instances & 43 \\
All test instances & 1,872 \\
Default matrix executions$^{a}$ & 359,424 \\
Budget sweep executions$^{b}$ & 539,136 \\
Total nominal agent executions & 898,560 \\
\midrule
Matched H1$\rightarrow$H2 pairs at Default & 87,792 \\
Matched Skill$\times$Harness interaction units & 21,948 \\
Repeat-1 common-support source failures & 3,939 \\
Root-labeled source failures & 2,225 \\
\bottomrule
\end{tabular}
\end{adjustbox}

\end{table}

\subsection{Evaluation Metrics and Statistical Abbreviations}
\label{app:evaluation-metrics}

We distinguish benchmark performance, matched intervention response, success, interaction, predictive discrimination, and routing cost. Let $Y_i(c)\in[0,1]$ be the native evaluator outcome for task instance $i$ under configuration $c$; tables and figures report $100Y$ unless stated otherwise. A cell mean averages $Y_i(c)$ over its task instances and three repeated executions. A matched intervention change is
\begin{equation}
  \Delta_{i,a}=Y_i(c+a)-Y_i(c),
  \label{eq:matched-change}
\end{equation}
where the task instance and all axes other than intervention $a$ are held fixed. Positive observed change, change exceeding execution noise, and crossing the benchmark's success threshold are separate events; Section~\ref{app:validity} defines the latter two and states their calibration status.

For matched marginal summaries, let $\bar Y_s$, $\bar Y_h$, $\bar Y_b$, and $\bar Y_m$ denote balanced marginal means. We report four \emph{within-axis observed envelopes}, each clipped at zero and each defined against its own reference cell:
\begin{equation}
\begin{aligned}
  \Delta_S^+ &= \max\!\left(0,\max_{s\neq \mathrm{NoSkill}}\bar Y_s-\bar Y_{\mathrm{NoSkill}}\right),
  &\quad \Delta_H^+ &= \max\!\left(0,\max_{h\in\{H1,H2,H3\}}\bar Y_h-\bar Y_{H0}\right), \\
  \Delta_B^+ &= \max\!\left(0,\max_{b\in\{\mathrm{Low},\mathrm{Medium},\mathrm{High}\}}\bar Y_b-\bar Y_{\mathrm{Low}}\right),
  &\quad \Delta_M &= \max_m \bar Y_m-\min_m \bar Y_m.
\end{aligned}
\label{eq:marginal-effects}
\end{equation}
Each envelope is a maximum over a different candidate set and is therefore selection-biased; the four values are descriptive within-axis summaries rather than directly comparable treatment effects. $\Delta_M$ is a best-to-worst span because no Model is a natural baseline. For Skill $\times$ Harness, the matched interaction contrast is
\begin{equation}
  I_{S,H}=\left[Y(S,H2)-Y(0,H2)\right]-\left[Y(S,H1)-Y(0,H1)\right].
  \label{eq:skill-harness-interaction}
\end{equation}

A/A denotes an unchanged-configuration repeat comparison; API, application programming interface; AUPRC, area under the precision--recall curve; AUROC, area under the receiver-operating-characteristic curve; BH, Benjamini--Hochberg; CI, confidence interval; LOO, leave-one-cell-out; LOBO, leave-one-benchmark-out; OOF, out-of-fold; PRESS, prediction sum of squares; SD, standard deviation; SOP, standard operating procedure; and UID, unique identifier. The abbreviation pp denotes percentage points, and NA or N/A denotes unavailable or not applicable as specified by the table.

Table~\ref{tab:evaluation-metrics} collects the remaining metrics used in the main text and detailed analyses. These definitions fix the denominator and interpretation of each quantity; they do not turn descriptive matched comparisons into randomized causal effects.

\begin{table*}[!t]
\centering
\caption{Evaluation metrics used throughout the paper. Native outcomes are displayed as percentages unless noted otherwise.}
\label{tab:evaluation-metrics}
\scriptsize
\setlength{\tabcolsep}{4pt}
\begin{tabularx}{\textwidth}{p{0.19\textwidth}p{0.34\textwidth}X}
\toprule
Metric & Definition & Interpretation boundary \\
\midrule
Native performance & Benchmark-native evaluator outcome $Y_i(c)$; a cell mean averages task instances and three repeats. & Native scores are combined across benchmarks only by equal-weight macro averaging. \\
Matched change & $\Delta_{i,a}=Y_i(c+a)-Y_i(c)$ with the task instance and non-target axes fixed. & An observed positive change is not automatically larger than execution noise. \\
Within-axis envelope & $\Delta_S^+$, $\Delta_H^+$, $\Delta_B^+$, or $\Delta_M$ in Eq.~\ref{eq:marginal-effects}. & A selected maximum or span, not an unbiased effect or an action-count-matched ranking. \\
Skill--Harness interaction & Difference-in-differences $I_{S,H}$ in Eq.~\ref{eq:skill-harness-interaction}. & The four named patterns use a prespecified five-point descriptive tolerance; they are not A/A-calibrated significance tests. \\
Intervention opportunity & Fraction of failed source task instances with a strictly positive matched target change; standardized summaries give equal weight to the candidate actions being compared. & Measures observed rescue opportunity, not noise-qualified improvement or deployable predictability. \\
Success crossing & Target outcome reaches the benchmark-native threshold $\tau_b$ after a failed source outcome. & Partial score improvement need not constitute success. \\
Rescue, preservation, regression, persistent failure & Matched binary transitions $0\!\to\!1$, $1\!\to\!1$, $1\!\to\!0$, and $0\!\to\!0$, respectively. Net gain is rescues minus regressions over all matched units. & Rescue must be interpreted jointly with preservation and regression. \\
Predictive fit and discrimination & $R^2$ is the coefficient of determination under LOO held-cell reconstruction; AUPRC and AUROC measure ranking performance, and log loss evaluates probabilistic predictions. & Grouped or leave-one-benchmark-out splits are used where stated; none is a causal estimand. \\
Routing quality and cost & Equal-weight macro performance and mean full-chain tokens, including required probes and restarts. One policy strictly dominates another if it has higher quality and lower cost. & The Observed Oracle uses matched outcomes and measures headroom rather than deployable performance. \\
Uncertainty & Task-instance/UID cluster bootstrap confidence intervals preserve all configurations for a resampled instance; paired sign-flip or stratified permutation tests are used where specified. & Sampling uncertainty does not substitute for A/A execution-noise calibration. \\
\bottomrule
\end{tabularx}
\end{table*}

\paragraph{Abbreviations.}

\subsection{Measurement validity and operational thresholds}
\label{app:validity}

\subsubsection{Measurement Floors and Answer-Contract Effects}
\label{sec:measurement-floors}

\begin{table}[!htbp]
\centering
\caption{Root-cause composition of the 1,920 labeled \NoSkill{}+H1+Medium source failures. Column order follows the taxonomy: Procedural, Reasoning, Execution/tool, State/grounding, Horizon/termination, Factual knowledge/recall, Abstention/calibration, Evaluation/answer contract. Labeled $n$ is the benchmark denominator and sums to the same 1,920 source failures reported in Table~\ref{tab:root-intervention-opportunity}. LiveMath (305 judge labels) is excluded from root-conditioned analyses. $^*$Deterministic answer-rule hypotheses on SimpleQA, validated by an independent judge ($\kappa=0.77$) and an author check; Appendix~\ref{sec:root-cause-appendix} describes the judge and its evidence boundary. Hotpot Context and Hotpot FullWiki have one and ten labeled failures here, so their composition explains nothing about their benchmark-wide results.}
\label{tab:appendix-copy-root-composition}
\scriptsize
\setlength{\tabcolsep}{3.5pt}
\begin{adjustbox}{max width=\linewidth,max totalheight=0.85\textheight}
\begin{tabular}{lrrrrrrrrr}
\toprule
Benchmark & $n$ & Proc. & Reas. & Exec. & State & Horiz. & Fact. & Abst. & Contr. \\
\midrule
Spreadsheet & 588 & 0.00 & 0.00 & 99.83 & 0.00 & 0.17 & 0.00 & 0.00 & 0.00 \\
UCSB SOP & 266 & 78.57 & 0.00 & 7.52 & 13.91 & 0.00 & 0.00 & 0.00 & 0.00 \\
Amazon & 113 & 0.00 & 0.00 & 100.00 & 0.00 & 0.00 & 0.00 & 0.00 & 0.00 \\
SimpleQA & 783 & 0.00 & 0.00 & 0.00 & 0.00 & 0.00 & 74.84$^*$ & 24.78$^*$ & 0.38 \\
SQuAD & 159 & 0.00 & 0.00 & 0.00 & 0.00 & 0.00 & 0.00 & 0.00 & 100.00 \\
Hotpot Context & 1 & 0.00 & 0.00 & 0.00 & 0.00 & 0.00 & 0.00 & 0.00 & 100.00 \\
Hotpot FullWiki & 10 & 0.00 & 0.00 & 100.00 & 0.00 & 0.00 & 0.00 & 0.00 & 0.00 \\
\bottomrule
\end{tabular}
\end{adjustbox}

\end{table}

Before interpreting any axis mechanistically we isolate a confound that inflates the largest contrasts in the matrix. Of the 512 Default cells, 16 sit at a native-score floor below 2.0 on the 0--100 scale, and their distribution is not uniform (Table~\ref{tab:measurement-floors}). Twelve occur at H0, the one level supplying no standardized output or tool interface; the other four are all on Amazon SOP, where \Direct{} on Qwen3.5-9B returns exactly 1.7 at each of H0--H3. A score identical to one decimal across four harness levels is one reproducible failure signature, not four graded outcomes.

\begin{table}[!htbp]
\centering
\caption{Native-score floor cells in the 512-cell Default matrix, defined as a three-repeat mean below 2.0 on the 0--100 scale. $\Delta_H^+$ is that benchmark's harness envelope from Table~\ref{tab:benchmark-marginal-summary}. Spreadsheet, LiveMath, UCSB SOP, and SimpleQA contain no floor cells and are omitted. The three benchmarks with the largest harness envelopes are exactly the three with the most floor cells. We therefore do not read those envelopes as evidence that execution support repairs residual reasoning or planning failures.}
\label{tab:measurement-floors}
\scriptsize
\begin{tabularx}{\columnwidth}{lrrXr}
\toprule
Benchmark & Floor cells & At H0 & Configurations at floor & $\Delta_H^+$ \\
\midrule
Amazon SOP & 6 & 2 & Qwen+\Direct{} at H0--H3; Sonnet 5+\Direct{} at H1; Sonnet 5+\SkillOpt{} at H0 & 48.03 \\
SQuAD & 3 & 3 & Sonnet 4.6+\NoSkill{}; Sonnet 4.6+\SkillOpt{}; Sonnet 5+\NoSkill{} & 44.34 \\
Hotpot Context & 6 & 6 & Qwen+\SkillOpt{}; Sonnet 4.6+\NoSkill{}/\SkillOpt{}/\TraceToSkill{}; Sonnet 5+\SkillOpt{}; GPT-5.6-sol+\SkillOpt{} & 48.18 \\
Hotpot FullWiki & 1 & 1 & Qwen+\Direct{} & 3.92 \\
\bottomrule
\end{tabularx}

\end{table}

Two consequences follow. First, the three largest harness envelopes in Table~\ref{tab:benchmark-marginal-summary}, Amazon SOP at 48.03, Hotpot Context at 48.18, and SQuAD at 44.34, belong to exactly the three benchmarks carrying the most floor cells. For SQuAD and Hotpot Context every floor cell is at H0, and SQuAD's labeled failures are 100\% Evaluation/answer contract (Table~\ref{tab:appendix-copy-root-composition}). The parsimonious reading is that H0 leaves the answer contract unspecified, so the native scorer credits little and H0$\rightarrow$H1 largely purchases a satisfiable answer format. That is a real effect but a claim about \emph{contract specification} rather than about executable feedback repairing a residual bottleneck, so we label these contrasts contract-level and exclude them from mechanism-level interpretation. Second, a Skill can collapse a working configuration to the floor: on Hotpot Context, GPT-5.6-sol with \SkillOpt{} scores 0.0/13.6/12.2/12.2 across H0--H3 against 22.6/62.0/61.9/61.3 for \NoSkill{}, a persistent deficit near 49 points more consistent with the skill overriding the answer contract than with graded interference.

Separating a format or schema violation from a capability failure requires re-scoring the saved trajectories under a contract-relaxed evaluator, which Appendix~\ref{app:relaxed-rescoring} does. It confirms that the SQuAD and Hotpot Context envelopes are almost entirely answer-contract compliance, whereas about half of the Amazon SOP envelope is substantive repair. The two Amazon SOP interaction examples in Table~\ref{tab:canonical-interactions}, marked $\S$ there, remain floor-adjacent illustrations rather than calibrated magnitudes.

\subsubsection{Relaxed Rescoring of Answer-Format Effects}
\label{app:relaxed-rescoring}

We re-score saved trajectories with relaxed evaluators fixed before rescoring; no new rollouts are run. For SQuAD and Hotpot Context, we re-score all 115,200 trajectories (Default budget, H0--H3, four Models $\times$ four Skills $\times$ three repeats) with the official SQuAD answer normalization. Each rule credits a run if any of these candidates passes: the native extracted answer, the last \texttt{<answer>} span, the whole final response, the text after an ``Answer:'' marker, or the last line. For Amazon SOP, we re-score all 9,600 executions with a field-level parser. Re-scoring with the native rules reproduces the paper's numbers exactly.

\begin{table}[!htbp]
\centering
\caption{Relaxed rescoring of SQuAD and Hotpot Context: harness marginals on a 0--100 scale, averaged equally over Models and Skills.}
\label{tab:relaxed-rescoring}
\scriptsize
\setlength{\tabcolsep}{3pt}
\begin{adjustbox}{max width=\linewidth,max totalheight=0.85\textheight}
\begin{tabular}{llrrrrrr}
\toprule
Benchmark & Scoring & H0 & H1 & H2 & H3 & $\Delta_H^+$ & Floor cells \\
\midrule
SQuAD & Native EM & 36.65 & 76.17 & 80.31 & 80.99 & 44.34 & 3 \\
 & Extraction-tolerant EM & 36.66 & 80.56 & 80.33 & 81.01 & 44.35 & 3 \\
 & Gold span, $\le3\times$ gold length & 53.17 & 91.49 & 91.35 & 91.45 & 38.32 & 0 \\
 & Token F1 & 60.40 & 91.43 & 91.25 & 91.64 & 31.24 & 0 \\
 & Answer content present & 93.17 & 94.62 & 94.47 & 95.29 & 2.12 & 0 \\
\midrule
Hotpot Context & Native EM & 12.17 & 58.90 & 58.99 & 60.35 & 48.18 & 6 \\
 & Extraction-tolerant EM & 15.29 & 58.92 & 59.01 & 60.35 & 45.06 & 3 \\
 & Gold span, $\le3\times$ gold length & 22.77 & 65.46 & 65.69 & 67.33 & 44.56 & 1 \\
 & Token F1 & 34.65 & 74.36 & 74.57 & 76.01 & 41.37 & 0 \\
 & Answer content present & 76.82 & 73.95 & 74.24 & 76.79 & 0.00 & 0 \\
\bottomrule
\end{tabular}
\end{adjustbox}
\end{table}

\textit{Answer content present} credits a response that contains a gold answer and is at most 100 tokens long. Already at H0, 93.2\% (SQuAD) and 76.8\% (Hotpot Context) of responses contain a correct answer, and 89.2\% and 73.6\% of the native H0 failures do. Judged on content, the harness adds 2.12 and 0.00 points, so these envelopes are answer-contract compliance rather than execution repair: H0 answers usually wrap the gold span in a longer phrase, which H1's short-answer contract removes. H1$\rightarrow$H2 is interface-level as well: on SQuAD, its native $+4.14$ gain becomes $-0.24$ once answers given without tags are extracted. Under the content rule, all floor cells of both benchmarks leave the floor.

\textbf{Manual check.} We inspected 30 random native-fail $\rightarrow$ relaxed-pass flips per benchmark and rule. Weighted by stratum size, the content rule's credit is correct for 91.7--96.8\% of flips on SQuAD and 94.3--95.2\% on Hotpot Context (the lower bound counts over-inclusive or self-contradicting answers as wrong), and the gold-span rule's credit for 93.3--100\% and 96.7--100\%. The content rule therefore slightly overstates the format share.

\textbf{Cross-model transfer case.} On Hotpot Context, GPT-5.6-sol scores 62.0 at H1 without a skill but 13.6 with the \SkillOpt{} skill optimized on Qwen3.5-9B, because the skill makes the model place an ``Extracted Facts / Verification'' worksheet in its answer field. Under the content rule, a correct answer is still present in 74.1\% of its responses with the skill, against 76.8\% without.

\begin{table}[!htbp]
\centering
\caption{Composition of Amazon SOP's H1$\rightarrow$H2 gain, by why each H1 run failed. The field-level parser normalizes case, whitespace, underscores, and hyphens and accepts JSON or key--value text, but never relaxes whether the action or its values are correct. It leaves $\Delta_H^+$ almost unchanged (48.03 to 46.33) and H1$\rightarrow$H2 at $+41.70$ (native $+42.24$).}
\label{tab:amazon-relaxed}
\scriptsize
\begin{adjustbox}{max width=\linewidth,max totalheight=0.85\textheight}
\begin{tabular}{lrr}
\toprule
Reason the H1 run failed & Share of H1 failures & Share of H1$\rightarrow$H2 gain \\
\midrule
Purely cosmetic formatting & 1.6\% & 2.0\% \\
Submitted before collecting required tool observations & 31.9\% & 41.0\% \\
Wrong tool-argument values & 16.5\% & 9.2\% \\
Wrong final values & 2.8\% & 1.9\% \\
Answer-schema mismatch, e.g. extra fields (unclassified) & 39.6\% & 38.2\% \\
Other: parseable but still wrong, or ambiguous output & 7.6\% & 8.1\% \\
\bottomrule
\end{tabular}
\end{adjustbox}
\end{table}

About half of Amazon SOP's harness gain (52\%) is clearly substantive repair, at most about 40\% is answer-schema compliance, and about 2\% is surface formatting; the schema-mismatch runs could be either, so we assign them to neither side. Two of Amazon's six floor cells leave the floor under relaxed parsing. Because H0 and H1 stop at the first invalid action, offline rescoring cannot recover what the agent would have done next.

\paragraph{A/A and improvement.}
All scores below are on the native $[0,1]$ scale. An A/A comparison changes neither task instance nor configuration: for two executions $r\ne r'$, define $D^{AA}_{i,c}=Y^{(r)}_i(c)-Y^{(r')}_i(c)$. For a calibrated analysis at the task-instance level, define the benchmark tolerance as $\delta_b=Q_{0.95}(|D^{AA}_{i,c}|)$ using matched repeated executions from a prespecified, configuration-matched calibration set. A repeat-averaged contrast requires an A/A distribution of repeat-averaged differences with the same repeat count; a benchmark-average noise floor must not be substituted for a task-instance tolerance. Calibration should precede inspecting intervention contrasts. The noise-qualified improvement event and success-crossing event are distinct:
\begin{equation}
 I_{i,a}=\mathbf{1}\!\left[Y_i(a)-Y_i(c)>\delta_b\right],
 \qquad
 K_{i,a}=\mathbf{1}\!\left[Y_i(a)\geq\tau_b\right].
 \label{eq:improvement-versus-success}
\end{equation}
For binary single-execution scores, observed improvement is exactly $0\to1$. This is not automatically evidence beyond execution noise: if the task-instance A/A quantile equals 1, a single binary contrast cannot clear it and repeated evidence is required. For fractional Amazon scores, any positive difference is only an observed gain; a noise-qualified gain must clear the calibrated fractional-score tolerance.

\paragraph{Calibration status.}
The current opportunity tables report observed positive changes, not A/A-qualified gains. The saved historical A/A analyses concern behavioral metrics and do not provide a matched performance tolerance for this common-support cohort. Accordingly, numerical $\delta_b$ values remain uncalibrated for Spreadsheet, LiveMath, UCSB SOP, Amazon SOP, SimpleQA, SQuAD, Hotpot Context, and Hotpot FullWiki; no value is imputed as zero. Equation~\ref{eq:improvement-versus-success} specifies the calibration criterion and does not retroactively certify the reported counts. Bootstrap confidence intervals quantify sampling uncertainty but do not replace an execution-noise calibration.

\paragraph{Repeat design.}
We use three repeated executions per cell to capture material execution-level nondeterminism. In a controlled Spreadsheet H2 diagnostic comprising 3,360 trajectories (two Skill arms $\times$ three request seeds $\times$ three repeats $\times$ 280 task instances), the paired Skill effect had a pooled within-seed repeat SD of 2.26 points and an estimated request-seed SD of 1.56 points; fixed-seed repeat variation accounted for 68\% of the estimated variance. This diagnostic is limited to one benchmark and two Skill arms, so it motivates repeated execution but does not establish a universal variance decomposition.

\paragraph{Interaction tolerance.}
The reported four-pattern prevalence uses a fixed descriptive tolerance $\epsilon=0.05$ (five native-score points) applied to repeat-averaged four-cell contrasts. This threshold was not estimated from A/A. A noise-qualified interaction analysis would additionally require a matched null for the four-cell interaction contrast, not merely reuse a two-cell improvement tolerance. The existing prevalence therefore characterizes thresholded response patterns rather than effects certified to exceed A/A noise.

\paragraph{Label coverage and success.}
The repeat-1 common-support cohort contains 3,939 source failures; 2,225 have assigned labels (56.5\%) and 1,714 remain unassigned (43.5\%). Root-conditioned percentages describe the assigned subset. Evidence-backed labels and model-assisted hypotheses retain their distinct status. Capability crossings use $\tau_b=1$ for all eight benchmarks: binary task instances must pass their native scorer, and Amazon must attain full field accuracy. Implementations allow $10^{-9}$ numerical slack at 1. Partial Amazon improvements need not cross this strict full-success threshold. BigCode is excluded from this common-support analysis.

\section{Skill Provenance, Split Isolation, and Transfer Sensitivity}
\label{app:skill-provenance}

\paragraph{Construction and selection.}
NoSkill supplies no learned artifact. Direct synthesizes a procedure once from a frozen catalog, without rollout collection or a validation gate. SkillOpt uses training rollouts for iterative edits, and locally generated Trace2Skill distills training-trajectory lessons under its recorded method-validation procedure. The protocol reserves method validation for skill selection and separates harness qualification, which qualifies execution code and interfaces rather than selecting skill content. Frozen packages are reused unchanged across target models instead of regenerated for each target. Holding the artifact fixed does not remove its potential alignment to the construction model.

\paragraph{UID and artifact evidence.}
The anonymized supplementary archive contains a provenance inventory recording exact archived UIDs, split-file hashes, within-directory overlap checks, and saved freeze declarations, together with the cross-model artifact bundle from which every frozen skill package is imported. UCSB training and test UIDs come from the archived 280-item SOP split, whose original validation set is partitioned into method and harness validation. LiveMath uses the archived ten-item method-validation split; Hotpot stores its Context and FullWiki views beneath a common split root. The protocol excludes test instances from generation, selection, patching, and acceptance. Disjoint UID lists and saved freeze declarations support this rule but are not a complete audit of every historical generation call. Hash-freezing specifies unchanged cross-model import; the inventory does not independently compare artifact bytes in every provider execution.

\paragraph{Spreadsheet.}
\SkillOpt{} and \TraceToSkill{} both use the 80 training and 24 method-validation items. \TraceToSkill{} starts from the authors' released initial skill bundle and runs the full evolution locally with Qwen3.5-9B on these items (seed 42); the evolved \texttt{xlsx/SKILL.md} is the frozen artifact, and the two released reference skills are carried over unchanged. The \Direct{} catalog contains only the 80 training items: an archived validation manifest lists 80 training plus 24 method-validation items, but the frozen generation record and the prompt contain only the training items. The archived provenance records freeze entrypoints and package hashes before evaluation.

\paragraph{Qwen-excluded sensitivity.}
We retain Sonnet 4.6, Sonnet 5, and GPT-5.6-sol on the same task instances and frozen artifacts. Skill gains average matched NoSkill contrasts over H0--H3 at Default and over retained models. Interaction prevalence retains the original H1/H2 classifier, three-repeat means, $\epsilon=0.05$, and equal benchmark weights. BigCode is excluded. Intervals use 2,000 instance-UID cluster bootstrap draws with seed 42, preserving all configurations per sampled task instance. Removing Qwen reduces interaction units from 21,948 to 16,461. No new model executions are used.

\begin{table*}[!t]
\centering
\caption{Skill transfer sensitivity at Default, averaged over H0--H3. Gains are paired against NoSkill; API-only excludes Qwen. Intervals resample task UIDs within benchmark (2,000 draws), retaining all models and harnesses for each task.}
\label{tab:api-skill-gains}
\small
\begin{adjustbox}{max width=\linewidth,max totalheight=0.85\textheight}
\begin{tabular}{llrr}
\toprule
Benchmark & Skill & Four-model gain (pp) & API-only gain [95\% CI] (pp) \\
\midrule
Amazon & Direct & -13.29 & +2.20 [+0.33, +4.09] \\
Amazon & SkillOpt & -0.09 & +3.01 [+1.71, +4.25] \\
Amazon & Trace2Skill & -2.92 & -0.25 [-1.70, +1.15] \\
Hotpot Context & Direct & +1.63 & +2.75 [+1.92, +3.65] \\
Hotpot Context & SkillOpt & -13.48 & -14.26 [-15.64, -12.81] \\
Hotpot Context & Trace2Skill & +1.71 & +1.75 [+1.06, +2.44] \\
Hotpot FullWiki & Direct & -3.02 & -0.68 [-1.99, +0.61] \\
Hotpot FullWiki & SkillOpt & -4.67 & -7.21 [-8.75, -5.75] \\
Hotpot FullWiki & Trace2Skill & +10.58 & +10.93 [+9.17, +12.61] \\
LiveMath & Direct & -1.50 & -1.16 [-3.07, +0.83] \\
LiveMath & SkillOpt & +2.20 & +1.14 [-1.52, +3.79] \\
LiveMath & Trace2Skill & +8.15 & +3.99 [+1.07, +6.83] \\
SQuAD & Direct & +12.26 & +10.50 [+9.02, +11.88] \\
SQuAD & SkillOpt & +6.86 & +4.49 [+3.30, +5.65] \\
SQuAD & Trace2Skill & +17.13 & +15.61 [+14.19, +17.10] \\
SimpleQA & Direct & -1.83 & -2.56 [-3.81, -1.31] \\
SimpleQA & SkillOpt & +0.12 & -0.20 [-1.07, +0.65] \\
SimpleQA & Trace2Skill & -0.12 & -0.28 [-0.96, +0.45] \\
Spreadsheet & Direct & -1.64 & -2.85 [-3.81, -1.86] \\
Spreadsheet & SkillOpt & -4.87 & -5.58 [-6.84, -4.37] \\
Spreadsheet & Trace2Skill & -6.23 & -7.49 [-9.00, -5.91] \\
UCSB SOP & Direct & -0.04 & -1.02 [-2.38, +0.25] \\
UCSB SOP & SkillOpt & +5.55 & +3.59 [+2.16, +5.05] \\
UCSB SOP & Trace2Skill & +1.89 & -0.17 [-1.16, +0.86] \\
\bottomrule
\end{tabular}
\end{adjustbox}
\end{table*}
\begin{table}[!htbp]
\centering
\caption{Interaction prevalence with and without Qwen. Original H1/H2 rules, $\epsilon=0.05$, complete three-repeat cells, and benchmark-equal weighting are retained.}
\label{tab:api-regimes}
\small
\begin{adjustbox}{max width=\linewidth,max totalheight=0.85\textheight}
\begin{tabular}{lrr}
\toprule
Regime & Four models (\%) & API only [95\% CI] (\%) \\
\midrule
Absorption & 3.77 & 2.79 [2.28, 3.31] \\
Activation & 2.35 & 2.02 [1.65, 2.46] \\
Complementarity & 1.48 & 1.27 [0.90, 1.70] \\
Interference & 6.46 & 5.13 [4.59, 5.67] \\
\bottomrule
\end{tabular}
\end{adjustbox}
\end{table}

FullWiki, SQuAD, LiveMath, and UCSB retain positive gains for their responsive skills (Table~\ref{tab:api-skill-gains}). LiveMath's Trace2Skill gain contracts from +8.15 to +3.99 points. Amazon also changes: its marginal SkillOpt gain moves from $-0.09$ to +3.01 points, illustrating dependence on model composition. All four named patterns persist with lower combined macro prevalence. These results establish transfer to the tested API targets, not construction-model neutrality or a causal explanation of alignment. Intervals reflect task sampling, not sampling of model families.

\section{Self-Evolved Skill Experiments}
\label{app:skill-experiments}

This section gives the designs and full results of the four experiments behind Section~\ref{sec:selfevo}. Unless stated otherwise, skills are evaluated on the locked test split with the paper's protocol (Default budget, H0--H3, three repeats, the paper's evaluators), and intervals are paired task-cluster bootstrap 95\% intervals (2,000 draws, seed 42).

\subsection{Stronger Constructor}
\label{app:skill-constructor}
\textbf{Design.} We rebuild \SkillOpt{} and \TraceToSkill{} on UCSB SOP, a procedure task, and on SimpleQA, the missing-information control, with GPT-5.6-sol replacing Qwen3.5-9B throughout construction: GPT-5.6-sol is the rollout model, the optimizer or analyst, and, on SimpleQA, the construction-time grader. Training and method-validation data, initial skills, hyperparameters, and validation gates are unchanged. Both GPT-5.6-sol-built \TraceToSkill{} candidates pass their gates; the \SkillOpt{} artifacts were not revalidated. The frozen skills are evaluated on Qwen3.5-9B (UCSB SOP 175 tasks, SimpleQA 300). On both benchmarks the Qwen-built \TraceToSkill{} artifact is the human-written seed, because its evolved candidate was rejected (Table~\ref{tab:trace2skill-gate}).

\textbf{Results.} Table~\ref{tab:app-skill-constructor}. On UCSB SOP, the GPT-5.6-sol-built \TraceToSkill{} skill beats the human-written seed, which the Qwen-built candidate could not, whereas the GPT-5.6-sol-built \SkillOpt{} skill is worse than its Qwen-built version. Neither changes SimpleQA. On SimpleQA, 3,596 (\SkillOpt{}) and 3,592 (\TraceToSkill{}) of 3,600 cells completed; the missing cells are failed requests.

\begin{table}[!htbp]
\centering
\caption{Stronger constructor: mean scores and paired differences on Qwen3.5-9B.}
\label{tab:app-skill-constructor}
\small
\setlength{\tabcolsep}{4pt}
\begin{adjustbox}{max width=\linewidth,max totalheight=0.85\textheight}
\begin{tabular}{lrr}
\toprule
 & UCSB SOP & SimpleQA \\
\midrule
\NoSkill{} & 53.90 & 6.33 \\
Human-written seed & 62.00 & 6.75 \\
\SkillOpt{} (Qwen-built) & 65.33 & 7.47 \\
\SkillOpt{} (GPT-5.6-sol-built) & 59.29 & 7.42 \\
\TraceToSkill{} (GPT-5.6-sol-built) & 67.38 & 6.85 \\
\midrule
GPT \TraceToSkill{} $-$ seed & $+5.38$ [$+1.71,+9.14$] & $+0.09$ [$-1.87,+1.97$] \\
GPT \SkillOpt{} $-$ Qwen \SkillOpt{} & $-6.05$ [$-10.24,-1.90$] & $-0.06$ [$-1.86,+1.58$] \\
GPT \SkillOpt{} $-$ seed & $-2.71$ [$-5.33,-0.10$] & -- \\
GPT \TraceToSkill{} $-$ \NoSkill{} & $+13.48$ [$+9.76,+17.57$] & $+0.61$ [$-1.58,+2.69$] \\
GPT \SkillOpt{} $-$ \NoSkill{} & $+5.38$ [$+1.81,+9.14$] & $+0.75$ [$-1.17,+2.75$] \\
\bottomrule
\end{tabular}
\end{adjustbox}
\end{table}

\subsection{Same Constructor, Different Procedure}
\label{app:skill-procedure}
\textbf{Design.} All artifacts are built by Qwen3.5-9B: \Direct{} by a single call, \SkillOpt{} by iterative optimization from scored outcomes, and \TraceToSkill{} by evolution from trajectories (a genuinely evolved artifact only on Spreadsheet and LiveMath). Holding the constructor fixed isolates the procedure. Scores are averaged over all four target Models.

\textbf{Results.} Table~\ref{tab:app-skill-procedure}. \Direct{} beats \SkillOpt{} on four of eight benchmarks, and \SkillOpt{} falls below \NoSkill{} on Hotpot Context, Hotpot FullWiki, and Spreadsheet. Iteration helps on the procedure-heavy tool tasks (UCSB SOP and Amazon SOP). Against the evolved \TraceToSkill{} artifact, \Direct{} is better on Spreadsheet ($+4.59$ [$+3.47,+5.73$]) and worse on LiveMath ($-9.64$ [$-13.05,-6.35$]), where the evolved skill carries the meta-option shortcut. Each method has one construction seed, so these compare single artifacts, not method distributions.

\begin{table}[!htbp]
\centering
\caption{Same Qwen3.5-9B constructor, different procedure, averaged over Models.}
\label{tab:app-skill-procedure}
\small
\setlength{\tabcolsep}{4pt}
\begin{adjustbox}{max width=\linewidth,max totalheight=0.85\textheight}
\begin{tabular}{lrrrl}
\toprule
Benchmark & \NoSkill{} & \Direct{} & \SkillOpt{} & \Direct{} $-$ \SkillOpt{} \\
\midrule
Hotpot Context & 50.14 & 51.77 & 36.66 & $+15.11$ [$+13.51,+16.73$] \\
SQuAD & 59.47 & 71.73 & 66.33 & $+5.40$ [$+4.13,+6.69$] \\
Spreadsheet & 34.78 & 33.15 & 29.91 & $+3.24$ [$+2.31,+4.15$] \\
Hotpot FullWiki & 27.31 & 24.28 & 22.63 & $+1.65$ [$+0.44,+2.83$] \\
SimpleQA & 33.98 & 32.15 & 34.10 & $-1.96$ [$-2.92,-0.90$] \\
LiveMath & 39.05 & 37.55 & 41.25 & $-3.70$ [$-6.37,-1.06$] \\
UCSB SOP & 62.45 & 62.42 & 68.00 & $-5.58$ [$-7.25,-3.93$] \\
Amazon SOP & 67.43 & 54.14 & 67.35 & $-13.20$ [$-14.67,-11.88$] \\
\bottomrule
\end{tabular}
\end{adjustbox}
\end{table}

\subsection{Rejected \texorpdfstring{\TraceToSkill{}}{Trace2Skill} Candidates}
\label{app:skill-rejected}
\textbf{Design.} On the benchmarks where \TraceToSkill{}'s evolved candidate failed its validation gate, the frozen artifact is the human-written seed. We run the rejected candidates themselves on the locked test split with Qwen3.5-9B, which tests the distillation step separately from the gate. The seed arm is the paper's frozen \TraceToSkill{} arm, read unchanged. Scores use each benchmark's master-table metric (SimpleQA accuracy, SQuAD and Hotpot F1, Amazon SOP field accuracy, UCSB SOP success); for SQuAD we also report exact match.

\textbf{Results.} Table~\ref{tab:app-skill-rejected}. The rejected candidate is significantly worse than its seed on UCSB SOP and on SQuAD exact match and indistinguishable from it on Amazon SOP, Hotpot Context, and SimpleQA, so on these benchmarks distillation did not produce a better skill than the few-line seed. On Hotpot FullWiki, the candidate is better; its gate was decided on 20 method-validation items (0.55 versus 0.50), so the small validation set rejected a better candidate. On SimpleQA, 3,594 of 3,600 matched cells completed.

\begin{table}[!htbp]
\centering
\caption{Rejected \TraceToSkill{} candidates against their human-written seeds on Qwen3.5-9B. Gate: method-validation score of seed / candidate.}
\label{tab:app-skill-rejected}
\small
\setlength{\tabcolsep}{4pt}
\begin{adjustbox}{max width=\linewidth,max totalheight=0.85\textheight}
\begin{tabular}{lcrrrl}
\toprule
Benchmark & Gate & Tasks & Seed & Candidate & Candidate $-$ Seed \\
\midrule
UCSB SOP & 0.40 / 0.35 & 175 & 62.00 & 49.19 & $-12.81$ [$-17.76,-8.19$] \\
Amazon SOP & 0.00 / 0.00 & 50 & 50.52 & 45.38 & $-5.14$ [$-16.00,+5.24$] \\
Hotpot Context & 0.55 / 0.50 & 300 & 63.69 & 61.94 & $-1.75$ [$-3.46,+0.05$] \\
SimpleQA & 0.10 / 0.10 & 300 & 6.75 & 6.34 & $-0.41$ [$-2.00,+1.22$] \\
SQuAD (F1) & 0.80 / 0.75 & 300 & 83.39 & 85.43 & $+2.04$ [$-0.18,+4.29$] \\
SQuAD (exact match) & & 300 & 72.97 & 67.03 & $-5.94$ [$-9.78,-2.11$] \\
Hotpot FullWiki & 0.55 / 0.50 & 300 & 48.95 & 60.57 & $+11.62$ [$+8.11,+15.16$] \\
\bottomrule
\end{tabular}
\end{adjustbox}
\end{table}

\subsection{SimpleQA Knowledge Access}
\label{app:skill-simpleqa}
\textbf{Design.} If the bottleneck on SimpleQA is missing information, a skill built from the agent's own data should not help, a skill can at most make the model recognize the gap, and supplying the information should move the score. We run Qwen3.5-9B under the paper's H1 protocol (no browsing, skill in the fixed package block, answer in \texttt{<answer>} with an ``I don't know'' option) on the 300 locked test items, with one repeat, temperature 0, seed 42, and the paper's SimpleQA grader. Only the skill changes, plus one knowledge-access turn in two arms (Table~\ref{tab:app-skill-simpleqa}). The search corpus contains, for each item, the source pages cited in its SimpleQA metadata (293 of 300 items have text), split into 120-word chunks and retrieved with BM25. It therefore measures what happens once the information is available, not the quality of open web search, and the external-knowledge skill is an upper bound rather than a valid construction method. $\Delta$ is a paired item-bootstrap interval against \NoSkill{}.

\textbf{Results.} A skill built from the agent's own data does not help. The knowledge-boundary skill replaces most wrong answers with explicit requests but cannot raise accuracy on its own. The same skill with search on request reaches 60.0\%, a gain comparable to switching to a stronger model in the matrix ($\Delta_M=60.85$). Knowledge written into a static skill helps only in proportion to what fits: it reaches 43.3\% at full prompt budget and 13.0\% at the size of the own-data skill. Only Qwen3.5-9B is tested, with one repeat and the paper's Qwen3.5-9B grader.

\begin{table}[!htbp]
\centering
\caption{SimpleQA knowledge-access arms (Qwen3.5-9B, H1, 300 test items).}
\label{tab:app-skill-simpleqa}
\small
\setlength{\tabcolsep}{3pt}
\begin{tabularx}{\textwidth}{>{\raggedright\arraybackslash}p{0.22\textwidth}Xrlr}
\toprule
Arm & Skill and knowledge access & Acc. & $\Delta$ vs.\ \NoSkill{} & Wrong \\
\midrule
\NoSkill{} & None & 5.0 & -- & 263 \\
Own-data facts & The 90 training and method-validation questions with their answers (disjoint from test) & 6.3 & $+1.3$ [$-1.3,+3.7$] & 251 \\
Knowledge boundary & Human-written: answer only facts you recall; otherwise answer ``I don't know'' and state a search query & 2.0 & $-3.0$ [$-5.3,-1.0$] & 45 \\
\quad + search on request & One search turn returns the top-3 passages when the model requests it & 60.0 & $+55.0$ [$+49.0,+60.7$] & 54 \\
\NoSkill{} + search always & The top-3 passages are always given with the question & 72.0 & $+67.0$ [$+61.0,+72.7$] & 34 \\
External-knowledge skill & 292 notes from the items' source pages, no questions or answers (upper bound) & 43.3 & $+38.3$ [$+32.7,+44.0$] & 96 \\
\quad skill-sized & The same notes cut to the size of the own-data skill & 13.0 & $+8.0$ [$+4.0,+12.7$] & 44 \\
\bottomrule
\end{tabularx}
\end{table}

\section{Detailed Analyses for the Results}
\label{app:rq-analyses}

This section mirrors the order of Sections~\ref{sec:when} and~\ref{sec:which}. It first documents the full factorial response surface and the two observations derived from it, then provides the detailed estimands, empirical evidence, uncertainty analyses, and interpretation boundaries for RQ1--RQ5.

\subsection{No Universally Optimal Agent Configuration}
\label{app:no-context-free-winner-details}

\subsubsection{Detailed Evidence for Context-Dependent Rankings}
\label{app:context-free-winner-details}

The absence of a context-free winner is first visible along the Model axis. Among the 32 Benchmark$\times$Skill rows of the Default matrix, 30 have a unique best Model, distributed 12/10/6/2 across GPT-5.6-sol, Sonnet 4.6, Sonnet 5, and Qwen3.5-9B, respectively. These are not merely near-tie reversals: Sonnet 4.6 reaches the benchmark-wide maximum on Spreadsheet at 53.1 but ranks last on LiveMath at 37.1, whereas GPT-5.6-sol reaches 69.4 on SimpleQA but ranks last on Hotpot FullWiki at 32.6, behind Sonnet 5 at 45.0. Figure~\ref{fig:heterogeneity}a visualizes the SQuAD+\NoSkill{} reversal under changing Harness support.

The other intervention axes exhibit the same conditionality. After balanced marginalization, \TraceToSkill{} contributes to positive Skill envelopes on Hotpot FullWiki ($\Delta_S^+=10.58$ points) and LiveMath ($8.15$), whereas no fixed alternative Skill outperforms \NoSkill{} on Spreadsheet or Amazon SOP ($\Delta_S^+=0$ for both). Harness effects are large on Amazon SOP and on the answer-interface-sensitive SQuAD and Hotpot Context, but are nearly flat on LiveMath and SimpleQA; Appendix~\ref{sec:measurement-floors} shows that the largest of these contrasts also contain floor and answer-contract effects. Budget exhibits an even sharper task boundary: Amazon SOP improves dramatically when Low is replaced by a feasible Medium or High resource ceiling because Low does not provide enough turns or wall-clock time to complete its long tool workflow ($\Delta_B^+=81.13$), whereas SimpleQA is essentially saturated ($\Delta_B^+=0.10$). Thus, stronger Skills, richer Harnesses, and larger Budgets can each improve performance, have little effect, or reduce performance depending on the task and the components already present. Figure~\ref{fig:heterogeneity}b contrasts three representative task profiles.

The same contrast can appear within a single Benchmark$\times$Model block. A floor-free example is Qwen3.5-9B on Spreadsheet. Moving from H1 to H2 raises \NoSkill{} from 9.3 to 24.9, \Direct{} from 13.3 to 28.6, \SkillOpt{} from 9.3 to 22.5, and \TraceToSkill{} from 10.1 to 22.7, showing a consistent execution-support gain. Skill effects remain conditional: \Direct{} exceeds \NoSkill{} at H0 and H1, but \NoSkill{} is strongest at H3 (30.4 versus 28.8, 26.2, and 25.1 for \Direct{}, \SkillOpt{}, and \TraceToSkill{}). Figure~\ref{fig:heterogeneity}c visualizes the complete $4\times4$ block.

\subsection{Within-Benchmark Response Heterogeneity}
\label{app:within-benchmark-response-heterogeneity}

A benchmark mean compresses distinct transitions at the task-instance level into one number:
\begin{equation}
  \mathbb{E}[\Delta]
  = p_{\mathrm{rescue}}\,\mathbb{E}[\Delta\mid\mathrm{rescue}]
  + p_{\mathrm{neutral}}\,\mathbb{E}[\Delta\mid\mathrm{neutral}]
  + p_{\mathrm{regression}}\,\mathbb{E}[\Delta\mid\mathrm{regression}].
  \label{eq:mean-transition-mixture}
\end{equation}
Figure~\ref{fig:heterogeneity}d illustrates this compression directly. Under the frozen post-hoc LiveMath partition, \TraceToSkill{} versus \NoSkill{} changes Sonnet 5 performance by $+6.60$ points on 53 meta-option instances and $-4.93$ points on 71 non-meta instances; the weighted whole-benchmark contrast averaged over H0--H3 is approximately zero. Because the partition was defined after Skill selection and evaluation, this is descriptive subgroup evidence rather than a preregistered causal contrast. Table~\ref{tab:surface-heterogeneity-support} provides four additional matched-unit views. Across 21,948 Skill$\times$Harness units, 85.94\% are weak or neutral under the prespecified five-point rule, while the remainder separates into absorption, activation, complementarity, and interference. On Amazon SOP, 1,070 of 2,400 fractional-score executions improve, 1,322 remain unchanged, and eight degrade. Thus, a positive or near-zero benchmark mean can conceal substantial and opposing changes at the task-instance level.

\begin{table*}[t]
\centering
\caption{Direct empirical support for within-benchmark response heterogeneity. Skill$\times$Harness percentages are benchmark-equal prevalence estimates under a prespecified five-point descriptive threshold. H1$\rightarrow$H2 entries are pooled matched transition counts for the seven binary-outcome benchmarks; conditional rates divide rescues by H1 failures and regressions by H1 successes. Full benchmark distributions and uncertainty are reported in Appendix Table~\ref{tab:appendix-h1-h2-transitions}. These are descriptive matched outcomes. The LiveMath partition is frozen but post hoc; the interaction labels depend on an uncalibrated five-point tolerance; and the H1/H2 executions are matched replays rather than randomized prospective continuations.}
\label{tab:surface-heterogeneity-support}
\small
\begin{tabularx}{\textwidth}{p{0.19\textwidth}p{0.16\textwidth}X}
\toprule
Matched analysis & Population & Empirical composition \\
\midrule
Skill$\times$Harness & 21,948 task-instance--model--alternative-Skill units & Weak/neutral 85.94\%; Absorption 3.77\%; Activation 2.35\%; Complementarity 1.48\%; Interference 6.46\%. \\
H1$\rightarrow$H2 & 85,392 binary-outcome executions & Rescue 3,886 (8.54\% of H1 failures); Preserve 37,568; Regression 2,340 (5.86\% of H1 successes); Persistent failure 41,598. \\
LiveMath Skill subgroup & 124 Sonnet 5 task instances, averaged over Default H0--H3 and three repeats & \TraceToSkill{} minus \NoSkill{}: meta-option $+6.60$ points (53 instances); non-meta $-4.93$ points (71 instances); all instances, averaged over H0--H3, $0.00$ points. \\
Amazon H1$\rightarrow$H2 & 2,400 fractional-score executions & Improve 1,070; unchanged 1,322; degrade 8; mean change $+42.24$ points. \\
\bottomrule
\end{tabularx}

\end{table*}

\subsubsection{LiveMath Meta-Option Diagnostic}
\label{app:livemath-meta}
The frozen post-hoc partition marks an item as meta when it contains the recurring meta option; 53 of the 124 test items meet this rule. The frozen \SkillOpt{} and \TraceToSkill{} documents explicitly call attention to this choice, but differ in when they recommend selecting it. For Qwen3.5-9B, averaging three repeats over Default H0--H3, \TraceToSkill{} scores 71.5\% versus 11.2\% for \NoSkill{} on meta items, but 43.7\% versus 52.7\% on the 71 non-meta items. The subgroup was defined after skill selection and evaluation, so this is a descriptive interaction; no ablation isolates the meta-option instruction from the rest of either Skill.

\subsection{How Interventions Interact}
\label{app:rq2-details}
\label{app:interaction-details}
\label{sec:interaction-regimes}

\begin{figure}[!htbp]
\centering
\includegraphics[width=0.8\linewidth,height=0.85\textheight,keepaspectratio]{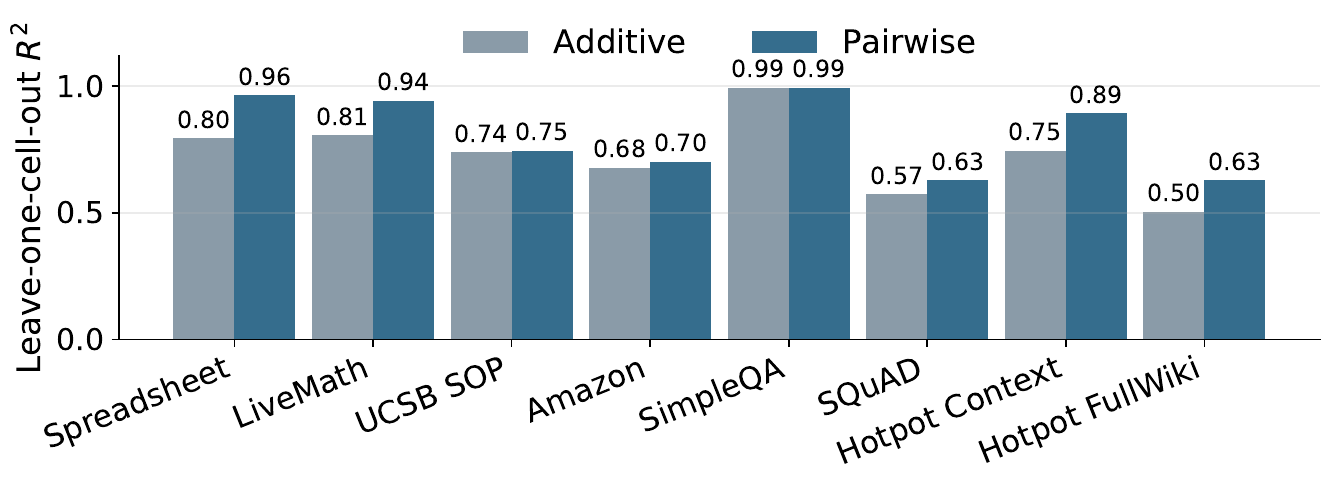}
\caption{Leave-one-cell-out $R^2$ of additive and pairwise-interaction models, fitted within each benchmark over the 64 Default-budget cells.}
\label{fig:interaction}
\end{figure}

\paragraph{Fitting procedure and test.} We fit categorical response surfaces separately within each benchmark after min--max normalization. The additive specification contains Model, Skill, and Harness main effects; the interaction specification additionally contains all Model$\times$Skill, Model$\times$Harness, and Skill$\times$Harness terms. We evaluate both with PRESS leave-one-cell-out prediction over the 64 Default-budget cells per benchmark. The interaction specification increases mean LOO $R^2$ from 0.729 to 0.812, with a mean paired increase of 0.083 and improvement on seven of eight benchmarks (Figure~\ref{fig:interaction}). An exact one-sided sign-flip test over the eight benchmark-level $R^2$ differences gives $p=0.0078$. This test supports non-additive context dependence within the observed benchmark surfaces; it does not establish generalization to unseen benchmarks or identify residual failure as its cause.

For matched four-cell Skill$\times$Harness comparisons, let $Y_{00}=Y(\mathrm{NoSkill},H1)$, $Y_{10}=Y(s,H1)$, $Y_{01}=Y(\mathrm{NoSkill},H2)$, and $Y_{11}=Y(s,H2)$, with $I_{S\times H}=Y_{11}-Y_{10}-Y_{01}+Y_{00}$. Using the fixed descriptive native-score tolerance $\epsilon=0.05$ specified in Appendix~\ref{app:validity}, Absorption occurs when both interventions help but their joint result stays within tolerance of the stronger one; Activation when neither alone clears tolerance but the combination does; Complementarity when the joint result exceeds the stronger individual intervention with positive interaction residual; and Interference when the combination underperforms a useful individual intervention.

\paragraph{Interaction patterns are recurrent but sparse.}
Across 21,948 matched task-instance--model--alternative-Skill units, Absorption accounts for 3.77\%, Activation 2.35\%, Complementarity 1.48\%, and Interference 6.46\% under benchmark-equal weighting. The four named patterns cover 14.06\% in total; the remaining 85.94\% are weak, neutral, or outside the prespecified definitions. An interaction therefore exists only in a minority of matched units, but its sign changes systematically enough to reject a uniformly additive view of external support.

The empirical taxonomy uses the same four H1/H2 cells for each task instance and Model. \emph{Absorption} means both interventions help but their combination stays near the stronger one; \emph{activation} means neither alone clears the fixed improvement tolerance while their combination does; \emph{complementarity} means the combination exceeds the stronger individual intervention with a positive interaction residual; and \emph{interference} means stacking reduces the benefit of a useful intervention. The exact five-point rules use the matched four-cell contrast defined above. Table~\ref{tab:canonical-interactions} reports representative panels averaged over all task instances from the frozen Default matrix.

\begin{table*}[!t]
\centering
\caption{Recurring Skill--Harness patterns using all task instances at Default budget in Table~\ref{tab:default-matrix-transposed}. Every raw H0--H2 score is copied from the corresponding Benchmark--Model--Skill cell of Table~\ref{tab:default-matrix-transposed}; Skill gains and $I(H2\mid H1)$ are computed from the displayed one-decimal means. These four patterns are illustrated at the panel-mean level, whereas population prevalence is classified per task-instance--model unit. Model superscripts: $^{\dagger}$ Qwen3.5-9B; $^{\ddagger}$ Sonnet 5. $^{\S}$ One endpoint of the contrast is a native-score floor cell (Appendix~\ref{sec:measurement-floors}); the pattern label is qualitative and the magnitude is not a calibrated interaction estimate. Blue/red bold highlights selected favorable/unfavorable outcomes and contrasts; it does not indicate statistical significance, and no cell in this table carries a standard error. H0 provides context; the pattern rule uses H1/H2 with a five-point descriptive threshold that was chosen a priori and not derived from A/A variation. Outcome-selected subset illustrations use different denominators and appear separately in Appendix Figure~\ref{fig:interaction-regimes}.}
\label{tab:canonical-interactions}
\scriptsize
\resizebox{\textwidth}{!}{%
\begin{tabular}{lllrrrl}
\toprule
Pattern & Benchmark & Skill/contrast & H0 & H1 & H2 & $I(H2\mid H1)$ \\
\midrule
\multirow{3}{*}{Absorption$^{\dagger}$} & \multirow{3}{*}{SQuAD} & \NoSkill{} & 10.7 & 36.6 & 78.6 & \\
& & \SkillOpt{} & 29.0 & 77.0 & 77.6 & \\
& & Skill gain & +18.3 & \gain{+40.4} & \harm{$-1.0$} & \harm{$-41.4$} \\
\midrule
\multirow{3}{*}{Activation$^{\dagger}$} & \multirow{3}{*}{Hotpot FullWiki} & \NoSkill{} & 24.2 & \textbf{25.9} & \textbf{26.8} & \\
& & \SkillOpt{} & 21.1 & \textbf{24.0} & \gain{34.9} & \\
& & Skill gain & $-3.1$ & $-1.9$ & +8.1 & \gain{+10.0} \\
\midrule
\multirow{3}{*}{Complementarity$^{\ddagger\S}$} & \multirow{3}{*}{Amazon SOP} & \NoSkill{} & 2.6 & \textbf{4.2} & \gain{58.9} & \\
& & \Direct{} & 2.6 & \harm{1.7} & \gain{85.7} & \\
& & Skill gain & 0.0 & $-2.5$ & +26.8 & \gain{+29.3} \\
\midrule
\multirow{3}{*}{Interference$^{\dagger\S}$} & \multirow{3}{*}{Amazon SOP} & \NoSkill{} & 28.3 & 33.9 & 88.6 & \\
& & \Direct{} & 1.7 & 1.7 & 1.7 & \\
& & Skill gain & \textbf{$-26.6$} & \harm{$-32.2$} & \harm{$-86.9$} & \harm{$-54.7$} \\
\bottomrule
\end{tabular}}

\end{table*}

\paragraph{Absorption arises when both components address the same bottleneck.}
On Qwen3.5-9B SQuAD, \SkillOpt{} gains 40.4 points over \NoSkill{} at H1, but its incremental gain contracts to $-1.0$ point at H2. A consistent explanation is that the Skill and stronger Harness both help satisfy the answer procedure or contract; once H2 has removed that bottleneck, the Skill has little residual opportunity. This is descriptive overlap rather than an identified mediator, especially because SQuAD contains answer-contract floor effects.

\paragraph{Activation and complementarity arise when components address different bottlenecks.}
On Qwen3.5-9B Hotpot FullWiki, neither \SkillOpt{} alone nor H2 alone improves the H1 reference by five points, but the combination gains 9.0 points. The pattern is consistent with a division of labor: the Skill supplies a reusable search policy, while the Harness supplies the execution support needed to carry it out. The Sonnet 5 Amazon example is an even larger complementarity pattern, but one endpoint lies at the native-score floor; it is retained as a qualitative illustration rather than a calibrated magnitude.

\paragraph{Interference arises when stronger execution preserves a mismatched policy.}
On Qwen3.5-9B Amazon SOP, H2 raises \NoSkill{} from 33.9 to 88.6, whereas \Direct{} remains at 1.7 under H0--H3. The result is consistent with a mismatched Skill steering the system toward a bad policy that additional execution support does not undo. Because the 1.7 endpoint is an instrument floor, this row establishes catastrophic incompatibility within the tested configuration but does not quantify a general causal interaction.

\paragraph{The interaction sign follows the residual bottleneck.}
\emph{The sign of an interaction depends on what failure remains after the first intervention is applied.} If the first component removes the bottleneck targeted by the second, incremental value is absorbed; if it exposes a different bottleneck that the second can repair, the pair activates or becomes complementary; if it imposes a mismatched policy that the second component executes or preserves, the pair can interfere. The API-model subset appears in Table~\ref{tab:api-regimes}.

\begin{figure*}[!tp]
  \centering
  \includegraphics[width=\textwidth,height=0.85\textheight,keepaspectratio]{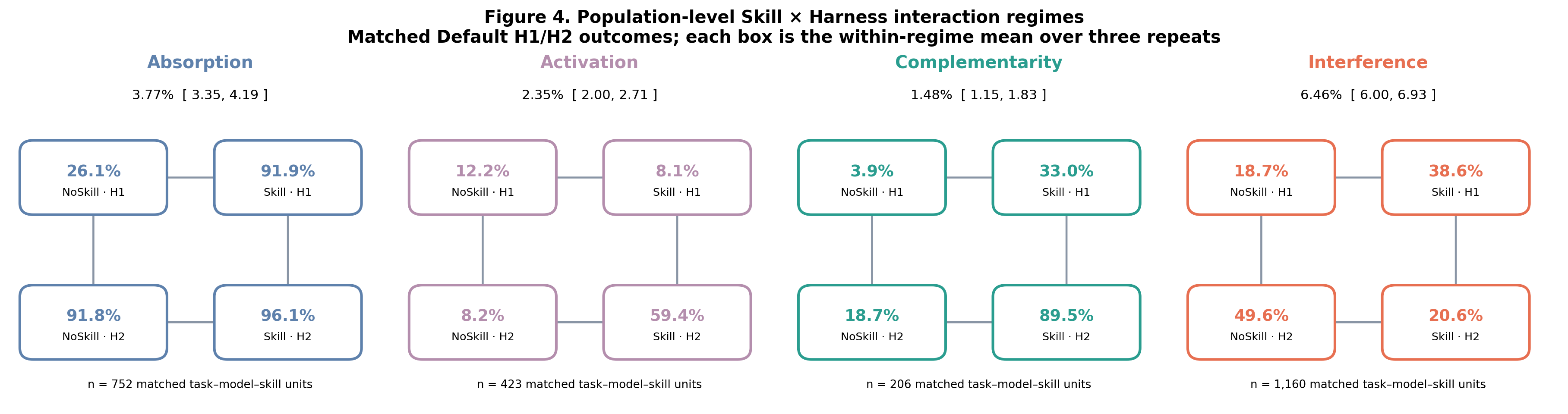}
  \caption{Outcome-selected task-instance--model subset examples of the four matched Skill $\times$ Harness interaction patterns. Each panel compares the same task instances under \NoSkill{} and an alternative skill across H1 and H2. These conditional subset means differ from the Table~\ref{tab:canonical-interactions} panel means over all task instances.}
  \label{fig:interaction-regimes}
\end{figure*}

\subsection{Which Interventions Help Under Which Failure}
\label{app:rq3-details}
\label{sec:root-cause-appendix}

We assign the earliest decisive error supported by the saved trajectory to one of the eight categories in Table~\ref{tab:appendix-root-taxonomy}. Every label comes from that trajectory alone. The human-authored task properties of Table~\ref{tab:appendix-copy-task-properties}, including their expected bottlenecks, are never used as evidence here, so the agreement between a benchmark's expected bottleneck and its measured root composition is a result rather than an assumption. A label is evidence-backed when a logged event or evaluator record localizes the error. When deterministic evidence does not determine the category, a fixed Qwen3.5-9B judge supplies an AgentRx-inspired, constraint-based causal hypothesis. Model-assisted and answer-level hypotheses remain marked separately from deterministic labels. Composition analyses normalize the 2,225 labeled source failures within each benchmark. Every root-conditioned analysis excludes LiveMath, whose 305 judge labels an independent judge does not reproduce (see below), leaving 1,920 labeled failures.

\subsubsection{Root Labeling and Evidence Tiers}
\label{app:root-labeling}
Each source failure receives one primary label from the taxonomy in Table~\ref{tab:appendix-root-taxonomy}, frozen before any matched outcome is loaded. Table~\ref{tab:root-evidence-tiers} gives the evidence behind each label. Only the 305 LiveMath labels come from the Qwen3.5-9B judge; the other 1,920 come from logged errors, evaluator records, or deterministic answer rules. Before any execution, each benchmark's expected bottleneck was annotated by hand (Table~\ref{tab:appendix-copy-task-properties}); these annotations are never used as labeling evidence.

\begin{table}[!htbp]
\centering
\caption{Evidence tiers of the 2,225 root labels.}
\label{tab:root-evidence-tiers}
\small
\setlength{\tabcolsep}{3pt}
\begin{tabularx}{\columnwidth}{Xrl}
\toprule
Evidence & Labels & Benchmarks \\
\midrule
Logged runtime or artifact error, localized to its decisive step & 585 & Spreadsheet \\
Evaluator flag, rejected action, tool error, or answer-contract violation, not step-localized & 555 & UCSB SOP 266, SQuAD 159, Amazon 113, FullWiki 10, other 7 \\
Deterministic answer rule (wrong answer $\rightarrow$ Factual; explicit refusal $\rightarrow$ Abstention) & 780 & SimpleQA \\
Qwen3.5-9B judge hypothesis & 305 & LiveMath \\
\midrule
Total & 2,225 & \\
\bottomrule
\end{tabularx}
\end{table}

\paragraph{Coverage and missingness.}
The repeat-1 common-support cohort holds 3,939 source failures and 2,225 receive labels, so 1,714 are unassigned. The missingness is concentrated rather than random, and the two Hotpot views account for the large majority of it, which is why their one and ten labeled failures cannot explain their benchmark-wide behavior. We therefore report root-conditioned quantities on the assigned subset only and provide no weighted bound for the unassigned remainder. A stratified audit of the unassigned cases, followed by inverse-probability weighting or a worst-case missingness bound, is required before any root-conditioned percentage can be read as a property of the full failure population.
Of the 1,714 unassigned failures, 1,300 are in the two Hotpot views, 214 in SQuAD, and 171 in Spreadsheet. They also respond less to the same actions: H2 rescues 5.4\% of them versus 16.8\% of the labeled failures, and a model change 19.4\% versus 30.7\%. A later pass of the same Qwen3.5-9B judge over all 1,714 proposes a label for 1,351 of them (raising coverage from 56.5\% to 90.8\%), dominated by Factual knowledge on Hotpot FullWiki (790 of 864) and answer contract on SQuAD (165) and Hotpot Context (162). A spot check found questionable FullWiki Factual labels, so these unvalidated candidates enter no reported table.

\subsubsection{Label Validation}
\label{app:label-validation}
\textbf{Independent judge.} GPT-6-sol, which is not an evaluated model, relabeled all 305 Qwen-judged LiveMath labels and all 780 rule-based SimpleQA labels, blind to the original labels. It received exactly the original requests (system prompt, evidence, output schema); only the model changed. Three deviations were necessary: the provider's default temperature was used because GPT-6-sol does not accept temperature 0; the output limit was raised because reasoning tokens count toward it; and on SimpleQA the Abstention category was also admitted for a plain ``I don't know'' (194 cases), which the original phrase filter missed. All 1,085 judgments were returned and valid. Table~\ref{tab:label-validation}a gives the agreement. On LiveMath, answer-only failures leave nothing to diagnose, so both judges default to Reasoning, and on failures with a trace the answer choices differ only in fine conditions of research-level theorem statements, so misreading the statement cannot be separated from faulty inference. We therefore exclude LiveMath from every root-conditioned analysis.

\textbf{Author verification.} An author verified a stratified random sample of 200 labels (seed 42, LiveMath excluded), with the strata and a test fixed before labeling. An LLM assistant proposed each root blind to the original label, and the author reviewed each proposal before the original label was revealed; all 200 proposals were accepted. Table~\ref{tab:label-validation}b gives the agreement. SimpleQA labels are confirmed (119 of 120; $\kappa=0.98$; the prespecified test of accuracy $\le 0.80$ gives $p=7.3\times10^{-11}$). The evidence-linked labels agree less on the exact root ($\kappa=0.33$), mostly because Execution labels are reassigned to Procedural or answer contract on Amazon SOP and Hotpot FullWiki, or judged undetermined on Spreadsheet. At the level of the intervention axis, which is what the root-aligned contrast uses, 51 of the 63 determinable evidence-linked cases (81\%) map to the same axis. Evidence-linked labels are therefore validated at the level of the intervention axis, not of the fine root. Because one author verified proposals from a single assistant, this check is not an independent multi-annotator study; a formal multi-annotator study of the decisive step remains future work.

\textbf{Propagating the label errors.} We redraw every label from the confusion matrix observed in its stratum (Dirichlet(1)-smoothed per benchmark) and recompute the root-aligned contrasts 2,000 times. Execution$\rightarrow$H2 is $+9.89$ [$+6.45,+13.23$] and Capability$\rightarrow$Model $+34.8$ [$+22.8,+46.9$], both positive in every draw (observed $+9.58$ and $+64.54$).

\begin{table}[!htbp]
\centering
\caption{Root-label validation. (a) Independent GPT-6-sol judge; $\kappa$ with 95\% task-cluster bootstrap intervals. (b) Author verification by stratum; accuracy with 95\% Wilson intervals.}
\label{tab:label-validation}
\small
\setlength{\tabcolsep}{3pt}
(a)\\[2pt]
\begin{adjustbox}{max width=\linewidth,max totalheight=0.85\textheight}
\begin{tabular}{lrrl}
\toprule
Labels & $n$ & Agreement (\%) & $\kappa$ \\
\midrule
SimpleQA (answer rule) & 780 & 89.9 & 0.77 [0.73, 0.81] \\
LiveMath (Qwen judge) & 305 & 73.1 & 0.08 [$-$0.01, 0.18] \\
\quad answer only & 125 & 90.4 & 0.00 \\
\quad with reasoning trace & 180 & 61.1 & 0.05 \\
\bottomrule
\end{tabular}
\end{adjustbox}\\[6pt]
(b)\\[2pt]
\begin{adjustbox}{max width=\linewidth,max totalheight=0.85\textheight}
\begin{tabular}{lrrl}
\toprule
Stratum (population) & $n$ & Agree & Accuracy (\%) \\
\midrule
SimpleQA answer rule (780) & 120 & 119 & 99.2 [95.4, 99.9] \\
SQuAD evidence-linked (159) & 15 & 15 & 100 [79.6, 100] \\
UCSB SOP evidence-linked (266) & 20 & 12 & 60.0 [38.7, 78.1] \\
Spreadsheet localized (585) & 20 & 7 & 35.0 [18.1, 56.7] \\
Amazon SOP evidence-linked (113) & 15 & 3 & 20.0 [7.0, 45.2] \\
Hotpot FullWiki evidence-linked (10) & 10 & 0 & 0 [0, 27.8] \\
\midrule
All & 200 & 156 & 78.0 [71.8, 83.2] \\
\bottomrule
\end{tabular}
\end{adjustbox}
\end{table}

\begin{table*}[!t]
\centering
\caption{Domain-general root-cause taxonomy. Each failed trajectory receives one primary label corresponding to its earliest decisive error; later symptoms may be retained as secondary evidence.}
\label{tab:appendix-root-taxonomy}
\small
\begin{tabularx}{\textwidth}{p{0.22\textwidth}X}
\toprule
Root cause & Operational definition \\
\midrule
Procedural & The required information is available, but the trajectory violates a mandatory step, ordering constraint, schema, or task rule. \\
Reasoning & The relevant facts or state are available, but the plan, calculation, inference, or final conclusion is incorrect. \\
Execution/tool & The intended action is plausible, but a tool call, API invocation, program, syntax, or runtime operation fails. \\
State/grounding & The trajectory acts on the wrong entity, document, cell, evidence item, or environment state. \\
Horizon/termination & No earlier decisive error is found, but the trajectory stops too early, loops, or exhausts its turn, token, or time allowance. \\
Factual knowledge/recall & The required fact is neither supplied by the task nor correctly recalled, and no earlier procedural, reasoning, execution, or grounding error explains the failure. \\
Abstention/calibration & The model explicitly declines or withholds an answer despite the task requiring one. \\
Evaluation/answer contract & The substantive answer may be available, but the submitted answer violates the evaluator-visible answer contract or cannot be credited under it. \\
\bottomrule
\end{tabularx}
\end{table*}

\subsubsection{Root-Conditioned Intervention Opportunity}

\paragraph{Failure type provides partial, but not universal, information about intervention opportunity.}
The factorial surface and interaction patterns establish that intervention value changes with the failure left by the surrounding configuration, but they do not identify that failure directly. If the residual-failure hypothesis is correct, a trajectory-level root label should provide information about intervention opportunity beyond benchmark identity. We therefore freeze each source trajectory before inspecting matched target outcomes, assign its earliest supported failure label, and test that incremental association.

\paragraph{Freeze the failure before testing its response.}
We fix the \NoSkill{}+H1+Medium source condition and assign the earliest decisive error supported by each failed trajectory before inspecting any matched target outcome. The eight labels are Procedural, Reasoning, Execution/tool, State/grounding, Horizon/termination, Factual knowledge/recall, Abstention/calibration, and Evaluation/answer contract, defined in Appendix Table~\ref{tab:appendix-root-taxonomy} and distributed over the 2,225 labeled failures in Table~\ref{tab:appendix-copy-root-composition}. That distribution is the central obstacle to interpreting roots: six of eight benchmarks are effectively dominated by one root, with UCSB SOP 78.57\% Procedural, Spreadsheet and Amazon predominantly Execution/tool, SQuAD entirely Evaluation/answer contract, and SimpleQA 74.84\% Factual with 24.78\% Abstention. Only LiveMath, SimpleQA, and UCSB SOP provide the within-benchmark overlap used below. Root and benchmark are therefore strongly associated by construction, and the Reasoning, State, Factual, and Abstention assignments are model-assisted hypotheses rather than deterministic labels.

\paragraph{Link each source root to matched intervention opportunity.}
For each labeled failure we ask whether changing one axis improves that same task instance. Table~\ref{tab:root-intervention-opportunity} averages candidate-specific positive-change indicators over the three Skill and three Model alternatives, with H2 and High Budget each testing one action. Contract-labeled SQuAD failures respond to H2 in 77.3\% of cases, and SimpleQA's factual-recall failures respond to a Model change in 30.1\% against 6.7\% for H2. The Model column is the largest entry in six of the seven rows, including Procedural, where the Skill average is 12.4\% against 23.1\% for Model; among actions that keep the model fixed, however, Skill is the largest for Procedural failures (12.4\% against 6.7\% for H2). A root label can thus suggest an axis without identifying the best action.

\begin{table*}[!t]
\centering
\caption{Pooled descriptive root cause and action-count-standardized matched intervention opportunity. Skill and Model average candidate-specific improvement indicators over three actions; H2 and High budget each test one action. Columns are overlapping improvement opportunities rather than mutually exclusive outcomes, so rows need not sum to 100\%. Improvement is a strictly positive matched native-score change ($\delta_b=0$), preserving the original estimand. Each axis has total weight one despite different candidate counts. Bold marks the largest standardized opportunity in each row; underlining marks the second largest. $^{\dagger}$The single Horizon case is descriptive. LiveMath is excluded (1,920 labeled failures).}
\label{tab:root-intervention-opportunity}
\scriptsize
\resizebox{\textwidth}{!}{%
\begin{tabular}{lrrrrr}
\toprule
H1 root cause & Failures & Skill avg. & H2 & High budget & Model avg. \\
\midrule
Procedural & 209 & \underline{12.4\%} & 6.7\% & 3.3\% & \textbf{23.1\%} \\
Execution/tool & 730 & 5.0\% & \underline{20.8\%} & 4.0\% & \textbf{26.6\%} \\
State/grounding & 37 & \underline{41.4\%} & 8.1\% & 10.8\% & \textbf{73.0\%} \\
Horizon/termination$^{\dagger}$ & 1 & \underline{33.3\%} & 0.0\% & 0.0\% & \textbf{100.0\%} \\
Factual knowledge/recall & 586 & \underline{7.8\%} & 6.7\% & 6.3\% & \textbf{30.1\%} \\
Abstention/calibration & 194 & \underline{2.7\%} & 2.1\% & 2.6\% & \textbf{16.2\%} \\
Evaluation/answer contract & 163 & 63.0\% & \textbf{77.3\%} & 1.8\% & \underline{76.7\%} \\
\bottomrule
\end{tabular}}

\end{table*}

Because root and benchmark are strongly associated, the pooled rows alone cannot show that the label adds information beyond benchmark membership. We therefore use the within-benchmark overlap of SimpleQA and UCSB SOP (1,046 failures; LiveMath excluded). A benchmark- and source-model-stratified permutation test finds a root $\times$ intervention association ($p=0.0002$). In grouped out-of-fold prediction, adding Root to Benchmark+Model reduces log loss by 0.0096 (95\% CI [+0.0020,+0.0170]) and improves AUPRC by 6.99 points ([+3.48,+10.72]), but not AUROC (+0.81 [$-0.67$,+2.29]). Failure labels therefore carry information beyond benchmark identity where several roots occur within a benchmark, but they do not yet predict, independently of the benchmark, which intervention will help.

A separate outcome-blind root-to-action mapping gives the same qualified picture. H2 is comparatively responsive for Execution/tool, State/grounding, and Evaluation/answer contract failures (+9.58 points over the non-aligned mean, 95\% CI [+4.77,+13.93]), and a Model change for Factual knowledge/recall and Abstention/calibration failures (+64.54 [+58.68,+70.16], SimpleQA). Among actions that keep the model fixed, the mapped action is the strongest in every row, including Procedural$\rightarrow$Skill (one benchmark); a stronger model still rescues more (Appendix~\ref{app:root-aligned-contrast}).

\subsubsection{Frozen Root-Aligned Intervention Contrast}
\label{app:root-aligned-contrast}

\begin{table}[!htbp]
\centering
\caption{Execution failures (428, six benchmarks, LiveMath excluded): the mapped H2 action against the best action that keeps Qwen3.5-9B (rerun, Skill, H2, or High budget) and against the stronger model. Success rates in \%; 95\% paired task-cluster bootstrap intervals.}
\label{tab:root-aligned-same-model}
\small
\setlength{\tabcolsep}{4pt}
\begin{adjustbox}{max width=\linewidth,max totalheight=0.85\textheight}
\begin{tabular}{rrlrl}
\toprule
H2 & Best same-model (Skill) & H2 $-$ best same-model & Stronger model & H2 $-$ model \\
\midrule
33.50 & 18.41 & $+15.09$ [$+7.98,+21.48$] & 48.18 & $-14.67$ [$-25.48,-6.62$] \\
\bottomrule
\end{tabular}
\end{adjustbox}
\end{table}

\begin{figure*}[!tp]
  \centering
  \includegraphics[width=\textwidth,height=0.85\textheight,keepaspectratio]{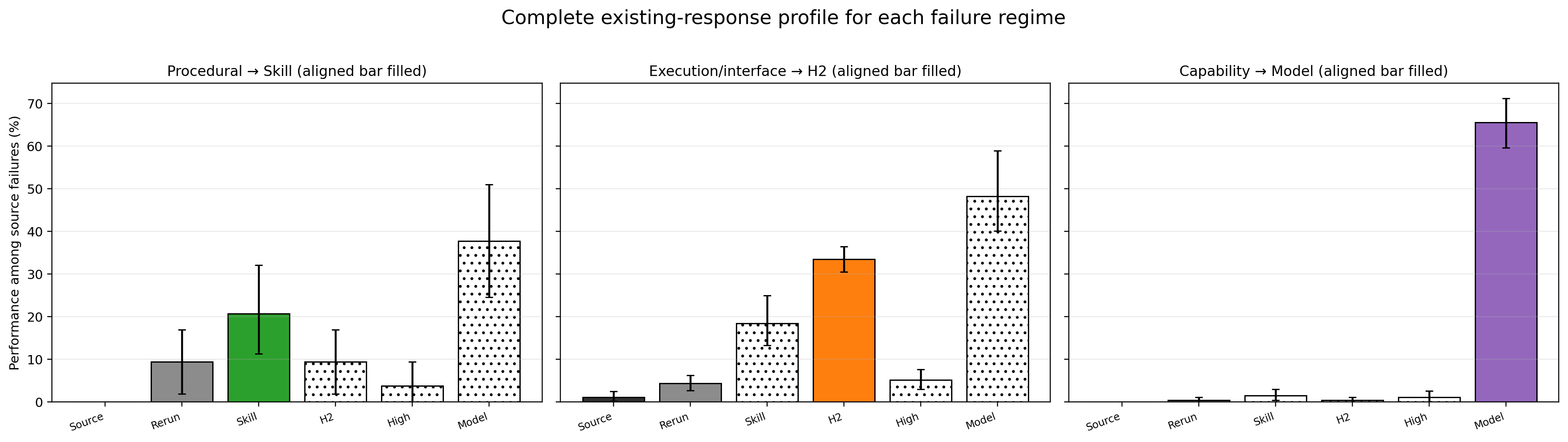}
  \caption{Complete action profile behind the compact root-aligned contrast, without LiveMath. Every panel shows the failed source, same-configuration rerun, Trace2Skill, H2, High budget, and GPT-5.6-sol target performance. The outlined bar is the action fixed by the semantic mapping; a high non-aligned mean can therefore be traced to its individual action.}
  \label{fig:appendix-root-aligned-all-actions}
\end{figure*}

The outcome-blind mapping fixes Procedural$\rightarrow$Skill, Execution/tool/State/answer contract$\rightarrow$H2, and Reasoning/Factual knowledge/Abstention$\rightarrow$Model before loading matched target outcomes. Figure~\ref{fig:aligned} and Table~\ref{tab:root-aligned-compact} show the compact contrast on 748 Qwen \NoSkill{}+H1+Medium repeat-1 failures with complete targets (LiveMath excluded). The H2 and Model mappings exceed the non-aligned three-axis mean; Procedural$\rightarrow$Skill does not, because that mean includes a stronger model, but Skill is the strongest action that keeps the model fixed. The H2 mapping does not exceed the stronger model, and capability now rests on SimpleQA alone. Figure~\ref{fig:appendix-root-aligned-all-actions} exposes each action. The reconstruction uses existing target executions and is descriptive rather than prospective randomized intervention.

\begin{table*}[!t]
\centering
\caption{Compact root-aligned intervention contrast. $n/B$ gives source failures and contributing benchmarks. The non-aligned column averages the other three active intervention axes. Horizon/termination is omitted because its single source case cannot support a pattern-level preset. Source: Qwen3.5-9B + \NoSkill{} + H1 + Medium, repeat 1. Same-configuration rerun is kept separate from the three-axis non-aligned mean. High remains an action-level comparison but is not assigned a root pattern.}
\label{tab:root-aligned-compact}
\scriptsize
\setlength{\tabcolsep}{4pt}
\resizebox{\textwidth}{!}{%
\begin{tabular}{lrrrrr}
\toprule
Failure pattern & $n/B$ & Source & Non-aln. & Root-aligned action & Aligned$-$mean [95\% CI] \\
\midrule
Proc.$\rightarrow$Skill & 53/1 & 0.00 & 16.98 & Skill: 20.75 & +3.77 [$-5.66$,+14.47] \\
Exec./interface$\rightarrow$H2 & 428/6 & 1.15 & 23.92 & H2: 33.50 & +9.58 [+4.77,+13.93] \\
Capability$\rightarrow$Model & 267/1 & 0.00 & 1.00 & GPT-5.6-sol: 65.54 & +64.54 [+58.68,+70.16] \\
\bottomrule
\end{tabular}}

\end{table*}

\begin{table*}[!t]
\centering
\caption{Complete performance table for the frozen root-aligned intervention contrast. Scores are paired within task and benchmark-equal aggregated. ``Strongest other'' is reported to distinguish a taxonomy-aligned signal from empirical action optimality. Performance values are percentages; blue bold values mark the intervention fixed by the root-to-axis mapping. $B$ is the number of contributing benchmarks. LiveMath is excluded. Bootstrap $p$ values (5,000 draws) are .4560, .0000, and .0000 in row order; the corresponding BH $q$ values are shown in the final column. Horizon/termination is excluded because $n=1$ cannot support a pattern-level preset.}
\label{tab:appendix-root-aligned-full}
\scriptsize
\setlength{\tabcolsep}{4.5pt}
\textbf{(a) Per-action performance}\\[2pt]
\resizebox{\textwidth}{!}{%
\begin{tabular}{lrrrrrrrr}
\toprule
Failure pattern & $n$ & $B$ & Source & Rerun & Skill & H2 & High & Model \\
\midrule
Proc.$\rightarrow$Skill & 53 & 1 & 0.00 & 9.43 & \gain{20.75} & 9.43 & 3.77 & 37.74 \\
Exec./interface$\rightarrow$H2 & 428 & 6 & 1.15 & 4.34 & 18.41 & \gain{33.50} & 5.17 & 48.18 \\
Capability$\rightarrow$Model & 267 & 1 & 0.00 & 0.37 & 1.50 & 0.37 & 1.12 & \gain{65.54} \\
\bottomrule
\end{tabular}}

\vspace{6pt}
\textbf{(b) Aligned-versus-alternative contrasts}\\[2pt]
\resizebox{\textwidth}{!}{%
\begin{tabular}{lrrlrl}
\toprule
Failure pattern & Other mean & Aligned$-$mean [95\% CI] & Strongest other & Aligned$-$strongest & BH $q$ \\
\midrule
Proc.$\rightarrow$Skill & 16.98 & +3.77 [$-5.66$,+14.47] & Model 37.74 & $-16.98$ & .4560 \\
Exec./interface$\rightarrow$H2 & 23.92 & +9.58 [+4.77,+13.93] & Model 48.18 & $-14.67$ & .0000 \\
Capability$\rightarrow$Model & 1.00 & +64.54 [+58.68,+70.16] & Skill 1.50 & +64.04 & .0000 \\
\bottomrule
\end{tabular}}

\end{table*}

\begin{table*}[!t]
\centering
\caption{Available cost fields for the frozen root-aligned contrast. Missing values mean that comparable full-chain telemetry was unavailable; they are not treated as zero. $^{a}$A reproducible \$1 per million accounted-token scenario, not an invoice. $^{b}$Actual provider attribution is available only when GPT-5.6-sol is the aligned target; it excludes the local-Qwen source cost. Qwen latency and actual dollar cost were not logged comparably.}
\label{tab:appendix-root-aligned-cost}
\scriptsize
\setlength{\tabcolsep}{4pt}
\begin{adjustbox}{max width=\linewidth,max totalheight=0.85\textheight}
\begin{tabular}{lrrrr}
\toprule
Failure pattern & Aligned tokens & Other-axis tokens & Aligned USD$^{a}$ & Provider USD / latency$^{b}$ \\
\midrule
Proc.$\rightarrow$Skill & NA & NA & NA & NA \\
Exec./interface$\rightarrow$H2 & 21,966.28 & 19,107.73 & \$0.021966 & NA \\
Capability$\rightarrow$Model & 4,380.55 & 6,903.82 & \$0.004381 & \$0.017542 / 15.80 s \\
\bottomrule
\end{tabular}
\end{adjustbox}

\end{table*}

\subsection{When Feedback Becomes Recovery}
\label{app:rq4-details}
\label{app:feedback-recovery-details}

\begin{figure}[!htbp]
  \centering
  \includegraphics[width=0.96\textwidth,height=0.85\textheight,keepaspectratio]{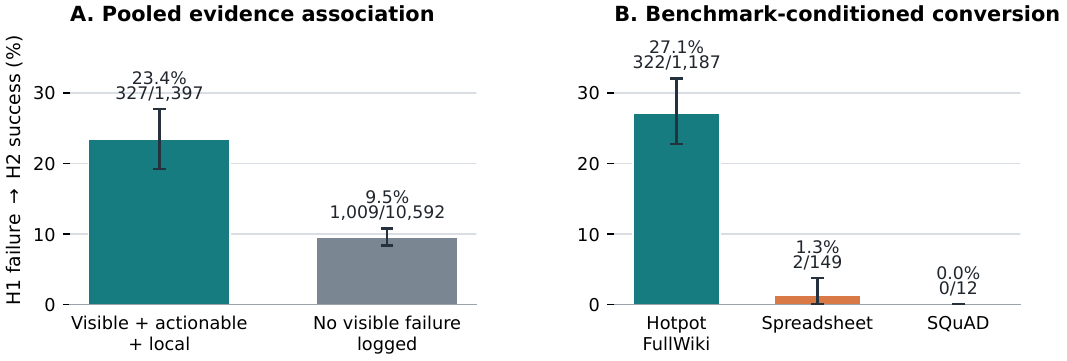}
  \caption{H1$\rightarrow$H2 rescue conditional on evidence recorded in the failed H1 trajectory. (a) Pooled binary failures with visible, actionable, locally correctable evidence have a higher rescue rate than failures with no visible signal logged. (b) The same evidence pattern converts very differently across benchmarks, and Hotpot FullWiki supplies 85.0\% of the pooled exposed-failure denominator. Whiskers are 95\% instance-UID cluster-bootstrap intervals; the interval is omitted for SQuAD because none of its 12 cases is rescued. ``No visible failure logged'' describes available telemetry rather than proving that no usable signal existed.}
  \label{fig:appendix-copy-recovery-evidence-conversion}
\end{figure}

\paragraph{Usable evidence is associated with recovery, but logged observability is not sufficient.}
H1$\rightarrow$H2 provides a concrete test of when feedback becomes repair. Mechanistically, H2 can help only if the failure is exposed to the model, the resulting signal identifies a legal corrective action, and the error can be repaired locally without restarting the whole solution:
\[
\boxed{\text{exposure} \;\longrightarrow\; \text{repair} \;\longrightarrow\; \text{preservation}.}
\]
We operationalize the first two stages using whether an H1 failure is logged as model-visible, actionable, and locally correctable. These telemetry fields are imperfect proxies: absence of a logged visible failure does not establish that no usable signal existed.

Among H1 failures satisfying all three recorded conditions, H2 rescues 23.4\% (327/1,397), compared with 9.5\% (1,009/10,592) when no visible failure is logged. The pooled difference is consistent with the observability hypothesis, but it is dominated by one benchmark. Hotpot FullWiki supplies 1,187 of the 1,397 exposed failures (85.0\% of the denominator) and rescues 27.1\% of them, whereas Spreadsheet rescues only 2 of 149 (1.3\%) and SQuAD 0 of 12 (Figure~\ref{fig:appendix-copy-recovery-evidence-conversion}). The pooled 23.4\% is therefore largely a weighted restatement of the retrieval case rather than evidence of a portable exposure-to-repair mechanism. Logged observability is associated with recovery, but it is not sufficient; its necessity also remains unproven because the logs may miss usable evidence.

\paragraph{Rescue alone is insufficient.}
A recovery intervention must repair failed executions without destroying executions that were already correct. We therefore report both
\[
\operatorname{RescueRate}=P(Y^{H2}=1\mid Y^{H1}=0)
\quad\text{and}\quad
\operatorname{RegressionRate}=P(Y^{H2}=0\mid Y^{H1}=1),
\]
with
\[
\operatorname{NetGain}=P(Y^{H1}=0,Y^{H2}=1)-P(Y^{H1}=1,Y^{H2}=0)
\]
over the full matched population. A successful repair that is overwritten later in the trajectory fails the preservation stage. H2 produces both rescues and recovery-induced regressions, so its utility is determined by the complete exposure--repair--preservation pathway rather than rescue alone (Figure~\ref{fig:appendix-copy-recovery-evidence-conversion} and Table~\ref{tab:appendix-h1-h2-transitions}).

A retrospective Qwen pre-H2 predictor attains AUROC 0.945 against 0.796 for benchmark identity, but this does not establish a deployable recoverability signal. Its features aggregate repeated H0 and H1 history for the same task instance while its label is whether H2 rescues that instance, so it has partial access to how the instance behaves under repetition and the gap to a single-trajectory predictor is unmeasured. It is an upper bound contaminated by task-instance identity information (Figure~\ref{fig:pre-recovery-prediction}).

For matched H1$\rightarrow$H2 transitions, let $Z=1$ indicate audited model-visible feedback/recovery exposure. For binary outcomes,
\begin{align}
\mu(H2)-\mu(H1)
&=\Pr(Y_1=0,Y_2=1,Z=1) \\
&\quad+\Pr(Y_1=0,Y_2=1,Z=0)-\Pr(Y_1=1,Y_2=0),
\label{eq:transition-decomposition}
\end{align}
corresponding to exposed gain, unexposed gain, and recovery-induced regression.

\begin{table*}[!t]
\centering
\caption{Action-set Oracle opportunity on the same 1,920 repeat-1 source failures (LiveMath excluded) as Table~\ref{tab:root-intervention-opportunity}. Skill and Model count a source as improved when any of three tested actions improves it; H2 and High budget each test one action. Candidate counts differ, so columns quantify action-set headroom and must not be used to rank intervention axes. A source counts as improved when any tested action on that axis improves it; Skill and Model each test three actions while H2 and High budget test one. These are overlapping retrospective action-set opportunities, not causal effects or router accuracy. $^{\dagger}$The single Horizon case is descriptive.}
\label{tab:appendix-root-action-set-oracle}
\scriptsize
\setlength{\tabcolsep}{5pt}
\begin{adjustbox}{max width=\linewidth,max totalheight=0.85\textheight}
\begin{tabular}{lrrrrr}
\toprule
H1 root cause & Failures & Skill (any) & H2 & High budget & Model (any) \\
\midrule
Procedural & 209 & 20.6\% (43) & 6.7\% (14) & 3.3\% (7) & 40.7\% (85) \\
Execution/tool & 730 & 11.2\% (82) & 20.8\% (152) & 4.0\% (29) & 47.0\% (343) \\
State/grounding & 37 & 64.9\% (24) & 8.1\% (3) & 10.8\% (4) & 89.2\% (33) \\
Horizon/termination$^{\dagger}$ & 1 & 1/1 & 0/1 & 0/1 & 1/1 \\
Factual knowledge/recall & 586 & 14.0\% (82) & 6.7\% (39) & 6.3\% (37) & 58.7\% (344) \\
Abstention/calibration & 194 & 6.7\% (13) & 2.1\% (4) & 2.6\% (5) & 40.2\% (78) \\
Evaluation/answer contract & 163 & 81.6\% (133) & 77.3\% (126) & 1.8\% (3) & 91.4\% (149) \\
\bottomrule
\end{tabular}
\end{adjustbox}

\end{table*}

\begin{table*}[!t]
\centering
\caption{Matched H1$\rightarrow$H2 transitions at Default budget for binary-outcome benchmarks. Rescue and regression rates condition on H1 failure and H1 success, respectively; Net gain is $(N_{0\rightarrow1}-N_{1\rightarrow0})/N$. Amazon uses a fractional metric and is not dichotomized: among 2,400 matched executions, 1,070 improve, 1,322 are unchanged, eight degrade, and mean performance rises from 40.47 to 82.71 (+42.24 points).}
\label{tab:appendix-h1-h2-transitions}
\scriptsize
\setlength{\tabcolsep}{4pt}
\begin{adjustbox}{max width=\linewidth,max totalheight=0.85\textheight}
\begin{tabular}{lrrrrrrrr}
\toprule
Benchmark & $N$ & Rescue & Preserve & Regr. & Persist. & Rescue \% & Regr. \% & Net \\
\midrule
Spreadsheet & 13,440 & 1,046 & 3,514 & 401 & 8,479 & 10.98\% & 10.24\% & +4.80 pp \\
LiveMath & 5,952 & 410 & 2,050 & 442 & 3,050 & 11.85\% & 17.74\% & $-0.54$ pp \\
UCSB SOP & 8,400 & 265 & 5,143 & 249 & 2,743 & 8.81\% & 4.62\% & +0.19 pp \\
SimpleQA & 14,400 & 365 & 4,366 & 396 & 9,273 & 3.79\% & 8.32\% & $-0.22$ pp \\
SQuAD & 14,400 & 736 & 10,828 & 140 & 2,696 & 21.45\% & 1.28\% & +4.14 pp \\
Hotpot Context & 14,400 & 294 & 8,201 & 280 & 5,625 & 4.97\% & 3.30\% & +0.10 pp \\
Hotpot FullWiki & 14,400 & 770 & 3,466 & 432 & 9,732 & 7.33\% & 11.08\% & +2.35 pp \\
\bottomrule
\end{tabular}
\end{adjustbox}

\end{table*}

\paragraph{Interpretation boundary.}
The taxonomy localizes the earliest supported source error, whereas the matched target outcomes indicate whether a tested intervention subsequently improves that case. The association does not establish that the named root cause is the sole causal mediator, and labels produced from answer-only or model-assisted evidence remain hypotheses. The transition analysis likewise describes replayed matched executions; it does not constitute prospective continuation from one shared H1 state.

\subsection{What Capability Boundaries Remain}
\label{app:rq5-details}
\label{app:capability-definition}

For failed task instance $i$ under source configuration $c$, let $\mathcal A_{i,J}$ be the available actions on axis $J\in\{S,H,B,M\}$ and $\tau_i$ the benchmark-native success criterion. To equalize candidate count, one action is selected uniformly per axis and the crossing set is
\begin{equation}
  a_{i,J}\sim\operatorname{Unif}(\mathcal A_{i,J}),
  \qquad
  \mathcal B_i(\mathbf a_i)=\left\{J:Y_i(c,a_{i,J})\geq\tau_i\right\}.
  \label{eq:capability-boundary}
\end{equation}
We enumerate all $3\times3$ Skill--Model pairs exactly; H2 and High budget are singleton actions. Reported boundary prevalence is the uniform expectation over these one-action-per-axis comparisons. If $\mathcal B_i$ is empty, no selected intervention reaches success; if several axes cross, deployment cost is required to identify the cheapest successful action in the full action set. The boundary is therefore empirical and configuration-dependent rather than an intrinsic hierarchy among intervention classes.

\section{Skill--Harness Regimes, Recovery Pathways, and Pre-Recovery Diagnosis}
\label{app:skill-harness-details}

\begingroup
\paragraph{Regime prevalence over the full matrix.} Table~\ref{tab:regime-prevalence} classifies every Default-budget skill--model--benchmark combination (8 benchmarks $\times$ 4 models $\times$ 3 skills $=96$) by its skill gain over \NoSkill{} at H0 ($g_0$) and at H2 ($g_2$), using the five-point descriptive tolerance $\epsilon$ of Appendix~\ref{app:validity} on the repeat-averaged cells of Table~\ref{tab:default-matrix-transposed}. \emph{Absorbed}: $g_0\ge\epsilon$, $g_2\le g_0-\epsilon$, and $g_2>-\epsilon$. \emph{Reversed}: $g_0\ge\epsilon$ and $g_2\le-\epsilon$. \emph{Preserved}: $g_0\ge\epsilon$ and $|g_2-g_0|<\epsilon$. \emph{Rescued}: $g_0\le-\epsilon$, $g_2-g_0\ge\epsilon$, and $g_2>-\epsilon$. \emph{Persistent harm}: $g_0\le-\epsilon$ and not rescued. \emph{Emerging gain} or \emph{harm}: $|g_0|<\epsilon$ and $g_2\ge\epsilon$ or $g_2\le-\epsilon$. \emph{No material effect}: $|g_0|<\epsilon$ and $|g_2|<\epsilon$. No combination is amplified ($g_0\ge\epsilon$, $g_2\ge g_0+\epsilon$). The six representative cells of Table~\ref{tab:regimes} receive the regime named there. All 18 absorbed combinations lie on SQuAD, Hotpot Context, and Amazon SOP, the three benchmarks with native-score floor cells (Appendix~\ref{sec:measurement-floors}). Of the four rescued combinations, the two of Qwen3.5-9B remove penalties of 19.5 and 22.8 points, whereas the two of Sonnet~4.6 remove penalties of 5.6 points. These counts are thresholded descriptions of cell means, not effects certified against A/A noise.

\begin{table}[!htbp]
\centering
\caption{Skill--harness regime of every Default-budget skill--model--benchmark combination (counts; H0 versus H2, $\epsilon=5$ points).}
\label{tab:regime-prevalence}
\small
\begin{tabular}{lrrrrr}
\toprule
Regime & Qwen3.5-9B & Sonnet 4.6 & Sonnet 5 & GPT-5.6-sol & All \\
\midrule
Absorbed & 4 & 4 & 6 & 4 & 18 \\
Preserved & 5 & 2 & 1 & 4 & 12 \\
Rescued & 2 & 2 & 0 & 0 & 4 \\
Persistent harm & 3 & 3 & 3 & 2 & 11 \\
Reversed & 0 & 1 & 0 & 0 & 1 \\
Emerging harm & 0 & 1 & 1 & 0 & 2 \\
Emerging gain & 1 & 0 & 1 & 1 & 3 \\
No material effect & 9 & 11 & 12 & 13 & 45 \\
\midrule
Total & 24 & 24 & 24 & 24 & 96 \\
\bottomrule
\end{tabular}
\end{table}
\endgroup

\paragraph{Answer format versus content.} Table~\ref{tab:qwen-format-content} re-scores the Qwen3.5-9B trajectories on SQuAD and Hotpot Context with a check that credits a response of at most 100 words containing a gold answer. On SQuAD the content check is near 95 in every cell, so the native Harness and Skill effects there are almost entirely answer format. On Hotpot Context, the \SkillOpt{} penalty at H1 and H2 remains under the content check ($-14.5$ and $-11.8$ against \NoSkill{}), so that penalty is substantive.

\begin{table}[!htbp]
\centering
\caption{Qwen3.5-9B native exact match versus the answer-content check on the same saved trajectories (0--100 scale, Default budget).}
\label{tab:qwen-format-content}
\scriptsize
\begin{adjustbox}{max width=\linewidth,max totalheight=0.85\textheight}
\begin{tabular}{llrrrrrrrr}
\toprule
 & & \multicolumn{4}{c}{Native exact match} & \multicolumn{4}{c}{Answer content present} \\
\cmidrule(lr){3-6}\cmidrule(lr){7-10}
Benchmark & Skill & H0 & H1 & H2 & H3 & H0 & H1 & H2 & H3 \\
\midrule
SQuAD & \NoSkill{} & 10.7 & 36.6 & 78.6 & 79.3 & 96.1 & 95.7 & 95.0 & 95.3 \\
 & \Direct{} & 46.7 & 76.4 & 75.9 & 76.3 & 95.2 & 96.4 & 96.8 & 96.2 \\
 & \SkillOpt{} & 29.0 & 77.0 & 77.6 & 77.4 & 95.3 & 95.0 & 95.0 & 95.0 \\
 & \TraceToSkill{} & 76.9 & 55.0 & 80.8 & 79.2 & 96.6 & 96.3 & 95.3 & 94.7 \\
\midrule
Hotpot Context & \NoSkill{} & 18.6 & 59.3 & 59.7 & 58.6 & 74.0 & 66.1 & 66.4 & 65.8 \\
 & \Direct{} & 21.3 & 55.2 & 55.3 & 57.3 & 68.8 & 64.6 & 64.7 & 66.9 \\
 & \SkillOpt{} & 0.0 & 44.0 & 47.9 & 59.7 & 70.0 & 51.6 & 54.6 & 69.4 \\
 & \TraceToSkill{} & 26.2 & 58.4 & 58.6 & 59.2 & 73.7 & 64.3 & 64.3 & 65.3 \\
\bottomrule
\end{tabular}
\end{adjustbox}
\end{table}

\paragraph{Cross-model regimes.} Table~\ref{tab:cross-model-regimes} reports the same regime cells for all four models, computed from Table~\ref{tab:default-matrix-transposed}.

\begin{table}[!htbp]
\centering
\caption{Skill gain over \NoSkill{} within each Harness (points) for the regime cells of Table~\ref{tab:regimes}, across the four Models at Default budget. Values are differences of one-decimal cell means.}
\label{tab:cross-model-regimes}
\scriptsize
\setlength{\tabcolsep}{2pt}
\resizebox{\linewidth}{!}{%
\begin{tabular}{lll*{16}{r}}
\toprule
 & & & \multicolumn{4}{c}{Qwen3.5-9B} & \multicolumn{4}{c}{Sonnet 4.6} & \multicolumn{4}{c}{Sonnet 5} & \multicolumn{4}{c}{GPT-5.6-sol} \\
\cmidrule(lr){4-7}\cmidrule(lr){8-11}\cmidrule(lr){12-15}\cmidrule(lr){16-19}
Regime & Benchmark & Skill & H0 & H1 & H2 & H3 & H0 & H1 & H2 & H3 & H0 & H1 & H2 & H3 & H0 & H1 & H2 & H3 \\
\midrule
Absorption & SQuAD & \TraceToSkill{} & 66.2 & 18.4 & 2.2 & -0.1 & 16.1 & 1.3 & 1.3 & 0.9 & 67.5 & 1.0 & 1.6 & 1.3 & 48.9 & 17.2 & 17.7 & 12.8 \\
 & SQuAD & \Direct{} & 36.0 & 39.8 & -2.7 & -3.0 & 42.4 & -5.0 & -5.3 & -4.3 & 35.8 & -0.8 & -0.7 & -2.0 & 35.4 & 10.2 & 11.3 & 8.9 \\
 & SQuAD & \SkillOpt{} & 18.3 & 40.4 & -1.0 & -1.9 & 0.0 & -3.0 & -3.5 & -3.7 & 48.5 & -2.9 & -2.1 & -4.7 & 25.8 & 0.4 & 1.0 & -1.7 \\
Penalty rescue & FullWiki & \Direct{} & -22.8 & -22.0 & 1.6 & 2.9 & -2.4 & -4.3 & -2.6 & -3.1 & -3.9 & -4.2 & -3.5 & -5.8 & 6.2 & 4.6 & 4.1 & 6.9 \\
 & FullWiki & \SkillOpt{} & -3.1 & -1.9 & 8.1 & 8.7 & -9.3 & -9.7 & -8.3 & -7.2 & -11.0 & -6.9 & -7.7 & -8.8 & -5.8 & -2.6 & -6.0 & -3.1 \\
 & Amazon SOP & \SkillOpt{} & -19.5 & -27.2 & 6.2 & 3.1 & 17.3 & 17.3 & 0.0 & 0.4 & -0.9 & 4.8 & 2.6 & -7.8 & 0.6 & 1.4 & 0.0 & 0.3 \\
 & Context & \SkillOpt{} & -18.6 & -15.3 & -11.8 & 1.1 & 0.0 & 2.0 & 2.2 & 0.8 & -3.4 & -0.4 & -1.6 & -0.9 & -22.6 & -48.4 & -49.7 & -49.1 \\
Preservation & FullWiki & \TraceToSkill{} & 8.5 & 8.2 & 9.2 & 12.2 & 14.9 & 11.2 & 12.4 & 11.6 & 10.0 & 14.1 & 11.9 & 8.7 & 10.6 & 7.5 & 9.5 & 9.0 \\
 & UCSB SOP & \SkillOpt{} & 13.5 & 15.3 & 13.3 & 3.7 & 5.7 & 7.0 & 5.7 & 4.9 & 1.9 & 1.9 & 2.6 & 3.4 & 2.7 & 3.8 & 1.3 & 1.9 \\
 & UCSB SOP & \TraceToSkill{} & 9.3 & 10.9 & 9.3 & 2.9 & -2.8 & -2.7 & 0.4 & 2.7 & -1.6 & 0.2 & -1.7 & -0.2 & 1.9 & 1.0 & 0.9 & -0.2 \\
\bottomrule
\end{tabular}}
\end{table}

\paragraph{Recovery pathway.} Table~\ref{tab:pathway-diagnostics} reports the Qwen3.5-9B H2 telemetry behind Section~\ref{sec:which}. Exposure is the share of executions with an audited model-visible recovery signal; each later rate uses the preceding stage as its denominator, and a dash denotes an unreached stage.

\begin{table}[t]
\caption{H2 failure-to-recovery diagnostics, pooled over three repeats (\%). The table distinguishes exposure, response, local correction, and final task recovery rather than treating them as a single repair event.}
\label{tab:pathway-diagnostics}
\centering
\scriptsize
\resizebox{\linewidth}{!}{%
\begin{tabular}{llrrrrr}
\toprule
Benchmark & Skill & Exposure & Response$\mid$exposure & Local repair$\mid$response & Success$\mid$response & H2$-$H1 \\
\midrule
SQuAD & \NoSkill{} & 52.9 & 100.0 & 100.0 & 78.2 & +42.0 \\
 & \Direct{} & 0.3 & 100.0 & 0.0 & 0.0 & -0.6 \\
 & \SkillOpt{} & 0.0 & -- & -- & -- & +0.6 \\
 & \TraceToSkill{} & 32.7 & 100.0 & 99.3 & 72.4 & +25.8 \\
\midrule
BigCode & \NoSkill{} & 69.8 & 100.0 & 0.0 & 0.0 & +0.8 \\
 & \Direct{} & 73.6 & 100.0 & 1.1 & 1.1 & +0.8 \\
 & \SkillOpt{} & 71.3 & 100.0 & 1.1 & 1.1 & +1.6 \\
 & \TraceToSkill{} & 76.0 & 100.0 & 2.0 & 2.0 & +1.6 \\
\midrule
Spreadsheet & \NoSkill{} & 73.2 & 100.0 & 65.2 & 22.3 & +15.6 \\
 & \Direct{} & 67.3 & 100.0 & 66.7 & 23.4 & +15.2 \\
 & \SkillOpt{} & 77.0 & 100.0 & 65.1 & 19.3 & +13.2 \\
 & \TraceToSkill{} & 69.0 & 100.0 & 65.5 & 17.9 & +12.6 \\
\bottomrule
\end{tabular}
}
\end{table}

\paragraph{Signal coding.} Each Qwen3.5-9B H1 failure is coded from its own H1 trajectory, before and independently of the H2 outcome: an interface or contract violation (missing answer tag, premature or rejected final submission), a runtime or tool error (execution error, invalid action, or tool-parameter error), or no error signal (a wrong answer that raises nothing). Rescue is H2 success in the same instance, Skill, and repeat.

\begin{table}[!htbp]
\caption{H2 rescue rate (\%) of H1 failures by the signal in the H1 trajectory. The last column is the within-benchmark difference between failures with and without a signal, with a 95\% task-cluster bootstrap interval.}
\label{tab:signal-rescue}
\centering
\scriptsize
\begin{adjustbox}{max width=\linewidth,max totalheight=0.85\textheight}
\begin{tabular}{lrrrrl}
\toprule
 & \multicolumn{2}{c}{With a signal} & \multicolumn{2}{c}{No signal} & \\
\cmidrule(lr){2-3}\cmidrule(lr){4-5}
Benchmark & $n$ & Rescued & $n$ & Rescued & Difference [95\% CI] \\
\midrule
SQuAD & 798 & 76.9 & 597 & 1.3 & $+75.6$ [$+69.0$, $+82.1$] \\
Hotpot Context & 52 & 63.5 & 1,595 & 1.8 & $+61.6$ [$+39.6$, $+82.2$] \\
Hotpot FullWiki & 1,430 & 24.2 & 1,379 & 1.5 & $+22.7$ [$+18.5$, $+27.0$] \\
Spreadsheet & 2,362 & 21.5 & 645 & 17.4 & $+4.1$ [$+0.0$, $+8.1$] \\
UCSB SOP & 25 & 4.0 & 832 & 10.6 & $-6.6$ [$-13.0$, $+3.3$] \\
BigCode & 377 & 3.2 & -- & -- & -- \\
\bottomrule
\end{tabular}
\end{adjustbox}
\end{table}

\begin{figure}[!htbp]
\centering
\includegraphics[width=0.9\linewidth,height=0.85\textheight,keepaspectratio]{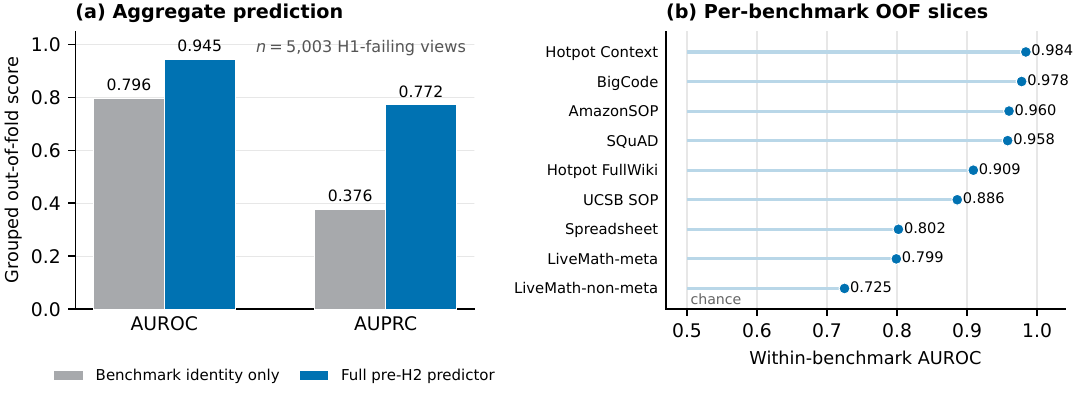}
\caption{Grouped out-of-fold prediction of H2 rescue among Qwen3.5-9B H1 failures. (a) Pre-H2 behavior outperforms benchmark identity. (b) Within-benchmark AUROC.}
\label{fig:pre-recovery-prediction}
\end{figure}

\paragraph{Pre-recovery diagnosis.}
\label{app:pre-recovery}
Among the 5,003 Qwen3.5-9B instance--Skill views that fail at H1 in a majority of the three repeats, 14.1\% succeed at H2 in a majority. A gradient-boosted classifier on H0 and H1 behavior (outcome summaries, token and turn counts, tool use, premature submission, and the modal H1 failure type), evaluated by five-fold cross-validation grouped by instance, ranks them with AUROC 0.945 and AUPRC 0.772, against 0.796 and 0.376 for benchmark identity alone (Figure~\ref{fig:pre-recovery-prediction}; Table~\ref{tab:recovery-prediction}). Table~\ref{tab:escalation-replay} replays escalation policies over all 7,488 views. Probe-first policies run H1, observe its outcome, and escalate only failures, so their cost includes the probe; regressions are H1 successes lost at H2 per 100 views, and tokens cover the benchmarks with token telemetry. The predictor scores in the replay come from a re-run of the same pipeline (AUROC 0.944, AUPRC 0.756). Because the replay reuses the repeats that define the label, its figures are retrospective upper bounds.

\begin{table}[t]
\caption{Grouped out-of-fold H2-rescuability prediction. Base rate is the positive-label prevalence. Per-benchmark rows use the full gradient-boosted predictor; SimpleQA has no positive cases, so ranking metrics are undefined.}
\label{tab:recovery-prediction}
\centering
\small
\resizebox{\linewidth}{!}{%
\begin{tabular}{llrrrr}
\toprule
Scope & Predictor & $n$ & Base rate & AUROC & AUPRC \\
\midrule
All & Full pre-H2 predictor & 5,003 & 0.141 & 0.945 & 0.772 \\
All & Benchmark identity only & 5,003 & 0.141 & 0.796 & 0.376 \\
\midrule
AmazonSOP & Full pre-H2 predictor & 179 & 0.575 & 0.960 & 0.956 \\
BigCode & Full pre-H2 predictor & 126 & 0.024 & 0.978 & 0.421 \\
Hotpot Context & Full pre-H2 predictor & 548 & 0.036 & 0.984 & 0.761 \\
Hotpot FullWiki & Full pre-H2 predictor & 941 & 0.131 & 0.909 & 0.608 \\
LiveMath-meta & Full pre-H2 predictor & 145 & 0.110 & 0.799 & 0.365 \\
LiveMath-non-meta & Full pre-H2 predictor & 151 & 0.172 & 0.725 & 0.311 \\
SQuAD & Full pre-H2 predictor & 463 & 0.438 & 0.958 & 0.927 \\
Spreadsheet & Full pre-H2 predictor & 1,036 & 0.181 & 0.802 & 0.469 \\
UCSB SOP & Full pre-H2 predictor & 286 & 0.091 & 0.886 & 0.512 \\
SimpleQA & Full pre-H2 predictor & 1,128 & 0.000 & -- & -- \\
\bottomrule
\end{tabular}
}
\end{table}

\begin{table}[!htbp]
\caption{Offline replay of escalation policies over 7,488 Qwen3.5-9B task--method views.}
\label{tab:escalation-replay}
\centering
\scriptsize
\begin{adjustbox}{max width=\linewidth,max totalheight=0.85\textheight}
\begin{tabular}{lrrrr}
\toprule
Policy & Escalated (\%) & Accuracy (\%) & Regressions & Tokens \\
\midrule
Always H1 & 0.0 & 33.71 & 0.00 & 3,190 \\
Always H2 (no probe) & 100.0 & 41.60 & 2.20 & 3,729 \\
Escalate every H1 failure & 66.8 & 42.13 & 0.87 & 6,267 \\
Predictor top 10\% of H1 failures & 6.7 & 38.45 & 0.16 & 3,944 \\
Random 10\% of H1 failures & 6.7 & 34.55 & 0.09 & 3,498 \\
Predictor top 20\% of H1 failures & 13.4 & 40.02 & 0.74 & 4,389 \\
Random 20\% of H1 failures & 13.4 & 35.39 & 0.17 & 3,810 \\
Predictor top 30\% of H1 failures & 20.0 & 41.13 & 0.87 & 4,719 \\
Random 30\% of H1 failures & 20.0 & 36.24 & 0.26 & 4,112 \\
Predictor top 50\% of H1 failures & 33.4 & 42.01 & 0.87 & 5,661 \\
Random 50\% of H1 failures & 33.4 & 37.92 & 0.44 & 4,730 \\
Oracle (escalate rescuable H1 failures) & 9.4 & 41.46 & 0.13 & 4,022 \\
Oracle (escalate any view where H2 > H1) & 14.4 & 43.12 & 0.13 & 4,241 \\
\bottomrule
\end{tabular}
\end{adjustbox}
\end{table}

\FloatBarrier

\section{Routing Intervention Choice}
\label{app:routing}
\label{app:rq6-details}
\label{sec:prospective-routing}
\label{sec:routing-appendix}

\subsection{Direct Routing}
\label{app:direct-routing}

Every router has two stages. \textbf{Stage A} is the predictor: for each task and candidate action, a ridge regression predicts whether the action beats the reference (\TraceToSkill{}+H2 at Default budget), using only inputs available before the outcome, under nested leave-one-benchmark-family-out cross-validation. \textbf{Stage B} picks the action with the highest predicted gain, and its outcome and full call-chain token cost are replayed from the saved runs.

The capability boundaries in Section~\ref{app:rq5-details} leave headroom wherever several actions succeed. Whether that headroom is usable depends on choosing the right action in advance, without first paying a probe to find out. Table~\ref{tab:direct-routing} compares five policies that must choose from the task text alone, under held-family historical replay. Trace H1 and Trace H2 apply one harness level to every instance without routing. The Text+action Router predicts the action directly from task text, whereas Text$\rightarrow$root$\rightarrow$action first predicts a root cause and then maps that label to an action. The Observed per-instance Oracle picks the best action using matched outcomes and is an unattainable reference. Every policy acts on Qwen3.5-9B, so ranking by mean full-chain tokens ranks by cost.

\begin{table}[!htbp]
\caption{Direct routing without a paid H1 probe under held-family historical replay. The Observed per-instance Oracle is an unattainable reference using matched outcomes. The cross-repeat Oracle picks the action on one repeat and scores it on the other two.}
\label{tab:direct-routing}
\centering
\small
\begin{adjustbox}{max width=\linewidth,max totalheight=0.85\textheight}
\begin{tabular}{lrr}
\toprule
Policy & Performance & Mean full-chain tokens \\
\midrule
Trace H1 & 38.99\% & 2,853.51 \\
Trace H2 reference & 45.29\% & 4,070.32 \\
Text+action Router & 39.42\% & 3,620.89 \\
Text$\rightarrow$root$\rightarrow$action & 40.10\% & 4,264.12 \\
Observed per-instance Oracle & 58.12\% & 2,084.54 \\
Cross-repeat Oracle & 49.24\% & 2,359 \\
\bottomrule
\end{tabular}
\end{adjustbox}
\end{table}

Predicting the root cause first helps slightly, 40.10\% against 39.42\%, but that difference is not statistically resolved. Both routers stay below the Trace H2 reference of 45.29\%, so no deployable router is yet established, while the Oracle reaches 58.12\%, or 49.24\% when the action is picked on one repeat and scored on the other two. That gap makes intervention selection a value-of-information problem: acquiring the signal can consume the benefit of acting on it.

\subsection{Recovery Routing}
\label{app:recovery-routing}

\begin{table}[!htbp]
\caption{Recovery routing under held-family historical replay. Full-chain token means charge the H1 probe and selected H2 restarts.}
\label{tab:recovery-routing}
\centering
\begin{adjustbox}{max width=\linewidth,max totalheight=0.85\textheight}
\begin{tabular}{lrr}
\toprule
Policy & Performance & Mean full-chain tokens \\
\midrule
Always H1 & 37.17\% & 5,334.19 \\
Always H2 & 51.01\% & 10,243.14 \\
H1-conditioned router & 49.25\% & 13,677.49 \\
Observed H1/H2 Oracle & 54.20\% & 11,374.29 \\
\bottomrule
\end{tabular}
\end{adjustbox}
\end{table}

The recovery setting purchases one H1 execution and decides whether to keep it or restart with H2; the direct setting avoids that probe and chooses among tested Qwen configurations from task text. The first exposes an information-acquisition cost, while the second exposes an identifiability limit. All comparisons use complete recorded call-chain token accounting, including probes and restarts.

\subsection{Routing Results and Interpretation}
\label{app:routing-interpretation}

\paragraph{Routing headroom exists, but deployable signals do not yet realize it.}
Section~\ref{sec:which} shows why routing could matter: some failures admit several successful axes with different costs, while others require a particular intervention. An Observed per-instance Oracle verifies that this heterogeneity contains substantial quality--cost headroom. On the direct action set, it reaches 58.12\% with 2,085 mean full-chain tokens, compared with 45.29\% and 4,070 tokens for the Trace H2 reference. Thus, a policy that knew the matched outcomes could be both more accurate and cheaper. The Oracle obtains that advantage by reading the target outcomes it is supposed to predict, however, and is therefore an unattainable boundary rather than a deployable policy. Part of its gap is also selection on noise, because it picks and scores the action on the same runs. Picking the action on one repeat and scoring it on the other two leaves $+3.95$ points [$+1.58,+6.36$] over the reference (49.24\% versus 45.29\%), which is our main estimate of direct-routing headroom.

The tested deployment-valid information paths do not realize this headroom. The H1-conditioned router pays for an H1 probe before deciding whether to keep H1 or restart H2. It reaches 49.25\% with 13,677 mean full-chain tokens, whereas Always H2 reaches 51.01\% with 10,243 tokens. This router has both lower quality and higher cost and is therefore \emph{strictly dominated}. Direct text-conditioned routing avoids the probe, but the Text+action Router reaches only 39.42\%, and Text$\rightarrow$root$\rightarrow$action reaches 40.10\%, both below the 45.29\% Trace H2 reference. The latter also spends 4,264 tokens against 4,070 for Trace H2.

The result is a negative deployment finding rather than evidence that routing has no value: routing headroom exists, but none of the tested lightweight deployable routing policies achieves a quality--cost Pareto improvement. In the terms of Section~\ref{sec:which},
\[
\text{failure information}\;\not\Rightarrow\;\text{a cheap deployable decision signal}.
\]
These experiments are frozen historical replays rather than live prospective executions, and their money and latency ledgers remain incomplete. We therefore make no positive routing claim. Instead, the gap between the Observed Oracle and learned policies identifies the remaining problem: intervention selection is a value-of-information problem in which acquiring the signal can consume the benefit of acting on it.

\subsection{Online Escalation Policies}
\label{app:online-policies}
\begingroup
Both studies replay existing Default-budget test records: every configuration was already run on the same task instances and repeats, so a policy ``runs'' a configuration by reading its record. Rules, metrics, and success criteria were preregistered before any outcome was computed. Scores reproduce the matrix cells within 0.05 points on LiveMath, SimpleQA (common Qwen grader), SQuAD, Hotpot Context, and Hotpot FullWiki; Amazon SOP is excluded because its per-record API-model scores do not reproduce the matrix, and Spreadsheet and UCSB SOP lack token telemetry. Cost is total tokens priced at the provider audit's mean rate (\$4.29, \$3.05, and \$4.28 per million tokens for Sonnet~4.6, Sonnet~5, and GPT-5.6-sol); local Qwen3.5-9B runs are priced at zero, and the cost of labeling pilot tasks is not included.

\paragraph{Per-task signal cascade.} Qwen3.5-9B+\TraceToSkill{}+H2 runs first; if its trajectory ends with a logged error, no answer or an invalid final action, an unresolved tool error, an abstention, or a short-answer violation (more than 20 words on SQuAD or Hotpot), the same skill is run at H2 on the stronger model and its answer is returned. Table~\ref{tab:online-cascade} reports the result. The rule is precise (98--100\% of escalated tasks were Qwen failures) but catches 11\% of Qwen3.5-9B's failures, none on LiveMath; among the caught failures the stronger model repairs only 22--23\%, fewer than among the missed ones (32--45\%), because the signal selects hard items such as SimpleQA abstentions. The cascade beats random escalation at the same rate for the two Sonnet targets but fails the preregistered criterion.

\begin{table}[!htbp]
\centering
\caption{Per-task signal cascade (\TraceToSkill{}, H2; macro score over five benchmarks). Random escalates at the cascade's per-benchmark rate; the oracle escalates exactly the Qwen3.5-9B failures and needs gold labels. CI: 95\% task-cluster bootstrap of cascade minus Qwen3.5-9B. Cost in USD per 1,000 tasks.}
\label{tab:online-cascade}
\small
\begin{adjustbox}{max width=\linewidth,max totalheight=0.85\textheight}
\begin{tabular}{lrrrrrcc}
\toprule
Target & Qwen & Target & Cascade & Random & Oracle & Cascade $-$ Qwen & Cost (cascade / target) \\
\midrule
Sonnet 4.6 & 47.5 & 51.9 & 48.7 & 48.0 & 63.1 & [+0.7, +1.8] & 2.01 / 13.59 \\
Sonnet 5 & 47.5 & 52.8 & 48.7 & 48.1 & 64.1 & [+0.8, +1.8] & 0.47 / 9.77 \\
GPT-5.6-sol & 47.5 & 60.0 & 48.6 & 48.3 & 70.4 & [+0.7, +1.7] & 0.45 / 8.19 \\
\bottomrule
\end{tabular}
\end{adjustbox}
\end{table}

\paragraph{Benchmark-level labeled pilot.} For each benchmark with everything else fixed, both the default and the alternative configuration run on the first $k$ tasks of a random order, and the whole benchmark switches to the alternative if its pilot score exceeds the default's by at least $\delta$; every task receives the committed configuration's answer, and the pilot pays for both. Results average 200 random orders and all units of a decision (Table~\ref{tab:online-pilot}). H1 versus H2 leaves nothing to learn, since H2 rarely costs more and never scores lower by five points. H2 versus H3 meets the preregistered criterion only narrowly. The model decision gains most: the stronger model is the better choice in only 11--16 of 20 units, and a 20-task pilot recovers most of the oracle's advantage over always switching. With $k=50$ the gap to the oracle falls to 0.2--0.4 points for every decision, at a higher pilot cost.

\begin{table}[!htbp]
\centering
\caption{Benchmark-level labeled pilot with $k=20$ tasks (macro score / USD per 1,000 tasks). Units: model$\times$skill$\times$benchmark for the harness decisions, skill$\times$benchmark for the model decisions.}
\label{tab:online-pilot}
\small
\begin{adjustbox}{max width=\linewidth,max totalheight=0.85\textheight}
\begin{tabular}{lrccccc}
\toprule
Decision & Units & Always default & Always alternative & Pilot, $\delta=0$ & Pilot, $\delta=5$ & Oracle \\
\midrule
H1 $\rightarrow$ H2 & 80 & 47.4 / 6.63 & 48.6 / 6.76 & 48.8 / 7.40 & 48.7 / 7.34 & 49.0 / 6.72 \\
H2 $\rightarrow$ H3 & 80 & 48.6 / 6.76 & 49.1 / 12.64 & 49.2 / 11.29 & 49.0 / 9.17 & 49.7 / 10.24 \\
Qwen $\rightarrow$ Sonnet 4.6 & 20 & 42.3 / 0.00 & 48.6 / 9.44 & 50.4 / 5.50 & 49.9 / 4.42 & 51.1 / 4.47 \\
Qwen $\rightarrow$ Sonnet 5 & 20 & 42.3 / 0.00 & 49.9 / 9.29 & 50.8 / 5.71 & 50.2 / 4.42 & 51.7 / 4.70 \\
Qwen $\rightarrow$ GPT-5.6-sol & 20 & 42.3 / 0.00 & 53.5 / 8.32 & 57.4 / 5.31 & 57.3 / 4.38 & 58.1 / 4.89 \\
\bottomrule
\end{tabular}
\end{adjustbox}
\end{table}
\endgroup

\section{Benchmark Case Studies}
\label{app:benchmark-case-studies}

This section reads the factorial results benchmark by benchmark. Task properties are the human-authored design annotations in Table~\ref{tab:appendix-copy-task-properties}; trajectory roots use the eight operational definitions in Table~\ref{tab:appendix-root-taxonomy}. Root frequencies refer only to the 2,225 labeled \NoSkill{}+H1+Medium source failures in Table~\ref{tab:appendix-copy-root-composition}; they are not benchmark labels, and the Hotpot estimates are too sparse for benchmark-wide interpretation. ``Best'' denotes the largest observed three-repeat mean in the Default-budget matrix (Table~\ref{tab:default-matrix-transposed}). It is descriptive selection over observed cells rather than a prospective policy.

\providecommand{\casetakeaway}[1]{\par\medskip\noindent\fbox{\parbox{\dimexpr\columnwidth-2\fboxsep-2\fboxrule\relax}{\textbf{Takeaway.} #1}}\par\medskip}

\subsection{SpreadsheetBench}
\label{app:case-spreadsheet}

\paragraph{Context, properties, and failures.}
SpreadsheetBench requires stateful edits through a spreadsheet tool. We therefore annotate high tool/state interaction and feedback demand, with Execution/tool as the expected bottleneck. The observed source labels agree: 587 of 588 labeled failures are Execution/tool, with one Horizon/termination case. This unusually concentrated composition supports an execution interpretation for this source condition, although it does not imply that every configuration fails for the same reason.

\paragraph{Best observed configuration and mechanism.}
The best Default cell is Sonnet 4.6+\NoSkill{}+H3 at 53.1. The same Model rises from 48.6 at H0 to 49.2/51.1/53.1 under H1/H2/H3, while every supplied Skill lowers its H3 score. Qwen shows a larger execution-support response: \NoSkill{} rises from 9.3 at H1 to 24.9 at H2 and 30.4 at H3. The pattern is consistent with feedback, bounded repair, and revision addressing runtime failures, while a frozen procedure can distract from the current workbook state.

\casetakeaway{Stateful spreadsheet work is primarily execution-limited in this matrix. A capable Model with observable execution and revision is more useful than an additional frozen procedure, and Skill injection should be treated as a possible source of interference.}

\subsection{LiveMathematicianBench}
\label{app:case-livemath}

\paragraph{Context, properties, and failures.}
LiveMath is a multiple-choice mathematical reasoning task with low tool demand and a recurring meta-option. Its source labels are 73.44\% Reasoning and 26.23\% State/grounding, both model-assisted hypotheses, plus 0.33\% Procedural. The post-hoc meta-option partition in Appendix~\ref{app:livemath-meta} shows that a reusable answer-selection regularity exists inside the benchmark but does not cover all instances.

\paragraph{Best observed configuration and mechanism.}
The best Default cell is GPT-5.6-sol+\TraceToSkill{}+H1 at 62.1. For GPT-5.6-sol, \TraceToSkill{} scores 61.6/62.1/57.8/60.5 across H0--H3; for Qwen it scores 53.2/58.1/56.5/54.6, far above Qwen+\NoSkill{} at 34.1/34.9/34.4/36.3. In contrast, additional Harness support is flat or non-monotone, and Sonnet 5+\TraceToSkill{} declines from 40.6 at H0 to 35.5 at H3. The benefit is therefore better explained by a reusable decision procedure plus model competence than by execution recovery.

\casetakeaway{LiveMath rewards a reusable but narrow decision regularity. Trace-derived procedures help broadly across observed contexts, yet subgroup reversal shows why the Skill should not be interpreted as a general increase in mathematical capability.}

\subsection{UCSB SOPBench}
\label{app:case-ucsb}

\paragraph{Context, properties, and failures.}
UCSB SOPBench requires following long, constraint-sensitive standard operating procedures while interacting with state. We annotate high procedural and tool demand. Among labeled source failures, 78.57\% are Procedural, 13.91\% State/grounding, and 7.52\% Execution/tool, giving the most direct root-level support for a procedure-oriented Skill hypothesis.

\paragraph{Best observed configuration and mechanism.}
The best Default cell is GPT-5.6-sol+\SkillOpt{}+H1 at 73.1, narrowly above H0 at 72.6; H2 and H3 fall to 71.6 and 70.3. Qwen shows the same Skill preference: \SkillOpt{} reaches 66.3 at H0/H1 versus 52.8/51.0 for \NoSkill{}. Outcome-guided optimization can encode ordering and verification constraints, whereas additional recovery or reflection is not uniformly useful once that policy is supplied.

\casetakeaway{For procedure- and constraint-heavy SOP execution, an outcome-optimized reusable procedure is the stable first intervention. Harness choice remains model-dependent and stronger recovery is not automatically better after the procedure is supplied.}

\subsection{Amazon SOP}
\label{app:case-amazon}

\paragraph{Context, properties, and failures.}
Amazon SOP is a long, state-changing inspection workflow with high procedure, tool, horizon, and feedback demand. All 113 labeled source failures are Execution/tool. Horizon is nevertheless a task-level design hypothesis rather than an observed root conclusion, because the frozen Medium-budget source cohort contains almost no labeled Horizon/termination failures.

\paragraph{Best observed configurations and mechanism.}
The Default ceiling is 100.0 and is non-unique. It is reached by Sonnet 4.6+\NoSkill{}+H2, all four Harnesses with Sonnet 4.6+\SkillOpt{}, Sonnet 4.6+\TraceToSkill{}+H2, GPT-5.6-sol+\NoSkill{}+H2, and GPT-5.6-sol+\SkillOpt{}+H2. The decisive transition is often execution feasibility: Qwen+\NoSkill{} rises 33.9$\rightarrow$88.6 from H1 to H2, and GPT-5.6-sol+\NoSkill{} rises 57.1$\rightarrow$100.0. A mismatched Skill can instead collapse the path: Qwen+\Direct{} remains at 1.7 across H0--H3. The Low-to-best Budget envelope is 81.13 points, consistent with a long workflow crossing a completion threshold rather than smooth scaling.

\casetakeaway{Amazon SOP is governed by path feasibility and execution reliability. H2 or sufficient Budget can unlock completion, but a mismatched procedure can prevent that path entirely; after feasibility is reached, several configurations saturate at the ceiling.}

\subsection{SimpleQA}
\label{app:case-simpleqa}

\paragraph{Context, properties, and failures.}
SimpleQA asks for short factual answers without supplied evidence or meaningful tool interaction. The labeled source failures are 74.84\% Factual knowledge/recall, 24.78\% Abstention/calibration, and 0.38\% Evaluation/answer contract; the first two categories are model-assisted hypotheses. This composition aligns with a Model intervention more directly than with a procedural or execution intervention.

\paragraph{Best observed configuration and mechanism.}
The best Default cell is GPT-5.6-sol+\NoSkill{}+H0 at 69.4. GPT-5.6-sol remains near this level under H1--H3 (67.8/68.0/68.8), whereas Qwen+\NoSkill{} is 10.7 at H0 and 4.7 at H2. Neither execution feedback nor a frozen procedure supplies a fact absent from the task and not recalled by the Model. The within-axis Model span is 60.85 points, compared with Skill, Harness, and Budget envelopes of 0.13, 0.00, and 0.10.

\casetakeaway{When the missing resource is factual recall, changing the Model dominates procedural and execution support. SimpleQA is the clearest case in which additional Harness, Skill, or Budget mostly adds cost without supplying the missing information.}

\subsection{SQuAD}
\label{app:case-squad}

\paragraph{Context, properties, and failures.}
SQuAD supplies the evidence passage, so the primary challenge in this setup is returning a creditable answer through the expected interface. All 159 labeled source failures are Evaluation/answer contract. This is strong evidence for the audited source condition, but the native-score floor analysis in Appendix~\ref{sec:measurement-floors} means large H0$\rightarrow$H1 gains should be read as contract specification rather than reasoning recovery.

\paragraph{Best observed configuration and mechanism.}
The best Default cell is Sonnet 5+\TraceToSkill{}+H1 at 91.0, followed closely by H2/H3 at 90.6/90.7. The interface transition is large: Qwen+\NoSkill{} rises 10.7$\rightarrow$36.6 from H0 to H1 and then 78.6/79.3 under H2/H3, while Sonnet 4.6+\NoSkill{} rises from the 0.0 floor to 85.4 at H1. A Skill can partially supply an extraction/answer procedure at H0, but once the Harness specifies the contract its incremental value is often absorbed.

\casetakeaway{SQuAD primarily tests whether the system exposes and satisfies the answer contract. H1 removes a measurement/interface bottleneck; after that transition, additional execution support or Skill often has little incremental value.}

\subsection{HotpotQA Context}
\label{app:case-hotpot-context}

\paragraph{Context, properties, and failures.}
Hotpot Context supplies gold and distractor paragraphs, reducing retrieval demand while retaining multi-hop evidence integration. We annotate Evaluation/answer contract as the expected bottleneck. Only one source failure is labeled, also as Evaluation/answer contract, so root composition cannot explain the benchmark; this is a coverage limitation rather than confirming evidence.

\paragraph{Best observed configuration and mechanism.}
The best Default cell is Sonnet 4.6+\Direct{}+H3 at 69.2. The main transition is H0$\rightarrow$H1: that configuration rises 16.6$\rightarrow$66.8, and Sonnet 4.6+\NoSkill{} rises 0.0$\rightarrow$66.2. Later Harness levels add little. Skill compatibility remains important: GPT-5.6-sol+\SkillOpt{} stays at 0.0/13.6/12.2/12.2, compared with 29.7/64.6/64.3/60.9 for \Direct{}. The former is consistent with a supplied policy overriding the answer contract, not with a graded loss of reasoning ability.

\casetakeaway{When evidence is already provided, interface compatibility dominates retrieval support. H1 captures most of the observed gain, and an incompatible Skill can erase it; the sparse root labels do not support a benchmark-wide mechanistic claim.}

\subsection{HotpotQA FullWiki}
\label{app:case-hotpot-fullwiki}

\paragraph{Context, properties, and failures.}
Hotpot FullWiki uses the same question UIDs as Context but removes the supplied passages and permits retrieval from a frozen Wikipedia index. This creates multi-hop search, grounding, execution, and horizon demands. All ten labeled source failures are Execution/tool, but ten labels are far too few to characterize the benchmark because missingness is concentrated in the Hotpot views.

\paragraph{Best observed configuration and mechanism.}
The best Default cell is Sonnet 5+\TraceToSkill{}+H1 at 45.0, compared with 30.9 for Sonnet 5+\NoSkill{} at H1. Qwen reveals the interaction mechanism: \Direct{} rises from 3.9 at H1 to 28.4 at H2, \SkillOpt{} from 24.0 to 34.9, and \TraceToSkill{} from 34.1 to 36.0. A reusable retrieval policy supplies what to search and how to integrate evidence; executable observations and recovery determine whether weaker configurations can carry out that policy. The strongest cell nevertheless uses H1, showing that more recovery is not universally necessary once Model and Skill are well matched.

\casetakeaway{FullWiki couples a reusable multi-hop retrieval policy with the ability to execute it. Trace2Skill is the stable Skill choice in the observed matrix, while Harness support is most valuable when the current Model--Skill pair cannot reliably realize that policy.}

\subsection{BigCodeBench}
\label{app:case-bigcode}

\paragraph{Context, properties, and evidence boundary.}
BigCodeBench-Hard requires program synthesis and execution against evaluator-private tests. We annotate medium procedural and horizon demand, high tool interaction and feedback demand, and Execution/tool as the expected bottleneck. We use all 148 available instances across frozen construction, validation, qualification, and test roles (Table~\ref{tab:benchmark-splits}).

\paragraph{Why no comparable winner is reported.}
BigCode is retained as supporting appendix evidence because its evaluator environment is not fully comparable across Models. It is therefore excluded from the 512-cell Table~\ref{tab:default-matrix-transposed} and the root-conditioned common-support cohort. Reporting a cross-model ``best intervention combination'' from those analyses would silently mix evaluator differences with intervention effects.

\paragraph{Interpretation.}
The task structure motivates executable feedback and Model capability as hypotheses, but the present evidence cannot distinguish them on the same matched scale used for the other eight benchmarks. A valid case analysis requires rerunning every Model under one frozen execution image, scorer, dependency set, timeout policy, and task split, followed by the same three-repeat factorial and root-label protocol.

\casetakeaway{Code execution is a plausible feedback-sensitive setting, but evaluator comparability is a prerequisite for attributing gains to Model, Skill, Harness, or Budget. BigCode remains supporting evidence rather than a ninth point in the comparative mechanism claims.}

\subsection{Cross-Case Synthesis}
\label{app:case-synthesis}

The cases separate four recurring structures without turning them into universal rules: missing factual knowledge favors Model changes; reusable decision, SOP, or retrieval regularities can favor Skills; observable runtime or answer-contract failures can favor Harness support; and long workflows can cross a Budget feasibility threshold. The categorical and boundary analyses show why these relations remain conditional. The mean-optimal observed configuration across the eight normalized benchmarks is GPT-5.6-sol+\TraceToSkill{}+H2 (mean 0.8515), whereas the maximin configuration is GPT-5.6-sol+\TraceToSkill{}+H3 (worst-benchmark score 0.6301, mean 0.8439). Relative to H2, the latter sacrifices 0.0076 mean performance for a 0.0417 increase in the worst-benchmark floor. Objective choice therefore remains part of intervention choice, even after the benchmark context is known.

\casetakeaway{Across cases, intervention value follows the bottleneck that remains in the current configuration. Benchmark identity narrows the plausible mechanism, but task-instance heterogeneity, component interactions, and measurement boundaries prevent a context-free intervention rule.}

\end{document}